\documentclass{article}
\usepackage[margin=1.5in]{geometry}

\usepackage[utf8]{inputenc} 
\usepackage[T1]{fontenc}    

\usepackage{setspace}	
\usepackage[colorlinks=true]{hyperref}
\usepackage{url}            
\usepackage{enumerate}
\usepackage{ifthen}
\usepackage{sidecap}
\usepackage{helvet}
\usepackage{fancyhdr} 
\usepackage{lastpage} 
\usepackage{comment}

\usepackage{amsmath}
\usepackage[mediummath]{nccmath} 
\usepackage{amsfonts}	
\usepackage{mathtools}	
\usepackage{amsthm}	
\usepackage{amssymb} 
\usepackage{nicefrac}       
\usepackage{microtype}      
\usepackage{dsfont} 
\usepackage{upgreek} 

\usepackage{eurosym}

\usepackage{graphicx}
\usepackage{caption}
\usepackage{subcaption}
\usepackage{wrapfig}
\usepackage{float} 

\usepackage{booktabs}       
\usepackage{multirow} 
\usepackage[table]{xcolor}
\usepackage{tabularx}
\usepackage{arydshln} 
\usepackage{makecell} 
\usepackage{ragged2e} 
\newcolumntype{L}[1]{>{\raggedright\let\newline\\\arraybackslash\hspace{0pt}}m{#1}}
\newcolumntype{C}[1]{>{\centering\let\newline\\\arraybackslash\hspace{0pt}}m{#1}}
\newcolumntype{J}[1]{>{\vspace*{-2ex}\justify\let\newline\\\arraybackslash\hspace{0pt}}m{#1}}
\newcolumntype{R}[1]{>{\raggedleft\let\newline\\\arraybackslash\hspace{0pt}}m{#1}}

\usepackage{tikz}
\usetikzlibrary{bayesnet} 

\usepackage{algorithm}
\usepackage{algorithmic}

\newcommand{\ie}{i.e., }

\newcommand{\eg}{e.g., }

\newcommand{\Real}{{\mathbb R}} 
\newcommand{\Natural}{{\mathbb N}} 

\newcommand{\Prob}[1]{\mathbb{P}\left( #1 \right)}
\newcommand{\myProb}[2]{\mathbb{P}_{#1}\left( #2 \right)}
\newcommand{\eValue}[2]{\mathbb{E}_{#1}\left\{ #2 \right\}}

\newcommand{\N}[1]{\mathcal{N}\left( #1\right)}

\newcommand{\T}[1]{t\left( #1\right)}

\newcommand{\Cat}[1]{{\rm Cat}\left( #1\right)}

\newcommand{\IG}[1]{\Gamma^{-1}\left( #1\right)}

\newcommand{\myind}[1]{\mathds{1}\left[#1\right]}
\newcommand{\dd}[1]{\mathrm{d} #1}

\newcommand{\A}{\mathcal{A}}

\newcommand{\X}{\mathcal{X}}

\newcommand{\HH}{\mathcal{H}}

\newcommand{\argmax}{\mathop{\mathrm{argmax}}}

\usepackage[square,colon]{natbib}

\usepackage{authblk}
\begin{document}

\title{Sequential Monte Carlo Bandits}

\author[1, 2]{
	I\~{n}igo Urteaga
}
\author[3]{
	Chris H.~Wiggins
}

\affil[ ]{{\footnotesize 
		\sf iurteaga@bcamath.org \qquad \qquad chris.wiggins@columbia.edu
	}
}

\affil[ ]{{\footnotesize}}
	
\affil[1]{
	{\small
	BCAM - Basque Center for Applied Mathematics,
	Bilbao,
	Spain
	}
}
\affil[2]{
	{\small 
	IKERBASQUE, Basque Foundation for Science,
	Bilbao,
	Spain
	}
}
\affil[3]{
	{\small
	Department of Applied Physics and Applied Mathematics,
	Columbia University,
	New York City,
	NY, USA
	}
}

\maketitle

\begin{abstract}
We extend Bayesian multi-armed bandit (MAB) algorithms
beyond their original setting
by making use of sequential Monte Carlo (SMC) methods.
A MAB is a sequential decision making problem
where the goal is to learn a policy that maximizes long term payoff,
where only the reward of the executed action is observed.
In the stochastic MAB, the reward for each action is generated from an unknown distribution, often assumed to be stationary.
To decide which action to take next,
a MAB agent must learn the characteristics of the unknown reward distribution, \eg compute its sufficient statistics.
However, closed-form expressions for these statistics are analytically intractable except for simple, stationary cases.
We here utilize SMC for estimation of the statistics Bayesian MAB agents compute,
and devise flexible policies that can address a rich class of bandit problems:
\ie MABs with nonlinear, stateless- and context-dependent reward distributions that evolve over time.
We showcase how non-stationary bandits,
where time dynamics are modeled via linear dynamical systems,
can be successfully addressed by SMC-based Bayesian bandit agents.
We empirically demonstrate good regret performance of the proposed SMC-based bandit policies in several MAB scenarios that have remained elusive,
\ie in non-stationary bandits with nonlinear rewards.
\end{abstract}

\section{Introduction}
\label{intro}

The multi-armed bandit (MAB) problem considers
the sequential strategy one must devise when playing a row of slot machines:
\ie which arm to play next to maximize cumulative returns.
This analogy extends to a wide range of real-world challenges
that require online learning, while simultaneously maximizing some notion of reward. 

The arm may be a medicine a doctor must prescribe to a patient,
the reward being the outcome of such treatment on the patient;
or the set of resources a manager needs to allocate for competing projects,
with the reward being the revenue attained at the end of the month;
or the ad/product/content an online recommendation algorithm must display
to maximize click-through rate in e-commerce.

The contextual MAB,
where at each interaction with the world
side information (known as `context') is available,
is a natural extension of this abstraction.
The `context' is the physiology of the patient,
the type of resources available for each project,
or the features of the website user.

Sequential decision processes have been studied for many decades, 
and interest has resurged
incited by reinforcement learning (RL) advancements developed within the machine learning community~\citep{j-Mnih2015,j-Silver2017}.
RL~\citep{b-Sutton1998} has been successfully applied to a variety of domains,
from Monte Carlo tree search~\citep{ic-Bai2013} and hyperparameter tuning for complex optimization in science, engineering and machine learning problems~\citep{ip-Kandasamy2018, Urteaga2023},
to revenue maximization~\citep{j-Ferreira2018} and marketing solutions~\citep{j-Schwartz2017} in business and operations research.
RL is also popular in e-commerce and digital services,
improving online advertising at LinkedIn~\citep{ip-Agarwal2013},
engagement with website services at Amazon~\citep{ip-Hill2017},
recommending targeted news at Yahoo~\citep{ip-Li2010},
and enabling full personalization of content and art at \citet{Netflix2017}.

The techniques used in these success stories are grounded on statistical advances on sequential decision processes and multi-armed bandits.
The MAB crystallizes the fundamental trade-off between exploration and exploitation in sequential decision making.
It has been studied throughout the 20th century, with important contributions
by \citet{j-Thompson1935} and later \citet{j-Robbins1952}.
Over the years, several algorithms have been proposed ---we provide an overview of state-of-the-art solutions in Section~\ref{ssec:mab}.
However, applied use cases raise challenges that these MAB algorithms often fail to address.

For instance, classic MAB algorithms do not typically generalize to problems
with nonlinear reward dependencies or non-Gaussian reward distributions,
as exact computation of their statistics of interest is intractable for distributions not in the exponential family~\citep{ic-Korda2013,j-Russo2018}.
More importantly, these algorithms are commonly designed under the assumption of stationary reward distributions,
\ie the reward function does not change over-time,
a premise often violated in practice.

We hereby relax these constraints,
and consider time-varying models and nonlinear reward functions.
We propose to use sequential Monte Carlo (SMC) for non-stationary bandits with nonlinear rewards,
where the world ---the reward function--- is time-varying,
and rewards are sequentially observed for the played arms.

SMC methods~\citep{j-Arulampalam2002,b-Doucet2001,j-Djuric2003} have been widely used
to estimate posterior densities and expectations in sequential problems with probabilistic models
that are too complex to treat analytically,
with many successful applications in science and engineering \citep{b-Ristic2004,j-Leeuwen2009,j-Ionides2006,j-Creal2012}.

In Bayesian MAB algorithms,
the agent must compute sufficient statistics of each arm's rewards over time,
for which sequential updates to the posterior of the parameters of interest must be computed.
We here show that SMC-based, sequentially updated random measures of per-arm parameter posteriors,
enable computation of any statistic a Bayesian MAB policy might require.

We generalize existing MAB policies beyond their original stationary setting,
and accommodate complex reward models:
those for which sampling may be performed even if analytic computation of summary statistics is infeasible.
We study latent dynamical systems with non-Gaussian and nonlinear reward functions,
for which SMC computes accurate posterior approximations.
Consequently, we devise a flexible SMC-based framework for solving non-stationary and nonlinear MABs.

Our \textbf{contribution} is a SMC-based MAB framework that:
\begin{enumerate}[(i)]
	\item computes SMC-based random measure posterior MAB densities utilized by Bayesian MAB policies;
	\item requires knowledge of the reward function only up to a proportionality constant,
		\ie it accommodates nonlinear and non-Gaussian bandit rewards; and,
	\item is applicable to time-varying reward models, \ie to restless or non-stationary multi-armed bandits.
\end{enumerate}

The proposed SMC-based MAB framework 
($i$) leverages SMC for both posterior sampling and estimation of sufficient statistics utilized by Bayesian MAB policies,
\ie Thompson sampling and Upper Confidence Bounds; 
($ii$) addresses restless bandits via the general linear dynamical system, and accommodates unknown parameters via Rao-Blackwellization; and
($iii$) targets nonlinear and non-Gaussian reward models,
accommodating stateless and context-dependent, discrete and continuous reward distributions. 

We introduce in Section~\ref{sec:background} the preliminaries for our work,
which combines sequential Monte Carlo techniques described in Section~\ref{ssec:smc},
with multi-armed bandit algorithms detailed in Section~\ref{ssec:mab}.
We present the SMC-based MAB framework in Section~\ref{sec:smc_mab},
and evaluate its performance for Thompson sampling and Bayes-Upper Confidence Bound policies in Section~\ref{sec:evaluation}.
We summarize and conclude with promising research directions in Section~\ref{sec:conclusion}.

\section{Background and preliminaries}
\label{sec:background}
\subsection{Multi-armed bandits}
\label{ssec:mab}

The MAB crystallizes the fundamental trade-off between exploration and exploitation in sequential decision making.
It formulates the problem of maximizing rewards observed from sequentially chosen actions $a\in\A$
---named \textit{arms} in the bandit literature---
when interacting with an uncertain environment.
The reward generating process is stochastic,
often parameterized with $\theta \in \Theta$
to capture the intrinsic properties of each arm.
It can potentially depend on context $x\in \X$; \eg a common choice is $\X=\Real^{d_X}$.
We use $p_{a}(\cdot |x,\theta)$ to indicate
per-arm reward distributions ---one for each of the $|\A|$ possible arms---
where subscript $_a$ indicates the conditional reward distribution for each arm $a$.

At each bandit interaction $t$, reward $y_t$ is observed for the played arm $a_t\in\A$ only,
which is independently and identically drawn from its context-conditional distribution
\begin{equation}
Y_t\sim p_{a}(Y|x_t,\theta_{t,a}^*) \;,
\end{equation}
parameterized by true $\theta_{t,a}^* \in \Theta$.
We use $Y_t$ for the stochastic reward variable with density $p_{a}(Y|x_t,\theta_{t,a}^*)$,
and denote with $y_t$ its realization at time $t$.
Recall that we accommodate time-varying context and parameters via the subscript $_t$ in both.

We denote with $\theta_t^*$ the union of all, per-arm, parameters at time $t$,
$\theta_t^* \equiv \left(\theta_{t,0}^*, \cdots, \theta_{t,|\A|-1}^* \right)$,
and with $\theta_{1:T}^*\equiv \left( \theta_{1}^*, \cdots, \theta_{T}^* \right)$,
the union of parameters over bandit interactions or time $t=1,\cdots,T$.
The above stochastic MAB formulation covers stationary bandits
(if parameters are constant over time, \ie $\theta_{t,a}^*=\theta_a^*, \; \forall t$)
and non-contextual bandits, by fixing the context to a constant value $x_t=x, \forall t$.

With knowledge of the true bandit model,
\ie the $\theta_t^* \in \Theta$
that parameterizes the reward distribution of the environment,
the optimal action to take is
\begin{equation}
a_t^* = \argmax_{a^\prime \in \A} \mu_{t,a^\prime}(x_t,\theta_t^*) \;,
\end{equation}
where $\mu_{t,a}(x_t,\theta_t^*)=\eValue{}{Y|a,x_t,\theta_t^*}$ is each arm's conditional reward expectation,
given context $x_t$ and true parameters $\theta_t^*$, at time $t$.

The challenge in MABs is the lack of knowledge about the reward-generating distribution,
\ie uncertainty about $\theta_t^*$ induces uncertainty about the true optimal action $a_t^*$.
Namely, the agent needs to simultaneously learn properties of the reward distribution,
and sequentially decide which action to take next.
MAB policies choose the next arm to play,
with the goal of maximizing attained rewards, based upon the history observed so far.

We use $\pi(A)$ to denote a multi-armed bandit policy,
which is in general stochastic ---$A$ is a random variable--- on its choices of arms,
and is dependent on previous history:
$\pi(A)=\myProb{}{A=a | \HH_{1:t}}, \forall a\in\A$.
Previous history $\HH_{1:t}$ contains the set of contexts, played arms, and observed rewards up to time $t$,
denoted as $\HH_{1:t}=\left\{x_{1:t}, a_{1:t}, y_{1:t}\right\}$,
with $x_{1:t} \equiv \left(x_1, \cdots , x_t\right)$,
$a_{1:t} \equiv \left(a_1, \cdots , a_t\right)$
and $y_{1:t} \equiv \left(y_{1,a_1}, \cdots , y_{t,a_t}\right)$.

Given history $\HH_{1:t}$, a MAB policy $\pi(A|\HH_{1:t})$ aims at maximizing its cumulative rewards,
or equivalently, 
minimizing its cumulative regret
(the loss incurred due to not knowing the best arm $a_t^*$ at each time $t$),
\ie $R_T=\sum_{t=1}^T y_{t,a^*_t}-y_{t,a_t}$,
where $a_t$ denotes the realization of the policy $\pi(A|\HH_{1:t})$
---the arm picked by the policy--- at time $t$.
Due to the stochastic nature of the problem,
we study the \emph{expected} cumulative regret at time horizon $T$ (not necessarily known a priori)
\begin{equation}
R_T=\eValue{}{\sum_{t=1}^T Y_{t,a^*_t}-Y_{t,A_t} } \; ,
\label{eq:mab_cumulative_regret}
\end{equation}
where the expectation is taken over the randomness in the outcomes $Y$, and the arm selection policy $A_t \sim \pi(A)$
for the frequentist regret.
In the Bayesian setting, the uncertainty over the true model parameters $\theta^*$ is also marginalized.

\subsubsection{MAB algorithms}
\label{sssec:mab_algos}
Over the years, many MAB policies have been proposed to overcome the exploration-exploitation tradeoff~\citep{b-Lattimore2020}.
$\epsilon$-greedy is a popular applied framework due to its simplicity
(\ie to be greedy with probability $1-\epsilon$,
and to play the arm with best averaged rewards so far,
otherwise to randomly pick any arm),
while retaining often good performance \citep{j-Auer2002}.
A more formal treatment was provided by
\citet{j-Gittins1979},
who devised the optimal strategy for certain bandit cases,
by considering geometrically discounted future rewards.
Since the exact computation of the Gittins index is complicated,
approximations have also been developed~\citep{j-Brezzi2002}.

\citet{j-Lai1985} introduced a new class of algorithms,
based on the upper confidence bound (UCB) of the expected reward of each arm,
for which strong theoretical guarantees have been proven~\citep{j-Lai1987},
and many extensions proposed~\citep{ip-Garivier2011,ip-Garivier2011a}.

Bayes-UCB \citep{ip-Kaufmann2012} is a Bayesian approach to UCB algorithms,
where quantiles are used as proxies for upper confidence bounds.
\citet{ip-Kaufmann2012} have proven the asymptotic optimality of Bayes-UCB's finite-time regret for the Bernoulli case,
and argued that it provides an unifying framework for several variants of the UCB algorithm.
However, its application is limited to reward models where the quantile functions are analytically tractable.

Thompson sampling (TS)~\citep{j-Thompson1935} is an alternative MAB policy that has been popularized in practice, and studied theoretically by many.
TS is a probability matching algorithm that randomly selects an action to play according to the probability of it being optimal~\citep{j-Russo2018}.
It has been empirically proven to perform satisfactorily, 
and to enjoy provable optimality properties,
both for problems with and without context \citep{ip-Agrawal2012,ip-Agrawal2013,ic-Korda2013,j-Russo2014,j-Russo2016}.

Bayes-UCB and TS can be viewed as different approaches to a Bayesian formulation of the MAB problem.
Namely, the agent views the unknown parameter of the reward function $\theta_t$ as a random variable, 
and as data from bandit interactions with the environment are collected,
a Bayesian policy updates its parameter posterior.
Because a bandit agent must take into account the uncertainty on the unknown parameters,
prior knowledge on the reward model and its parameters can be incorporated into Bayesian policies,
capturing the full state of knowledge via the parameter posterior
\begin{equation}
p(\theta_t|\HH_{1:t}) \propto p_{a_t}(y_t|x_t,\theta_t)p(\theta_t| \HH_{1:t-1}) \; ,
\label{eq:mab_param_posterior}
\end{equation}
where $p_{a_t}(y_t | x_t, \theta_t)$ is the likelihood of the observed reward $y_t$ after playing arm $a_t$ at time $t$.
Computation of this posterior is critical for Bayesian MAB algorithms.

In Thompson sampling,
one uses $p(\theta_t|\HH_{1:t})$ to compute the probability of an arm being optimal,
\ie $\pi(A|x_{t+1},\HH_{1:t}) = \Prob{A=a_{t+1}^*|x_{t+1}, \theta_t, \HH_{1:t}}$,
where the uncertainty over the parameters must be accounted for~\citep{j-Russo2018}.

Namely,
one marginalizes the posterior parameter uncertainty after observing  history $\HH_{1:t}$ up to time instant $t$, \ie
\begin{equation}
\begin{split}
\pi(A|x_{t+1},\HH_{1:t})&=\Prob{A=a_{t+1}^*|x_{t+1},\HH_{1:t}} \\
&= \int \Prob{A=a_{t+1}^*|x_{t+1},\theta_t,\HH_{1:t}} p(\theta_t|\HH_{1:t}) \dd{\theta} \\
&=\int \myind{A=\argmax_{a^\prime \in \A} \mu_{t+1,a^\prime}(x_{t+1},\theta_t)} p(\theta_t|\HH_{1:t}) \dd{\theta_t} \; .
\end{split}
\label{eq:theta_unknown_pr_arm_optimal}
\end{equation}

In Bayes-UCB,
$p(\theta_t|\HH_{1:t})$ is critical to determine the distribution of the expected rewards, \ie
\begin{equation}
p(\mu_{t+1,a}(x_{t+1})) = \int p(\mu_{t+1,a}|x_{t+1},\theta_{t}) p(\theta_t|\HH_{1:t}) \dd{\theta_t} \;,
\label{eq:density_expected_rewards}
\end{equation}
which is required for computation of the expected reward quantile $q_{t+1,a}(\alpha_{t})$, formally defined as
\begin{equation}
\Prob{\mu_{t+1,a}(x_{t+1})>q_{t+1,a}(\alpha_{t})}=\alpha_{t} \;,
\label{eq:quantile_expected_rewards}
\end{equation}
where the quantile value $\alpha_t$ may depend on time, as proposed by~\citet{ip-Kaufmann2012}.

Analytical expressions for the parameter posterior of interest $p(\theta_t|\HH_{1:t})$ are available only for few reward functions (\eg Bernoulli and linear contextual Gaussian models),
but not for many other useful cases, such as logistic or categorical rewards.
In addition,
computation of Equations~\eqref{eq:theta_unknown_pr_arm_optimal} and \eqref{eq:quantile_expected_rewards} can be challenging for many distributions outside the exponential family~\citep{ic-Korda2013}.
These issues become even more imperative
when dealing with dynamic parameters, \ie in environments that evolve over time,
and with nonlinear reward distributions.

\subsubsection{Beyond linear MABs.}
\label{sssec:mab_algos_complex}
To extend MAB algorithms to more realistic scenarios,
many have considered flexible reward functions and Bayesian inference.
For example,
the use of Laplace approximations~\citep{ic-Chapelle2011} 
or Polya-Gamma augmentations~\citep{ic-Dumitrascu2018}
for Thompson sampling.
These techniques however, are targeted to binary rewards only, modeled via the logistic function.

To accommodate complex, continuous reward functions, 
the combination of Bayesian neural networks with approximate inference has also been investigated.
Variational methods, stochastic mini-batches, and Monte Carlo techniques have been studied for uncertainty estimation of reward posteriors of these models~\citep{ip-Blundell2015, ic-Kingma2015, ic-Osband2016, ip-Li2016}.

\citet{ip-Riquelme2018} benchmarked some of these techniques, and reported that neural networks with approximate inference, even if successful for supervised learning, under-perform in the MAB setting.
In particular, \citet{ip-Riquelme2018} emphasize
the need for adapting the slow convergence uncertainty estimates of neural net based methods
for a successful identification of the exploration-exploitation tradeoff.

In parallel,
others have investigated how to extend Bayesian policies, such as Thompson sampling, 
to complex online problems~\citep{ip-Gopalan2014}
by leveraging ensemble methods~\citep{ip-Lu2017},
generalized sampling techniques~\citep{j-Li2013},
or via bootstrapped sampling ~\citep{j-Eckles2019,j-Osband2015}.
Solutions that approximate
the unknown bandit reward function with finite~\citep{ip-Urteaga2018}
or countably infinite Gaussian mixture models~\citep{j-Urteaga2018a} have also been proposed.

However, all these algorithms for MABs with complex rewards assume stationary distributions.

\subsubsection{Non-stationary MABs.}
\label{sssec:mab_algos_dynamic}
The study of bandits in a changing world go back to the work by Whittle~\citep{j-Whittle1988},
with subsequent theoretical efforts by many on characterizing restless, or non-stationary, bandits~\citep{j-Auer2002a, j-Bubeck2012}.
The special case of piecewise-stationary, or abruptly changing environments, has attracted a lot of interest in general~\citep{ip-Yu2009,ip-Luo2018},
and for UCB~\citep{ip-Garivier2011} and Thompson sampling~\citep{ip-Mellor2013} algorithms, in particular.
Often, these impose a reward `variation' constraint on the evolution of the arms~\citep{j-Raj2017},
or target specific reward functions, such as Bernoulli rewards in~\citep{ic-Besbes2014},
where discounting parameters for the prior Beta distributions can be incorporated.

More flexible restless bandit models,
based on the Brownian motion or discrete random walks~\citep{ip-Slivkins2008},
and simple Markov models~\citep{ip-Bogunovic2016} have been proposed,
showcasing the trade-off between the time horizon and the rate at which the reward function varies.
Besides, theoretical performance guarantees have been recently established for Thompson sampling in restless environments
where the bandit is assumed to evolve via a binary state Markov chain,
both in the episodic \citep{ip-Jung2019} and non-episodic~\citep{j-Jung2019} setting.

Here, we overcome constraints on both the bandit's assumed reward function and its time-evolving model,
by leveraging sequential Monte Carlo (SMC).
The use of SMC in the context of bandit problems was previously considered for probit~\citep{j-Cherkassky2013} and softmax~\citep{Urteaga2018} reward models,
and to update latent feature posteriors in a probabilistic matrix factorization model~\citep{ic-Kawale2015}.
\citet{ip-Gopalan2014} showed that utilizing SMC to compute posterior distributions that lack an explicit closed-form
is a theoretically grounded approach for certain online learning problems,
such as bandit subset arm selection or job scheduling tasks.

These efforts provide evidence that SMC can be successfully combined with Thompson sampling,
yet are different in scope from our work.
The SMC-based MAB framework we present generalizes existing Bayesian MAB policies beyond their original setting.
Contrary to existing MAB solutions, the SMC-based bandit policies we propose
($i$) are not restricted to specific reward functions,
but accommodate nonlinear and non-Gaussian rewards,
($ii$) address non-stationary bandit environments, and
($iii$) are readily applicable to state-of-the-art Bayesian MAB algorithms
---Thompson sampling and Bayes-UCB policies--- in a modular fashion.

\subsection{Sequential Monte Carlo}
\label{ssec:smc}

Monte Carlo (MC) methods are a family of numerical techniques based on repeated random sampling,
which have been shown to be flexible enough for both numerical integration and drawing samples from complex probability distributions of interest~\citep{b-Liu2001}.

With importance sampling (IS), one estimates properties of a distribution when obtaining samples from such distribution is difficult.
The basic idea of IS is to draw, from an alternative distribution,
samples that are subsequently weighted to guarantee estimation accuracy (and often reduced variance).
These methods are used both to approximate posterior densities,
and to compute expectations in probabilistic models, \ie
\begin{equation}
\bar{f}=\int f(\varphi) p(\varphi) \mathrm{d}\varphi \;,
\end{equation}
when these are too complex to treat analytically.
IS relies on a proposal distribution $q(\cdot)$,
from which one draws $M$ samples $\varphi^{(m)} \sim q(\varphi), \; m=1, \cdots , M$,
weighted according to
\begin{equation}
\widetilde{w}^{(m)}=\frac{p(\varphi^{(m)})}{q(\varphi^{(m)})} \;, \quad \text{with} \quad w^{(m)}=\frac{\widetilde{w}^{(m)}}{\sum_{m=1}^M\widetilde{w}^{(m)}} \; .
\end{equation}

If the support of $q(\cdot)$ includes the support of the distribution of interest $p(\cdot)$, one computes the IS estimator of a test function based on the normalized weights $w^{(m)}$,
\begin{equation}
\bar{f}_M=\sum_{m=1}^M w^{(m)} f\left(\varphi^{(m)}\right) \; ,
\end{equation}
with convergence guarantees under weak assumptions~\citep{b-Liu2001}.

IS can also be interpreted as a sampling method where the true posterior distribution is approximated by a random measure, \ie
\begin{equation}
p(\varphi) \approx p_M(\varphi) = \sum_{m=1}^M w^{(m)} \delta\left(\varphi^{(m)}-\varphi\right) \;,
\end{equation}
leading to estimates that integrate the test function with respect to such measure,
\begin{equation}
\bar{f}_M=\int f(\varphi) p_M(\varphi) \mathrm{d}\varphi =  \sum_{m=1}^M f\left(\varphi^{(m)}\right) w^{(m)} \; .
\end{equation}

The sequential counterpart of IS,
also known as sequential Monte Carlo (SMC)~\citep{b-Doucet2001}
or particle filtering (PF)~\citep{ib-Djuric2010},
provides a convenient solution to computing approximations to posterior distributions
with sequential or recursive formulations.
In SMC, one considers a proposal distribution that factorizes
---often, but not necessarily--- over time, \ie
\begin{equation}
q(\varphi_{0:t})=q(\varphi_t|\varphi_{1:t-1}) q(\varphi_{1:t-1})=\prod_{\tau=1}^{t} q(\varphi_{\tau}|\varphi_{1:\tau-1}) q(\varphi_0) \; ,
\end{equation}
which helps in matching the sequential form of the probabilistic model of interest $p(\varphi_t|\varphi_{1:t-1})$,
to enable a recursive evaluation of the importance sampling weights
\begin{equation}
w_t^{(m)} \propto \frac{p(\varphi_{t}|\varphi_{1:t-1})}{q(\varphi_{t}|\varphi_{1:t-1})} w_{t-1}^{(m)} \; .
\end{equation}

One problem with following the above weight update scheme is that,
as time evolves, the distribution of the importance weights becomes more and more skewed,
resulting in few (or just one) non-zero weights.

To overcome this degeneracy, an additional selection step, known as resampling \citep{j-Li2015}, is added.
In its most basic setting,
one replaces the weighted empirical distribution with an equally weighted random measure at every time instant,
where the number of offspring for each sample is proportional to its weight.
This is known as Sequential Importance Resampling (SIR)~\citep{j-Gordon1993}.

SIR and its many variants ~\citep{b-Doucet2001,j-Arulampalam2002}
have been shown to be of great flexibility and value
in many science and engineering problems~\citep{b-Ristic2004,j-Leeuwen2009,j-Ionides2006,j-Creal2012},
where data are acquired sequentially in time.
In these circumstances,
one needs to infer all the unknown quantities in an online fashion,
where often, the underlying parameters evolve over time.

SMC provides a flexible and useful framework
for these problems with probabilistic models and lax assumptions:
\ie when nonlinear observation functions, non-Gaussian noise processes and uncertainty over model parameters must be accommodated.
Here, we leverage SMC for flexible approximations to posterior of interest in non-stationary and nonlinear MAB problems.

\section{SMC for multi-armed bandits}
\label{sec:smc_mab}
We use sequential Monte Carlo to compute posteriors and sufficient statistics of interest for a rich-class of MABs:
non-stationary bandits (modeled via linear dynamical systems)
with complex (stateless and context-dependent) nonlinear reward functions,
subject to non-Gaussian stochastic innovations.

We model non-stationary, stochastic MABs in a state-space framework,
where for a given reward distribution $p_{a}(Y|x,\theta)$,
and parameters that evolve in-time via a transition distribution $p(\theta_{t}|\theta_{t-1})$,
we write
\begin{equation}
\begin{cases}
\theta_{t}^* \sim p(\theta_{t}^*|\theta_{t-1}^*) \; ,\\
Y_{t}\sim p_{a_t}(Y|x_t,\theta_{t}^*) \; ,
\end{cases}
\quad t=1, \cdots, T,
\label{eq:dynamic_mab}
\end{equation}
where we explicitly indicate with $\theta_t^*$ the true yet unknown parameters of the non-stationary multi-armed bandit.

Within this bandit framework, a Bayesian policy must characterize
the posterior of the unknown parameters $p(\theta_t|\HH_{1:t})$ as in Equation~\eqref{eq:mab_param_posterior},
in which the time-varying dynamics of the underlying transition distribution are incorporated.

The posterior of interest given observed reward $y_t$
can be written as
\begin{equation}
p(\theta_{t}|\HH_{1:t}) \propto p_{a_t}(y_t|x_t, \theta_{t}) p(\theta_{t} |\HH_{1:t-1}) \; ,
\label{eq:dynamic_posterior}
\end{equation}
where $p(\theta_t| \HH_{1:t-1}) = \int_{\theta_{t-1}} p(\theta_{t} | \theta_{t-1}) p(\theta_{t-1}|\HH_{1:t-1}) \dd{\theta_{t-1}}$.
Recall that the parameter predictive distribution $p(\theta_t| \HH_{1:t-1})$ and parameter posterior $p(\theta_{t}|\HH_{1:t})$ in Equation~\eqref{eq:dynamic_posterior} 
have analytical, closed-form recursive solutions only for limited cases.

We adhere to the standard MAB formulation,
in that each arm of the bandit is described by its own idiosyncratic parameters
(no information is shared across arms);
\ie $p_{a}(Y| x_t,\theta_{t}^*)=p_{a}(Y|x_t,\theta_{t,a}^*)$,
yet we allow for such parameters to evolve independently per-arm in time:
$p(\theta_{t}^*|\theta_{t-1}^*)=\prod_{a=1}^{|\A|} p(\theta_{t,a}^*|\theta_{t-1,a}^*)$.
Therefore, the posterior of interest factorizes across arms
\begin{equation}
	p(\theta_{t}|\HH_{1:t}) = \prod_{a=1}^{|\A|} p(\theta_{t,a}|\HH_{1:t})\; .
	\label{eq:dynamic_posterior_factorized}
\end{equation}

This standard MAB formulation with independent, per-arm parameter dynamics
enables Bayesian MAB policies to approximate each per-arm parameter posterior separately.
Given observed reward $y_t$ for played MAB arm $a_t$ at time instant $t$,
only the parameter posterior of the played arm is updated with this new observation,
\ie 
\begin{equation}
	p(\theta_{t,a_t}|\HH_{1:t}) \propto p_{a_t}(y_t|x_t, \theta_{t,a_t}) p(\theta_{t,a_t} | \theta_{t-1,a_t}) p(\theta_{t-1,a_t}|\HH_{1:t-1}) \; ;
	\label{eq:dynamic_posterior_factorized_aplayed_updated}
\end{equation}
while parameter posteriors of the non-played arms are only updated according to the latent parameter dynamics, \ie
\begin{equation}
	p(\theta_{t,a}|\HH_{1:t}) \propto p(\theta_{t,a} | \theta_{t-1,a}) p(\theta_{t-1,a}|\HH_{1:t-1}) \;, \forall a^\prime \neq a_t \;.
	\label{eq:dynamic_posterior_factorized_updated}
\end{equation}

We here combine SMC with Bayesian bandit policies ---Thompson sampling and Bayes-UCB, specifically---
for non-stationary bandits as modeled in Equation~\eqref{eq:dynamic_mab}.
The challenge is on computing the posteriors in Equations~\eqref{eq:dynamic_posterior}, \eqref{eq:dynamic_posterior_factorized_aplayed_updated} and \eqref{eq:dynamic_posterior_factorized_updated} for a variety of MAB models,
for which SMC enables us to accommodate 
($i$) any likelihood function that is computable up to a proportionality constant, and
($ii$) any time-varying model described by a transition density from which we can draw samples.

We use SMC to compute per-arm parameter posteriors at each bandit round,
\ie we approximate per-arm filtering densities $p(\theta_{t,a}|\HH_{1:t})$ with SMC-based random measures $p_M(\theta_{t,a}|\HH_{1:t})$, $\forall a$,
for which there are strong theoretical convergence guarantees~\citep{j-Crisan2002,j-Chopin2004}.

The dimensionality of this estimation problem depends on the size of per-arm parameters,
and not on the number of bandit arms $|\A|$.
Consequently, there will be no particle degeneracy due to increased number of arms.

We present in Section~\ref{ssec:sir-policies} and Algorithm~\ref{alg:sir-mab}
the SMC-based Bayesian MAB framework we devise for nonlinear and non-stationary bandits.
We describe in Section~\ref{ssec:linear_mixing_dynamics}
how to draw samples from the transition densities,
when modeling bandit non-stationarity via the general linear model;
and in Section~\ref{ssec:mab_reward_models},
we present examples of
non-Gaussian and nonlinear (continuous and discrete)
reward functions of interest in practice.
Throughout, we avoid assumptions on model parameter knowledge and resort to their Bayesian marginalization.

\subsection{SMC-based Bayesian MAB policies}
\label{ssec:sir-policies}

We combine SMC with both Thompson sampling and Bayes-UCB policies, 
by sequentially updating, at each bandit interaction $t$,
a SMC-based random measure to approximate the time-varying posterior of interest,
\begin{equation}
p(\theta_{t,a}|\HH_{1:t})\approx p_M(\theta_{t,a}|\HH_{1:t})=\sum_{m=1}^M w_{t,a}^{(m)} \delta\left(\theta_{t,a}^{(m)}-\theta_{t,a}\right) \;.
\end{equation}
Knowledge of $p_M(\theta_{t,a}|\HH_{1:t})$ enables computation
of any per-arm reward statistic Bayesian MAB policies require.

We present Algorithm~\ref{alg:sir-mab}
with the sequential Importance Resampling (SIR) method\footnote{
	We acknowledge that any of the methodological SMC advancements that improve and extend SIR,
	\eg advanced SMC algorithms~\citep{ip-Merwe2001, j-Andrieu2010}
	or alternative resampling mechanisms~\citep{j-Li2015},
	are readily applicable to the proposed SMC-based bandit framework,
	and are likely to have a positive impact on the corresponding SMC-based MAB policies' performance.
} 
as introduced by \citet{j-Gordon1993}, 
where:
\begin{itemize}
	\item The SMC proposal distribution $q(\cdot)$ at each bandit interaction $t$ obeys the assumed parameter dynamics:
		$\theta_{t,a}^{(m)} \sim p(\theta_{t,a}|\theta_{t-1,a}), \ \forall m$ ---Step (9.b) in Algorithm~\ref{alg:sir-mab};
	\item SMC weights are updated based on the likelihood of the observed rewards:
		$w_{t,a}^{(m)} \propto p_a(y_t|x_t,\theta_{t,a}^{(m)})$ ---Step (9.c) in Algorithm~\ref{alg:sir-mab}; and
	\item The SMC random measure is resampled at every time instant ---Step (9.a).
\end{itemize}

Independently of which SMC technique is used to compute the posterior random measure $p_M(\theta_{t,a}|\HH_{1:t})$,
the fundamental operation in the proposed SMC-based MAB Algorithm~\ref{alg:sir-mab} 
is to sequentially update the random measure $p_M(\theta_{t,a}|\HH_{1:t})$
to approximate the true per-arm posterior $p(\theta_{t,a}|\HH_{1:t})$
over bandit interactions.

This SMC-based random measure is key,
along with transition density $p(\theta_{t,a}|\theta_{t-1,a})$,
to sequentially propagate parameter posteriors per-arm,
and to estimate their sufficient statistics for any Bayesian bandit policy.
More precisely:
\begin{itemize}
	\item In Step 5 of Algorithm~\ref{alg:sir-mab},
	we estimate the predictive posterior of per-arm parameters,
	as a mixture of the transition densities conditioned on previous samples from $p_M(\theta_{t,a}|\HH_{1:t})$, 
	\begin{equation}
		p_M(\theta_{t+1,a}|\HH_{1:t}) = \sum_{m_{t,a}=1}^{M} w_{t,a}^{(m_{t,a})} p(\theta_{t+1,a}|\theta_{t,a}^{(m_{t,a})}) \; , \; m_{t,a}=1,\cdots, M, \; \forall a\in \A \; .
		\label{eq:smc_parameter_predictive_posterior}
	\end{equation}
	
	\item In Step 9 of Algorithm~\ref{alg:sir-mab},
	we propagate forward the sequential random measure $p_M(\theta_{t,a}|\HH_{1:t})$
	by drawing new samples from the transition density, conditioned on \textit{resampled} particles, \ie
	\begin{equation}
		\theta_{t+1,a}^{(m_{t+1,a})} \sim p(\theta_{t+1,a}|\overline{\theta}_{t,a}^{(m_{t+1,a})}) \; , \; m_{t+1,a}=1,\cdots, M, \; \forall a\in \A \; .
	\end{equation}
\end{itemize}
In both cases, one draws with replacement according to the importance weights in $p_M(\theta_{t,a}|\HH_{1:t})$,
\ie from a categorical distribution with per-sample probabilities $w_{t,a}^{(m)}$: $m_{t,a}^\prime \sim \Cat{w_{t,a}^{(m)}}$.

We now describe in detail how to use the SMC-based posterior random measure $p_M(\theta_{t+1,a}|\HH_{1:t})$ for both Thompson sampling and Bayes-UCB policies:
\ie which are the specific instructions to execute in steps 5 and 7 of Algorithm~\ref{alg:sir-mab}.
\begin{itemize}
	\item \textbf{SMC-based Thompson Sampling:}
	TS operates by drawing a sample parameter $\theta_{t+1}^{(s)}$ from its updated posterior $p(\theta_{t+1}|\HH_{1:t})$,
	and picking the optimal arm for such sample, \ie
	\begin{equation}
		a_{t+1}=\argmax_{a^\prime \in \A} \mu_{t+1,a^\prime}\left(x_{t+1},\theta_{t+1,a^\prime}^{(s)}\right) \; .
		\label{eq:mc_expected_reward}
	\end{equation}
	We use the SMC random measure $p_M(\theta_t|\HH_{1:t})$,
	and propagate it using the transition density $p(\theta_{t+1,a}|\theta_{t,a})$,
	to draw samples from the parameter posterior predictive distribution:
	\ie $\theta_{t+1}^{(s)}\sim p_M(\theta_{t+1}|\HH_{1:t})$
	in Equation~\eqref{eq:smc_parameter_predictive_posterior}.
	This SMC-based random measure provides an accurate approximation to the true posterior density with high probability.

	\vspace*{1ex}	
	\item \textbf{SMC-based Bayes-UCB:}
	We extend Bayes-UCB to reward models where the quantile functions are not analytically tractable,
	by leveraging the SMC-based parameter predictive posterior random measure  $p_M(\theta_{t+1}|\HH_{1:t})$.

	We compute the quantile function of interest by first 
	evaluating the expected reward at each round $t$ based on the available posterior samples,
	\ie $\mu_{t+1,a}^{(m)}\left(x_{t+1},\theta_{t+1,a}^{(m)}\right)$, $m=1,\cdots,M$;
	and compute
	$
	\Prob{\mu_{t+1,a}>q_{t+1,a}(\alpha_{t+1})} = \alpha_{t+1}
	$
	via
	\begin{equation}
		q_{t+1,a}(\alpha_{t+1}):=\max \left\{\mu \; \left|\sum_{m|\mu_{t+1,a}^m>\mu} w_{t,a}^m\ge\alpha_{t+1} \right. \right\} \; .
		\label{eq:mc_quantile_value}
	\end{equation}
	The convergence of quantile estimators generated by SMC methods has been explicitly proved in~\citep{j-Maiz2012}.
\end{itemize}

Random measure $p_M(\theta_{t+1,a}|\HH_{1:t})$ in Equation~\eqref{eq:smc_parameter_predictive_posterior} 
enables computation of the statistics Bayesian MAB policies require,
extending their applicability from stationary to time-evolving bandits.
The exposition that follows addresses dynamic bandits,
and we illustrate how to process classic, stationary bandits within the proposed framework in Appendix~\ref{asec:static_bandits}.

\begin{algorithm}
	\caption{SMC-based Bayesian MAB policies}
	\label{alg:sir-mab}
	\begin{algorithmic}[1]
		\REQUIRE $\A$, $p(\theta_a)$, $p(\theta_{t,a}|\theta_{t-1,a})$, $p_a(Y|x,\theta)$, $\forall a \in \A$.
		\REQUIRE Number of SMC samples $M$ (for UCB we also require $\alpha_t$)
		\STATE Draw initial samples from the parameter prior
			\vspace*{-1ex}
		\begin{equation}
			\overline{\theta}_{0,a}^{(m_{0,a})} \sim p(\theta_a), \quad \text{ and } \quad w_{0,a}^{^{(m_{0,a})}}=\frac{1}{M} \;, \; m_{0,a}=1,\cdots, M, \; \forall a \in \A \;.
			\nonumber
			\vspace*{-4ex}
		\end{equation}
		\FOR{$t=0, \cdots, T-1$}
		\STATE Receive context $x_{t+1}$
		\FOR{$a \in \A$}
		\STATE Estimate sufficient statistics of the MAB policy for all arms, \\
		given $\{w_{t,a}^{(m_{t,a})} \}$ and $\{\theta_{t,a}^{(m_{t,a})}\}$, $\forall m_{t,a}$, $\forall a\in\A$.\\
		
		\vspace*{1ex}
		\quad For \textit{Thompson sampling:}\\
		\qquad Draw a sample $s \sim \Cat{w_{t,a}^{(m_{t,a})}}$, \\
		\qquad Propagate the sample parameter $\theta_{t+1,a}^{(s)}\sim p\left(\theta_{t+1,a}|\theta_{t,a}^{(s)}\right)$, \\
		\qquad Set $\mu_{t+1,a}\left(x_{t+1}, \theta_{t+1,a}^{(s)}\right)=\eValue{}{Y|a,x_{t+1}, \theta_{t+1,a}^{(s)}}$ .\\
		
		\vspace*{1ex}
		\quad For\textit{Bayes-UCB:}\\
		\qquad Draw $M$ candidate samples $m_{a}^\prime \sim \Cat{w_{t,a}^{(m_{t,a})}}$,\\
		\qquad Propagate sample parameters $\theta_{t+1,a}^{(m_{a}^\prime)} \sim p\left(\theta_{t+1,a}|\theta_{t,a}^{(m_{a}^\prime)}\right)$, \\
		\qquad Set $\mu_{t+1,a}\left(x_{t+1}, \theta_{t+1,a}^{(m_{a}^\prime)}\right)=\eValue{}{Y|a,x_{t+1}, \theta_{t+1,a}^{(m_{a}^\prime)}}$,\\
		\qquad Estimate quantile $q_{t+1,a}(\alpha_{t+1})$ as in Equation~\eqref{eq:mc_quantile_value}.
		
		\vspace*{1ex}
		\ENDFOR
		\STATE Decide next action $a_{t+1}$ to play\\
		\vspace*{1ex}
		\quad For \textit{Thompson sampling:} \hspace*{0.6cm} $a_{t+1}=\argmax_{a^\prime \in \A} \mu_{t+1,a^\prime}\left(x_{t+1}, \theta_{t+1,a^\prime}^{(s)}\right)$ \\
		\vspace*{1ex}
		\quad For\textit{Bayes-UCB:} \hspace*{1.8cm} $a_{t+1}=\argmax_{a^\prime \in \A}q_{t+1,a^\prime}(\alpha_{t+1})$
		
		\vspace*{1ex}
		\STATE Observe reward $y_{t+1}$ for played arm
		
		\vspace*{1ex}
		\STATE Update the posterior SMC random measure $p_M(\theta_{t,a}|\HH_{1:t})$ for all arms\\
		
		\begin{enumerate}[(a)]
			\vspace*{-1ex}
			\item Resample $m_{t+1,a}=1,\cdots, M$ parameters $\overline{\theta}_{t,a}^{(m_{t+1,a})}=\theta_{t,a}^{(m_{t,a}^\prime)}$ per arm $a\in \A$,
			where $m_{t,a}^\prime$ is drawn with replacement according to the importance weights $w_{t,a}^{(m_{t,a})}$.

			\vspace*{-1ex}
			\item Propagate resampled parameters by drawing from the transition density
			\vspace*{-1ex}
			\begin{equation}		
				\theta_{t+1,a}^{(m_{t+1,a})} \sim p\left(\theta_{t+1,a} \middle| \overline{\theta}_{t,a}^{(m_{t+1,a})}\right) \; , \; m_{t+1,a}=1,\cdots, M, \; \forall a \in \A \; .
				\label{eq:sir-mab-propagate}
			\vspace*{-2ex}
			\end{equation}
			
			\vspace*{-1ex}
			\item Weight samples of the played arm $a_{t+1}$ based on the likelihood of observed $y_{t+1}$
			\vspace*{-1ex}
			\begin{equation}
				\widetilde{w}_{t+1,a_{t+1}}^{\left(m_{t+1,a_{t+1}}\right)} \propto p_{a_{t+1}}\left(y_{t+1} \middle|  x_{t+1},\theta_{t+1,a_{t+1}}^{\left(m_{t+1,a_{t+1}}\right)}\right) \; ,
				\label{eq:sir-mab-weights}
			\vspace*{-2ex}
			\end{equation}
			and normalize the weights
			\vspace*{-1ex}
			\begin{equation}
				w_{t+1,a_{t+1}}^{\left(m_{t+1,a_{t+1}}\right)}=\frac{\widetilde{w}_{t+1,a_{t+1}}^{\left(m_{t+1,a_{t+1}}\right)}}{\sum_{m_{t+1,a_{t+1}}=1}^M\widetilde{w}_{t+1,a_{t+1}}^{\left(m_{t+1,a_{t+1}}\right)}} \; , \; m_{t+1,a}=1,\cdots, M.
				\label{eq:sir-mab-weights-norm}
			\vspace*{-2ex}
			\end{equation}
		\end{enumerate}
		\vspace*{-2ex}
		\ENDFOR
	\end{algorithmic}
\end{algorithm}

\subsection{Non-stationary MABs}
\label{ssec:linear_mixing_dynamics}

The dynamic linear model is a flexible and widely used framework to characterize time-evolving systems~\citep{b-Whittle1951, b-Box1976, b-Brockwell1991, b-Durbin2001, b-Shumway2010, b-Durbin2012}.
Here, we model the latent parameters of the bandit $\theta \in \Real^{d_{\Theta}}$ to evolve over time according to
\begin{equation}
\theta_{t,a}=L_a \theta_{t-1,a}+\epsilon_a \;, \qquad \epsilon_a\sim\N{\epsilon_a|0, \Sigma_a} \; ,
\label{eq:linear_mixing_dynamics}
\end{equation}
with parameters $L_a \in \Real^{d_{\Theta_a} \times d_{\Theta_a}}$ and $\Sigma_a \in \Real^{d_{\Theta_a} \times d_{\Theta_a}}$
---recall that we specify distinct transition densities per-arm.

With \emph{known parameters} $L_a$ and $\Sigma_a$, the transition distribution $p(\theta_{t,a}|\theta_{t-1,a})$
is Gaussian with closed-form updates,
\ie $\theta_{t,a}\sim \N{\theta_{t,a}|L_a \theta_{t-1,a}, \Sigma_a} $.

For the more interesting case of \emph{unknown parameters},
we marginalize parameters $L_a$ and $\Sigma_a$ of the transition distributions
utilized by the proposed SMC-based Bayesian policies, \ie
we Rao-Blackwellize\footnote{
	Rao-Blackwellization is known to help reduce the degeneracy and variance of SMC estimates~\citep{ip-Doucet2000,ib-Djuric2010}.
} them.

The marginalized transition density is a multivariate t-distribution~\citep{j-Geisser1963,j-Tiao1964,j-Geisser1965,j-Urteaga2016,j-Urteaga2016a}:
\begin{align}
\theta_{t,a} \sim \T{\theta_{t,a}|\nu_{t,a}, m_{t,a}, R_{t,a}} \;, \text{ with } \quad
\begin{cases}
\nu_{t,a}=\nu_{0,a}+t-d \; ,\\
m_{t,a}=L_{t-1,a} \theta_{t-1,a} \; , \\
R_{t,a} = \frac{V_{t-1,a}}{\nu_{t,a}\left(1-\theta_{t-1,a}^\top(U_{t,a} U_{t,a}^\top)^{-1}\theta_{t-1,a}\right)} \; ,\\
\end{cases} 
\label{eqn:dynamics_rb} \\
\nonumber \\
\text{where } \begin{cases}
\Theta_{t_0:t_1,a}=[\theta_{t_0,a} \theta_{t_0+1,a} \cdots \theta_{t_1-1,a} \theta_{t_1,a}] \in \Real^{d\times (t_1-t_0)} \; , \\
B_{t-1,a} = \left(\Theta_{0:t-2,a}\Theta_{0:t-2,a}^\top + B_{0,a}^{-1} \right)^{-1} \; ,\\
L_{t-1,a} = \left(\Theta_{1:t-1,a}\Theta_{0:t-2,a}^\top + L_{0,a}B_{0,a}^{-1}\right) B_{t-1,a} \; ,\\
V_{t-1,a}= \left(\Theta_{1:t-1,a}-L_{t-1,a} \Theta_{0:t-2,a}\right)\left(\Theta_{1:t-1,a}-L_{t-1,a} \Theta_{0:t-2,a}\right)^\top \\
\qquad \qquad + \left(L_{t-1,a}-L_{0,a}\right) B_{0,a}^{-1} \left(L_{t-1,a}-L_{0,a}\right)^\top + V_{0,a} \; ,\\
U_{t,a} U_{t,a}^\top = \left(\theta_{t-1,a}\theta_{t-1,a}^\top+B_{t-1,a}^{-1}\right) \; .\\
\end{cases}
\label{eqn:dynamics_rb_aux}
\end{align}
Each distribution above holds separately for each arm $a$, and subscript $_{a,0}$ indicates assumed prior parameters for arm $a$.

These transition distributions are used when propagating per-arm parameter densities in Steps 5 and 9 of Algorithm~\ref{alg:sir-mab}.
They are fundamental for the accuracy of the sequential, random measure-based approximation to the posterior,
and the downstream performance of the proposed SMC-based MAB policies.

Caution must be exercised when using SMC to approximate the dynamic bandit model's posteriors.
Notably, the impact of non-Markovian transition distributions in SMC performance must be taken into consideration:
the sufficient statistics in Equations~\eqref{eqn:dynamics_rb}-\eqref{eqn:dynamics_rb_aux} depend on the full history of the model dynamics.
Here, we use general linear models,
for which it can be shown that, if stationarity conditions are met,
the autocovariance function decays quickly,
\ie the dependence of general linear models on past samples decays exponentially~\citep{j-Urteaga2016,j-Urteaga2016a}.

When exponential forgetting holds in the latent space
---\ie the dependence on past samples decays exponentially, and is negligible after a certain lag---
one can establish uniform-in-time convergence of SMC methods for functions that depend only on recent states, see~\citep{j-Kantas2015} and references therein.

More broadly, one can establish uniform-in-time convergence for path functionals that depend only on recent states,
as the Monte Carlo error of $p_M(\theta_{t-\tau:t}|\HH_{1:t})$ with respect to $p(\theta_{t-\tau:t}|\HH_{1:t})$ is uniformly bounded over time.
This quick forgetting property is fundamental
for the successful performance of SMC methods for inference of linear dynamical states in practice~\citep{j-Urteaga2017b,j-Urteaga2016,j-Urteaga2016a}.

Nevertheless, we acknowledge that any improved SMC solution that mitigates path-degeneracy issues can only be beneficial for the performance of the proposed SMC-based policies.

\subsection{MAB reward models}
\label{ssec:mab_reward_models}

Algorithm~\ref{alg:sir-mab} is described in terms of a generic reward likelihood function $p_a(Y|x_t,\theta_{t,a})$
that must be computable up to a proportionality constant.
We now introduce reward functions that are applicable in many MAB use-cases,
where the time subscript $_t$ has been suppressed for clarity of presentation,
and subscript $_0$ indicates assumed prior parameters.

\subsubsection{Contextual, categorical rewards}
\label{sssec:categorical_softmax_rewards}

For MAB problems where observed rewards are discrete,
\ie $Y=c$ for $c\in\{1,\cdots,C\}$,
and contextual information is available,
the softmax function is a natural reward density model.
In general, categorical variables assign probabilities to an unordered set of outcomes ---not necessarily numeric.
In this work,
we refer to categorical rewards where,
for each categorical outcome $c\in\Natural$,
there is a numeric reward $y=c$ associated with it.

Given a $d$-dimensional context vector $x\in\Real^{d}$,
and per-arm parameters $\theta_{a,c} \in \Real^{d}$ for each category $c\in\{1,\cdots,C\}$,
the contextual softmax reward model is
\begin{equation}
p_a(Y=c|x,\theta_a)=\frac{e^{(x^\top\theta_{a,c})}}{\sum_{c'=1}^C e^{(x^\top\theta_{a,c'})} } \; ,
\label{eq:softmax_rewards}
\end{equation}
where we denote with $\theta_a=\{\theta_{a,1}, \cdots, \theta_{a,C}\}$ the set of per-category parameters $\theta_{a,c}$ for arm $a$.
For this reward distribution,
the posterior of the parameters can not be computed in closed form,
and neither, the quantile function of the expected rewards $\mu_{t,a}=y_t\cdot(x_t^\top\theta_{t,a})$.

When returns are binary, \ie $Y=\{0,1\}$ (success or failure of an action),
but dependent on a $d$-dimensional context vector $x\in\Real^{d}$,
the softmax function reduces to the logistic reward model
\begin{equation}
p_a(Y|x,\theta)=\frac{e^{Y\cdot(x^\top\theta_a) }}{1+e^{(x^\top\theta_a)}} \; ,
\label{eq:logistic_rewards}
\end{equation}
with per-arm parameters $\theta_a \in \Real^{d}$ of same dimensionality $d$ as the context $x$.

The theoretical study of UCB and TS-based algorithms for logistic rewards is an active research area~\citep{ip-Dong2019,ip-Faury2020},
which we here extend to the discrete-categorical setting.
We accommodate discrete-categorical MAB problems by implementing Algorithm~\ref{alg:sir-mab}
with likelihoods as in Equations~\eqref{eq:softmax_rewards} or~\eqref{eq:logistic_rewards}.
Namely, we compute $p_M(\theta_{t,a}|\HH_{1:t})$ 
for both stationary and non-stationary discrete-categorical bandits,
by updating the weights of the posterior SMC random measure in Step 9-c.
To the best of our knowledge,
no existing work addresses non-stationary, discrete-categorical bandits.

\subsubsection{Contextual, linear Gaussian rewards}
\label{sssec:linear_gaussian_rewards}

For bandits with continuous rewards, Gaussian distributions are typically used,
where contextual dependencies can easily be included.
The contextual linear Gaussian reward model is well studied in the bandit literature~\citep{ic-Abbasi-Yadkori2011,ip-Chu2011,ip-Agrawal2013a},
where the expected reward of each arm is modeled as a linear combination of a $d$-dimensional context vector $x\in\Real^{d}$,
and the idiosyncratic parameters of the arm $w_a\in\Real^{d}$; \ie
\begin{equation}
\begin{split}
p_a(Y|x,\theta)&=\N{Y \middle|x^\top w_a, \sigma_a^2} =\frac{1}{\sqrt{2\pi\sigma_a^2}}e^{-\frac{(Y-x^\top w_a)^2}{2\sigma_a^2}} \; .
\end{split}
\end{equation}
We denote with $\theta\equiv\{w, \sigma\}$ the set of all parameters of the reward distribution,
and consider the normal inverse-gamma conjugate prior distribution for these, 
\begin{equation}
\begin{split}
p(w_a, \sigma_a^2|u_{0,a}, V_{0,a}, \alpha_{0,a}, \beta_{0,a})
& = \N{w_a|u_{0,a}, \sigma_a^2 V_{0,a}} \cdot \IG{\sigma_a^2|\alpha_{0,a}, \beta_{0,a}} \; .\\
\end{split}
\end{equation}

After observing actions $a_{1:t}$ and rewards $y_{1:t}$, the parameter posterior for each arm 
\begin{equation}
\begin{split}
p(w_a, \sigma_a^2|a_{1:t},y_{1:t},u_{0,a}, V_{0,a},\alpha_{0,a}, \beta_{0,a}) &= p\left(w_a, \sigma_a^2|u_{t,a}, V_{t,a},\alpha_{t,a}, \beta_{t,a}\right) \\
\end{split}
\end{equation}
follows an updated normal inverse-gamma distribution with sequentially updated hyperparameters
\begin{equation}
\begin{cases}
V_{t,a}^{-1} = V_{t-1,a}^{-1} + x_t x_t^\top \cdot \mathds{1}[a_t=a] \; ,\\
u_{t,a}= V_{t,a} \left( V_{t-1,a}^{-1} u_{t-1,a} + x_t y_{t}\cdot \mathds{1}[a_t=a]\right) \; ,\\
\alpha_{t,a}=\alpha_{t-1,a} + \frac{\mathds{1}[a_t=a]}{2} \; ,\\
\beta_{t,a}=\beta_{t-1,a} + \frac{\mathds{1}[a_t=a](y_{t_a}-x_t^\top u_{t-1,a})^2}{2\left(1+x_t^\top V_{t-1,a} x_t\right)} \; ,
\end{cases}
\end{equation}
or, alternatively, batch updates of the form
\begin{equation}
\begin{cases}
V_{t,a}^{-1}= V_{0,a}^{-1}+x_{{1:t}|t_a} x_{{1:t}|t_a}^\top \; ,\\
u_{t,a}=V_{t,a}\left(V_{0,a}^{-1}u_{0,a}+x_{{1:t}|t_a} y_{{1:t}|t_a}\right) \; ,\\
\alpha_{t,a}=\alpha_{0,a} + \frac{|t_a|}{2} \; ,\\
\beta_{t,a}=\beta_{0,a} + \frac{\left(y_{{1:t}|t_a}^\top y_{{1:t}|t_a} + u_{0,a}^\top V_{0,a}^{-1}u_{0,a} - u_{t,a}^\top V_{t,a}^{-1}u_{t,a} \right)}{2} \; ,
\end{cases}
\end{equation}
where $t_a=\{t|a_t=a\}$ indicates the set of time instances when arm $a$ is played.

With these, we can compute the Bayesian expected reward of each arm,
\begin{equation}
p(\mu_{a}|x, \sigma_a^2, u_{t,a}, V_{t,a}) = \N{\mu_{a} \middle|x^\top u_{t,a}, \; \sigma_a^2 \cdot x^\top V_{t,a} x} \; ,
\label{eq:gaussian_posterior_mean}
\end{equation}
and the quantile function for such distribution
\begin{equation}
q_{t+1,a}(\alpha_{t+1})=Q\left(1-\alpha_{t+1}, \N{\mu_{a} \middle|x^\top u_{t,a}, \; \sigma_a^2 \cdot x^\top V_{t,a} x}\right) \;.
\label{eq:gaussian_posterior_quantile}
\end{equation}

The reward variance $\sigma^2_a$ is unknown in practice,
so we marginalize it and obtain
\begin{equation}
\begin{split}
p(\mu_{a}|x, u_{t,a}, V_{t,a}) = \T{\mu_{a} \middle|2\alpha_{t,a}, x^\top u_{t,a}, \; \frac{\beta_{t,a}}{\alpha_{t,a}} \cdot x^\top V_{t,a} x} \;, 
\end{split}
\label{eq:t_posterior_mean}
\end{equation}
which leads to quantile function computations based on a Student's t-distribution
\begin{equation}
q_{t+1,a}(\alpha_{t+1})=Q\left(1-\alpha_{t+1}, \T{\mu_{a} \middle| 2\alpha_{t,a}, x^\top u_{t,a}, \; \frac{\beta_{t,a}}{\alpha_{t,a}} \cdot x^\top V_{t,a} x}\right) \;.
\label{eq:t_posterior_quantile}
\end{equation}

Equations~\eqref{eq:gaussian_posterior_mean} and~\eqref{eq:gaussian_posterior_quantile}
are needed in Step 5 when implementing TS or Bayes-UCB policies for the known $\sigma_a^2$ case;
while Equations~\eqref{eq:t_posterior_mean} and~\eqref{eq:t_posterior_quantile} are used for the unknown $\sigma_a^2$ case.
Note that one can use the above results for Gaussian bandits with no context, by replacing $x=I$ and obtaining $\mu_{a}=u_{t,a}$.

When these equations are combined
with the Rao-Blackwellized transition densities derived for the dynamic model in Section~\ref{ssec:linear_mixing_dynamics},
the proposed SMC-based MAB policies can be applied to non-stationary, linear Gaussian bandit problems with minimal assumptions:
\ie only the functional form of the transition and reward functions is known.

There are no competing MAB algorithms
for non-stationary bandit problems where no parameter knowledge is assumed.

\clearpage
\section{Evaluation}
\label{sec:evaluation}
We empirically evaluate the proposed SMC-based Bayesian MAB framework
in non-stationary bandit scenarios
with continuous, binary and discrete-categorical reward distributions.

Results in Appendix~\ref{assec:static_bandits_experiments_analytical}
validate the performance of SMC-based policies in stationary bandits.
We compare their performance to solutions based on analytically attainable posteriors
with Bernoulli and contextual linear Gaussian reward functions~\citep{ip-Kaufmann2012,ip-Garivier2011a,ic-Korda2013,ip-Agrawal2013a},
as well as for context-dependent binary rewards modeled with the logistic reward function~\cite{ic-Chapelle2011,j-Scott2015} ---Appendix~\ref{assec:static_bandits_experiments_logistic}.
Results showcase satisfactory performance across a wide range of stationary bandit parameterizations and sizes,
as SMC-based policies achieve the right exploration-exploitation tradeoff.

For results we present below, we simulate different parameterizations of dynamic linear models described in Section~\ref{ssec:linear_mixing_dynamics},
and present results for a variety of MAB environments with reward functions detailed in Sections~\ref{ssec:dynamic_bandits_gaussian},~\ref{ssec:dynamic_bandits_logistic} and~\ref{ssec:dynamic_bandits_categorical}.
Section~\ref{ssec:logged_data_bandits} illustrates
the ability of SMC-based bandit policies
to capture non-stationary trends in personalized news article recommendations.

The main evaluation metric is the cumulative regret of the bandit agent, as defined in Equation~\eqref{eq:mab_cumulative_regret},
with results averaged over 500 realizations.
We present results for SMC-based policies with $M=2000$ samples,
and provide an evaluation of the impact of $M$ in Appendix~\ref{asec:dynamic_bandits}.

\subsection{Non-stationary, linear Gaussian rewards}
\label{ssec:dynamic_bandits_gaussian}
We simulate the following two-armed, contextual ($x_t\in\Real^2, \forall t$), linear Gaussian bandit:
\begin{equation}
\text{Scenario A}
\begin{cases}
	\vspace*{1ex}
	p(\theta_{t,a=0}|\theta_{t-1,a=0}): \\ \vspace*{1ex}
	\hspace*{10ex}\begin{pmatrix}
	\theta_{t,a=0,0}\\
	\theta_{t,a=0,1}\\
	\end{pmatrix} = \begin{pmatrix}
	0.9 & -0.1 \\
	-0.1 & 0.9 \\
	\end{pmatrix} \begin{pmatrix}
	\theta_{t-1,a=0,0}\\
	\theta_{t-1,a=0,1}\\
	\end{pmatrix} + \epsilon_{a=0} \;, \\ \vspace*{1ex}
	\hspace*{40ex} \text{where } \;  \epsilon_{a=0} \sim \N{\epsilon|0,0.01 \cdot\mathrm{I}},\\
	
	\vspace*{1ex}
	p(\theta_{t,a=1}|\theta_{t-1,a=1}): \\ \vspace*{1ex}
	\hspace*{10ex}\begin{pmatrix}
	\theta_{t,a=1,0}\\
	\theta_{t,a=1,1}\\
	\end{pmatrix} = \begin{pmatrix}
	0.9 & 0.1 \\
	0.1 & 0.9 \\
	\end{pmatrix} \begin{pmatrix}
	\theta_{t-1,a=1,0}\\
	\theta_{t-1,a=1,1}\\
	\end{pmatrix} + \epsilon_{a=1} \;, \\ \vspace*{1ex}
	\hspace*{40ex} \text{where } \;  \epsilon_{a=1} \sim \N{\epsilon|0,0.01 \cdot\mathrm{I}},\\
	
	p_a(Y|x,\theta_{t,a})=\N{Y|x^\top \theta_{t,a}, \sigma_a^2} \;.
\end{cases}
\label{eq:linear_mixing_dynamics_a}
\end{equation}

\begin{equation}
\text{Scenario B}
\begin{cases}
	\vspace*{1ex}
	p(\theta_{t,a=0}|\theta_{t-1,a=0}): \\ \vspace*{1ex}
	\hspace*{10ex}\begin{pmatrix}
	\theta_{t,a=0,0}\\
	\theta_{t,a=0,1}\\
	\end{pmatrix} = \begin{pmatrix}
	0.5 & 0.0 \\
	0.0 & 0.5 \\
	\end{pmatrix} \begin{pmatrix}
	\theta_{t-1,a=0,0}\\
	\theta_{t-1,a=0,1}\\
	\end{pmatrix} + \epsilon_{a=0} \;, \\ \vspace*{1ex}
	\hspace*{40ex} \text{where } \;  \epsilon_{a=0} \sim \N{\epsilon|0,0.01 \cdot\mathrm{I}},\\
	
	\vspace*{1ex}
	p(\theta_{t,a=1}|\theta_{t-1,a=1}): \\ \vspace*{1ex}
	\hspace*{10ex}\begin{pmatrix}
	\theta_{t,a=1,0}\\
	\theta_{t,a=1,1}\\
	\end{pmatrix} = \begin{pmatrix}
	0.9 & 0.1 \\
	0.1 & 0.9 \\
	\end{pmatrix} \begin{pmatrix}
	\theta_{t-1,a=1,0}\\
	\theta_{t-1,a=1,1}\\
	\end{pmatrix} + \epsilon_{a=1} \;, \\ \vspace*{1ex}
	\hspace*{40ex} \text{where } \;  \epsilon_{a=1} \sim \N{\epsilon|0,0.01 \cdot\mathrm{I}}, \\
	
	p_a(Y|x,\theta_{t,a})=\N{Y|x^\top \theta_{t,a}, \sigma_a^2} \;.
\end{cases}
\label{eq:linear_mixing_dynamics_b}
\end{equation}
 
\begin{figure}[!h]
	\centering
	\begin{subfigure}[b]{0.45\textwidth}
		\includegraphics[width=\textwidth]{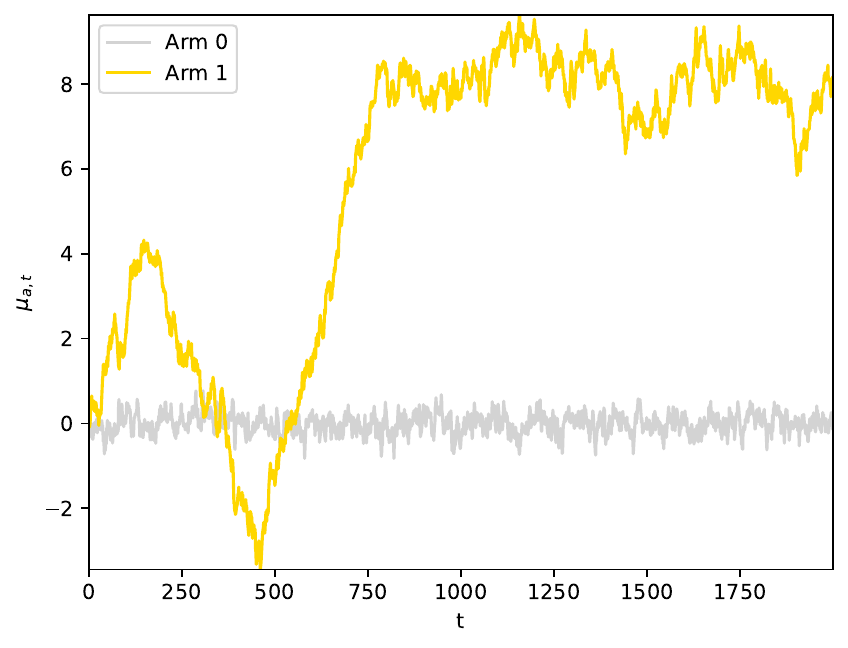}
		\caption{Expected per-arm rewards over time for Scenario A in Equation~\eqref{eq:linear_mixing_dynamics_a}.}
		\label{fig:linear_mixing_dynamics_a_gaussian}
	\end{subfigure}\qquad
	\begin{subfigure}[b]{0.45\textwidth}
		\includegraphics[width=\textwidth]{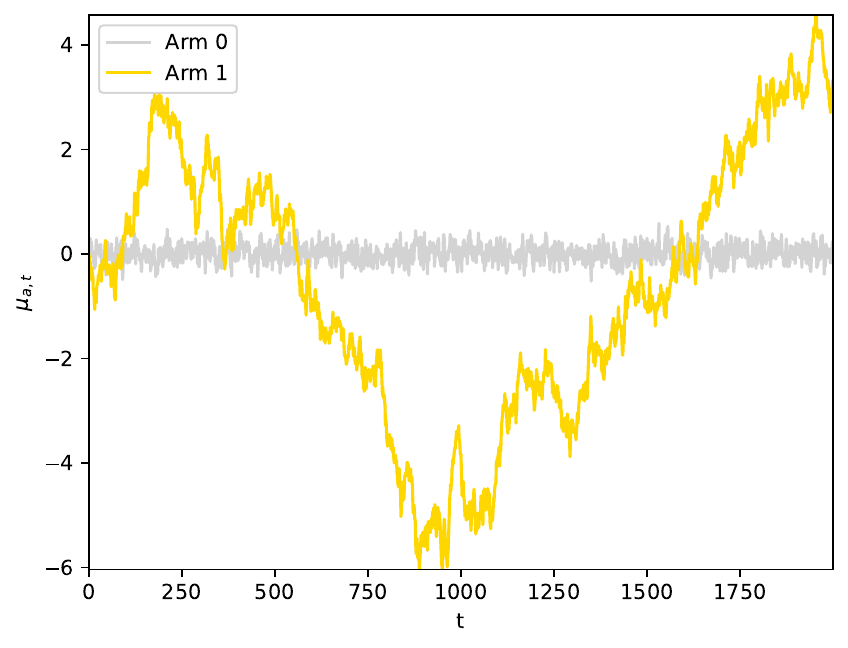}
		\caption{Expected per-arm rewards over time for Scenario B in Equation~\eqref{eq:linear_mixing_dynamics_a}.}
		\label{fig:linear_mixing_dynamics_b_gaussian}
	\end{subfigure}
	
	\begin{subfigure}[b]{0.47\textwidth}
		\includegraphics[width=\textwidth]{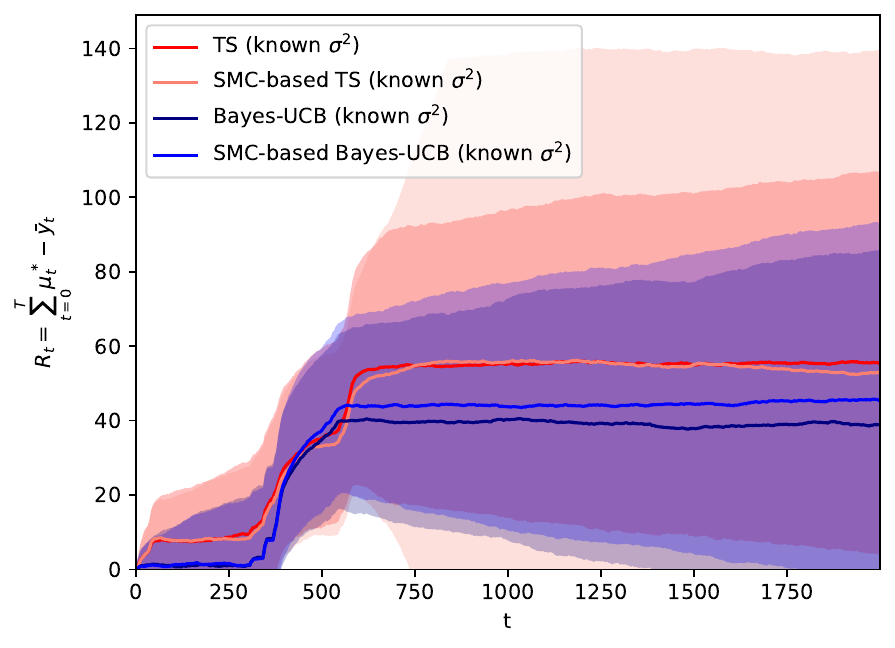}
		\caption{Cumulative regret for SMC-based Bayesian policies in scenario A: known dynamic parameters.}
		\label{fig:dynamic_bandits_linearGaussian_a_cstatic_dknown_knownsigma}
	\end{subfigure}\qquad
	\begin{subfigure}[b]{0.47\textwidth}
		\includegraphics[width=\textwidth]{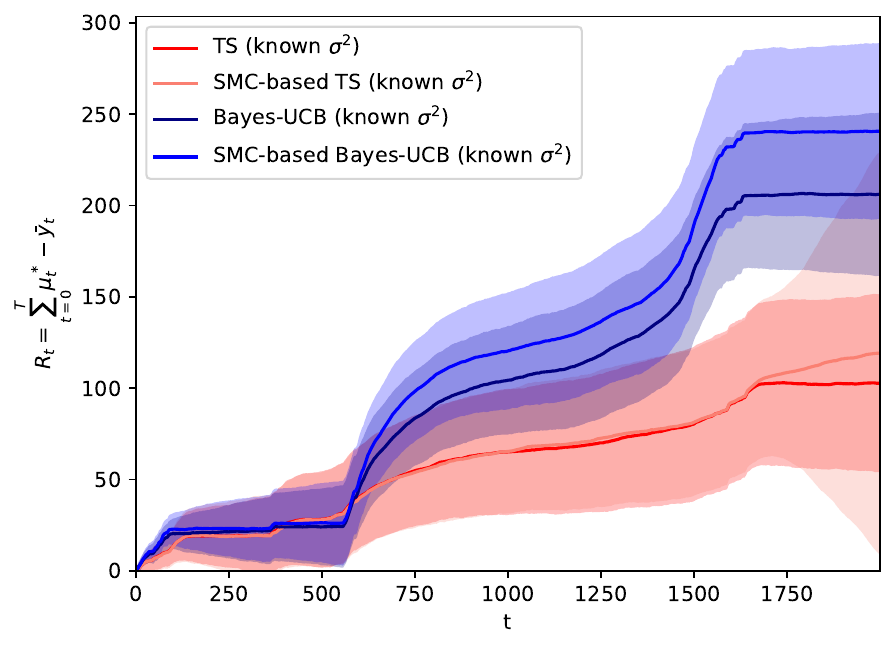}
		\caption{Cumulative regret for SMC-based Bayesian policies in scenario B: known dynamic parameters.}
		\label{fig:dynamic_bandits_linearGaussian_b_cstatic_dknown_knownsigma}
	\end{subfigure}
	
	\caption{
		Mean regret (standard deviation shown as shaded region) in contextual, linear Gaussian bandit Scenarios A and B
		described in Equations~\eqref{eq:linear_mixing_dynamics_a}--\eqref{eq:linear_mixing_dynamics_b},
		when the bandit agent knows the latent dynamic parameterization.
		Notice how
		in Figures~\ref{fig:dynamic_bandits_linearGaussian_a_cstatic_dknown_knownsigma}--\ref{fig:dynamic_bandits_linearGaussian_b_cstatic_dknown_knownsigma}
		regret increases when the optimal arms swap
		(as shown in Figures~\ref{fig:linear_mixing_dynamics_a_gaussian}--\ref{fig:linear_mixing_dynamics_b_gaussian}).
		SMC-based policies successfully find the right exploration-exploitation tradeoff,
		with minimal additional regret incurred in comparison to their analytical alternatives. 
	}
	\label{fig:dynamic_bandits_linearGaussian_dknown}
\end{figure}

The expected rewards driven by the dynamics of Equations~\eqref{eq:linear_mixing_dynamics_a} and~\eqref{eq:linear_mixing_dynamics_b} change over time,
inducing switches on the identity of the optimal arm.
For instance, for a given realization of Scenario A shown in Figure~\ref{fig:linear_mixing_dynamics_a_gaussian},
there is an optimal arm swap between time-instants $t=(300, 550)$, with arm 1 becoming the optimal for all $t\geq600$;
for a realization of Scenario B illustrated in Figure~\ref{fig:linear_mixing_dynamics_b_gaussian},
there is an optimal arm change around $t=100$, a swap around $t=600$,
with arm 1 becoming optimal again after $t\geq1600$.

\begin{figure}[!h]
	\centering
	\begin{subfigure}[b]{0.45\textwidth}
		\includegraphics[width=\textwidth]{./fods_figs/dynamic/linearGaussian/dynamics_a}
		\caption{Expected per-arm rewards over time for Scenario A in Equation~\eqref{eq:linear_mixing_dynamics_a}.}
		\label{fig:linear_mixing_dynamics_a_gaussian_2}
	\end{subfigure}\qquad
	\begin{subfigure}[b]{0.45\textwidth}
		\includegraphics[width=\textwidth]{./fods_figs/dynamic/linearGaussian/dynamics_b}
		\caption{Expected per-arm rewards over time for Scenario B in Equation~\eqref{eq:linear_mixing_dynamics_a}.}
		\label{fig:linear_mixing_dynamics_b_gaussian_2}
	\end{subfigure}
	
	\begin{subfigure}[b]{0.47\textwidth}
		\includegraphics[width=\textwidth]{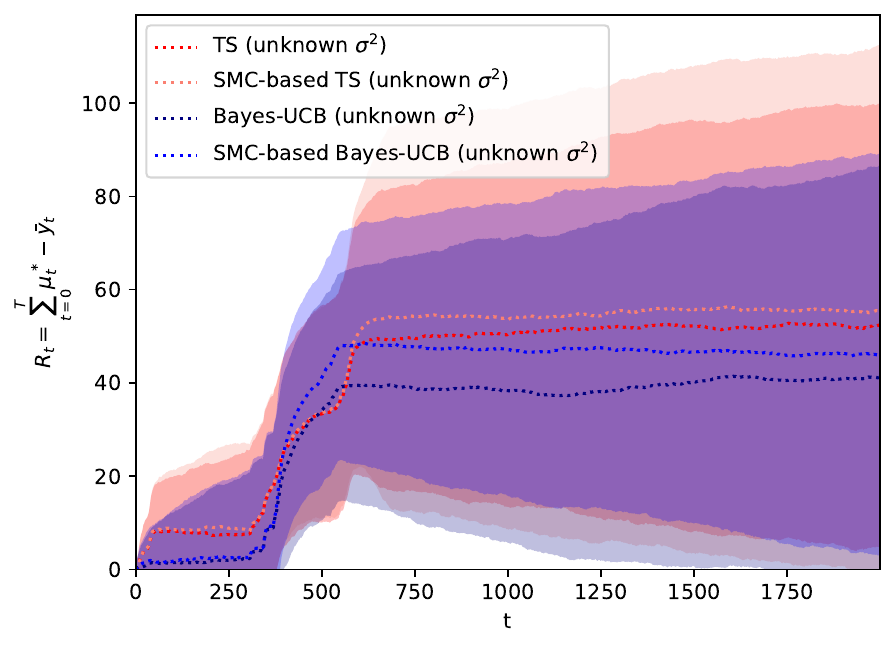}
		\caption{Cumulative regret for SMC-based Bayesian policies in scenario A: known dynamic parameters, unknown $\sigma_a, \forall a$.}
		\label{fig:dynamic_bandits_linearGaussian_a_cstatic_dknown_unknownsigma}
	\end{subfigure}\qquad
	\begin{subfigure}[b]{0.47\textwidth}
		\includegraphics[width=\textwidth]{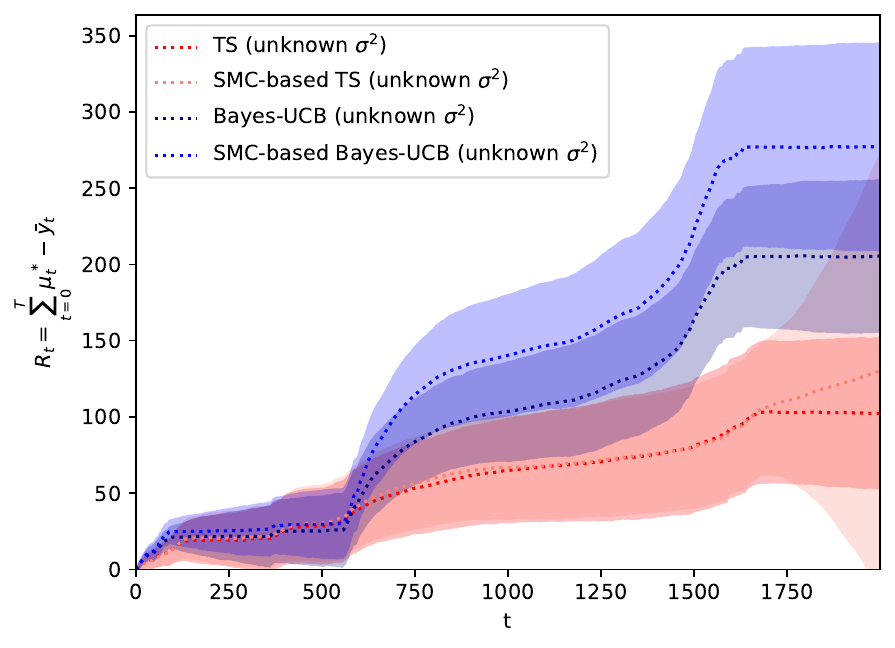}
		\caption{Cumulative regret for SMC-based Bayesian policies in scenario B: known dynamic parameters, unknown $\sigma_a, \forall a$.}
		\label{fig:dynamic_bandits_linearGaussian_b_cstatic_dknown_unknownsigma}
	\end{subfigure}
	
	\begin{subfigure}[b]{0.47\textwidth}
		\includegraphics[width=\textwidth]{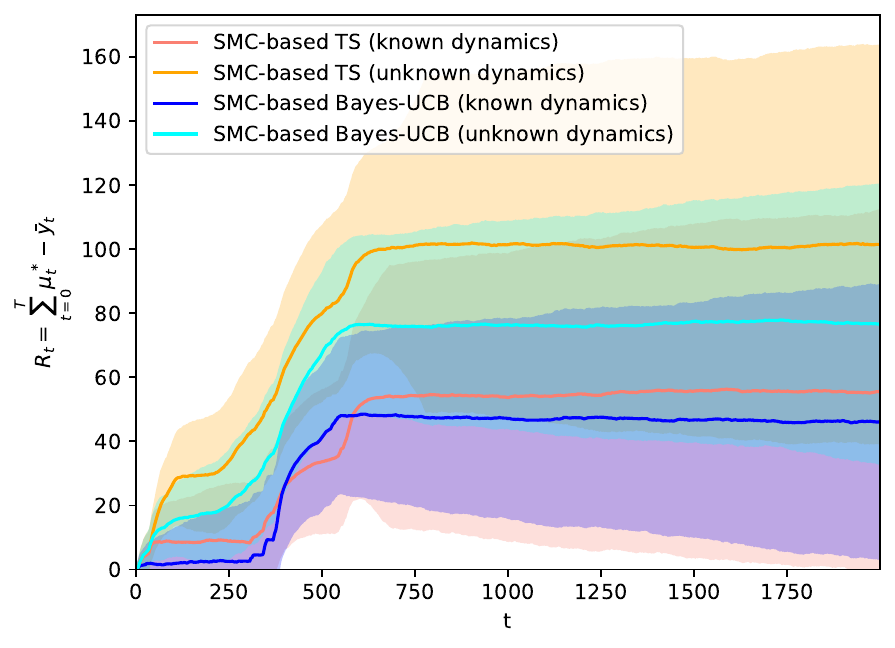}
		\caption{Cumulative regret for SMC-based Bayesian policies in scenario A: unknown dynamic parameters $L_a,\Sigma_a,\sigma_a^2, \forall a$.}
		\label{fig:dynamic_bandits_linearGaussian_a_cstatic_dunknown}
	\end{subfigure}\qquad
	\begin{subfigure}[b]{0.47\textwidth}
		\includegraphics[width=\textwidth]{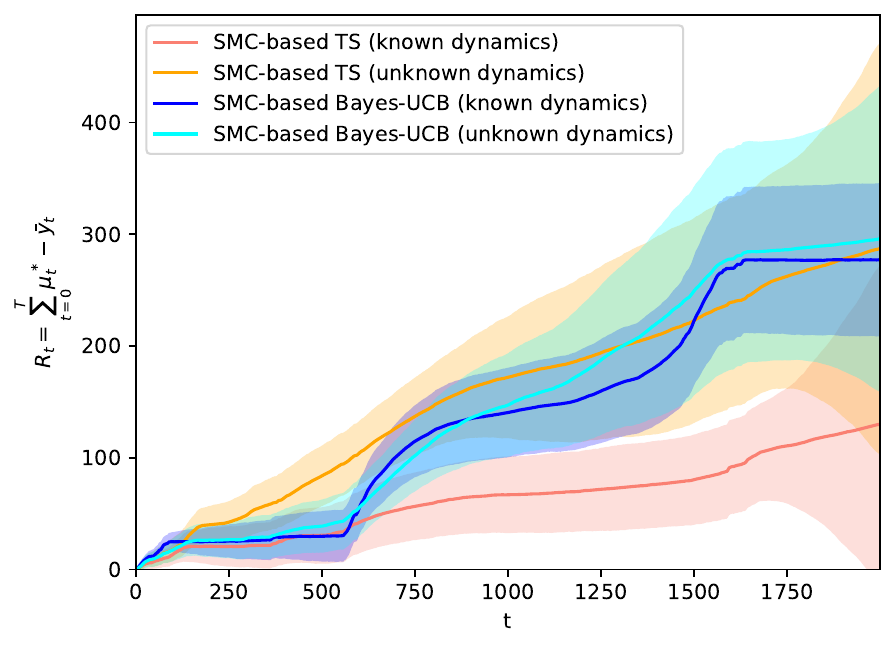}
		\caption{Cumulative regret for SMC-based Bayesian policies in scenario B: unknown dynamic parameters $L_a,\Sigma_a,\sigma_a^2, \forall a$.}
		\label{fig:dynamic_bandits_linearGaussian_b_cstatic_dunknown}
	\end{subfigure}
	\caption{
		Mean regret (standard deviation shown as shaded region) in contextual, linear Gaussian bandit Scenarios A and B
		described in Equations~\eqref{eq:linear_mixing_dynamics_a}--\eqref{eq:linear_mixing_dynamics_b},
		in the realistic setting of unknown dynamic parameters.
		Notice how
			in Figures~\ref{fig:dynamic_bandits_linearGaussian_a_cstatic_dknown_unknownsigma}--\ref{fig:dynamic_bandits_linearGaussian_b_cstatic_dunknown}
		the regret increases when the optimal arms swap
			(as shown in Figures~\ref{fig:linear_mixing_dynamics_a_gaussian_2}--\ref{fig:linear_mixing_dynamics_b_gaussian_2}).
		SMC-based policies find the right exploration-exploitation tradeoff even when the latent dynamic parameters are unknown.}
	\label{fig:dynamic_bandits_linearGaussian_dunknown}
\end{figure}

Empirical results for SMC-based Bayesian policies in scenarios described by Equations~\eqref{eq:linear_mixing_dynamics_a} and~\eqref{eq:linear_mixing_dynamics_b}
are shown in Figures~\ref{fig:dynamic_bandits_linearGaussian_dknown} and \ref{fig:dynamic_bandits_linearGaussian_dunknown}.

We study linear dynamics with Gaussian reward distributions with known parameters in Figure~\ref{fig:dynamic_bandits_linearGaussian_dknown},
of interest as it allows us to validate the SMC-based random measure in comparison to the optimal, closed-form posterior
---the Kalman filter~\cite{j-Kalman1960}---
under the assumption of known dynamic parameters.

We observe satisfactory cumulative regret performance in Figure~\ref{fig:dynamic_bandits_linearGaussian_dknown}:
\ie SMC-based Bayesian agents' cumulative regret is sublinear.
Policies that compute and use SMC random measure posteriors
incur in minimal regret loss 
in comparison to the optimal Kalman filter-based agent.
Namely, the shape of the regret curves of \textit{TS} and \textit{SMC-based TS}
(\textit{Bayes-UCB} and \textit{SMC based Bayes-UCB}, respectively) in Figure~\ref{fig:dynamic_bandits_linearGaussian_dknown} is equivalent,
with minimal differences in average cumulative regret when compared to the volatility across realizations.
Importantly, all policies are able to adapt to the changes over time of the identify of the optimal arm. 

We illustrate in Figure~\ref{fig:dynamic_bandits_linearGaussian_dunknown}
a more realistic scenario, where the dynamic parameterization is unknown to the bandit agent.

We observe in Figures~\ref{fig:dynamic_bandits_linearGaussian_a_cstatic_dknown_unknownsigma}--\ref{fig:dynamic_bandits_linearGaussian_b_cstatic_dknown_unknownsigma} that,
in the case of unknown reward variances ($\sigma_a^2, \forall a)$,
SMC-based policies perform comparably well.
In these cases,
the agents' reward model is not Gaussian,
but Student-t distributed, as per the marginalized posterior in Equation~\eqref{eq:t_posterior_mean}.
The regret loss associated with the uncertainty about $\sigma_a^2$ is minimal for SMC-based Bayesian agents,
and does not hinder the ability of the proposed SMC-based policies
to find the right exploration-exploitation balance:
\ie regret is sublinear, and the agents adapt to switches in the identity of the optimal arm.

We illustrate in Figures~\ref{fig:dynamic_bandits_linearGaussian_a_cstatic_dunknown}--\ref{fig:dynamic_bandits_linearGaussian_b_cstatic_dunknown}
the most realistic, yet challenging, non-stationary contextual Gaussian bandit case:
one where none of the parameters of the model are known.
In this case, the agent must sequentially learn both the underlying dynamics ($L_a,\Sigma_a; \forall a$)
and the conditional reward function's variance ($\sigma_a^2, \forall a)$,
in order to infer the posterior distribution over the latent, time-varying sufficient statistics of interest,
to enable informed sequential decision making.

Cumulative regret results in Figures~\ref{fig:dynamic_bandits_linearGaussian_a_cstatic_dunknown}--\ref{fig:dynamic_bandits_linearGaussian_b_cstatic_dunknown}
showcase a regret performance loss due to the need to learn all these unknown parameters.
We observe noticeable (almost linear) regret increases when the dynamics of the parameters swap the identity of the optimal arm.
However, SMC-based Thompson sampling and Bayes-UCB agents are able to learn the evolution of the dynamic latent parameters,
and the corresponding time-varying expected rewards,
with enough accuracy to attain good exploration-exploitation balance:
\ie sublinear regret curves indicate the agent identified and played the optimal arm repeatedly.
Figure~\ref{fig:dynamic_bandits_linearGaussian_a_cstatic_dunknown} is clear evidence of the SMC-based agents' ability to recover from linear to no-regret regimes.

\clearpage
\subsection{Non-stationary, logistic rewards}
\label{ssec:dynamic_bandits_logistic}
We here evaluate non-stationary, contextual, binary reward bandits.
We resort to the logistic reward function described in Equation~\eqref{eq:logistic_rewards},
with time-varying, latent parameter dynamics as described in the following scenarios:
\begin{equation}
\text{Scenario C}
\begin{cases}
\vspace*{1ex}
p(\theta_{t,a=0}|\theta_{t-1,a=0}): \\ \vspace*{1ex}
\hspace*{10ex}\begin{pmatrix}
\theta_{t,a=0,0}\\
\theta_{t,a=0,1}\\
\end{pmatrix} = \begin{pmatrix}
0.9 & -0.1 \\
-0.1 & 0.9 \\
\end{pmatrix} \begin{pmatrix}
\theta_{t-1,a=0,0}\\
\theta_{t-1,a=0,1}\\
\end{pmatrix} + \epsilon_{a=0} \;, \\ \vspace*{1ex}
\hspace*{40ex} \text{where } \;  \epsilon_{a=0} \sim \N{\epsilon|0,0.01 \cdot\mathrm{I}},\\

\vspace*{1ex}
p(\theta_{t,a=1}|\theta_{t-1,a=1}): \\ \vspace*{1ex}
\hspace*{10ex}\begin{pmatrix}
\theta_{t,a=1,0}\\
\theta_{t,a=1,1}\\
\end{pmatrix} = \begin{pmatrix}
0.9 & 0.1 \\
0.1 & 0.9 \\
\end{pmatrix} \begin{pmatrix}
\theta_{t-1,a=1,0}\\
\theta_{t-1,a=1,1}\\
\end{pmatrix} + \epsilon_{a=1}  \;, \\ \vspace*{1ex}
\hspace*{40ex} \text{where } \;  \epsilon_{a=1} \sim \N{\epsilon|0,0.01 \cdot\mathrm{I}},\\

p_a(Y|x,\theta_{t,a})=\frac{e^{y\cdot(x^\top\theta_{t,a}) }}{1+e^{(x^\top\theta_{t,a})}} \; .
\end{cases}
\label{eq:linear_mixing_dynamics_c}
\end{equation}

\begin{equation}
\text{Scenario D}
\begin{cases}
\vspace*{1ex}
p(\theta_{t,a=0}|\theta_{t-1,a=0}): \\ \vspace*{1ex}
\hspace*{10ex}\begin{pmatrix}
\theta_{t,a=0,0}\\
\theta_{t,a=0,1}\\
\end{pmatrix} = \begin{pmatrix}
0.5 & 0.0 \\
0.0 & 0.5 \\
\end{pmatrix} \begin{pmatrix}
\theta_{t-1,a=0,0}\\
\theta_{t-1,a=0,1}\\
\end{pmatrix} + \epsilon_{a=0}  \;, \\ \vspace*{1ex}
\hspace*{40ex} \text{where } \;  \epsilon_{a=0} \sim \N{\epsilon|0,0.01 \cdot\mathrm{I}},\\

\vspace*{1ex}
p(\theta_{t,a=1}|\theta_{t-1,a=1}): \\ \vspace*{1ex}
\hspace*{10ex}\begin{pmatrix}
\theta_{t,a=1,0}\\
\theta_{t,a=1,1}\\
\end{pmatrix} = \begin{pmatrix}
0.9 & 0.1 \\
0.1 & 0.9 \\
\end{pmatrix} \begin{pmatrix}
\theta_{t-1,a=1,0}\\
\theta_{t-1,a=1,1}\\
\end{pmatrix} + \epsilon_{a=1}  \;, \\ \vspace*{1ex}
\hspace*{40ex} \text{where } \;  \epsilon_{a=1} \sim \N{\epsilon|0,0.01 \cdot\mathrm{I}}, \\

p_a(Y|x,\theta_{t,a})=\frac{e^{y\cdot(x^\top\theta_{t,a}) }}{1+e^{(x^\top\theta_{t,a})}} \; .
\end{cases}
\label{eq:linear_mixing_dynamics_d}
\end{equation}

For bandits with logistic rewards,
there are no closed form posteriors;
hence, one needs to resort to approximations,
\eg a Laplace approximation as in~\citep{ic-Chapelle2011} for the stationary case.
However, there are no bandit algorithms for the non-stationary logistic scenarios described above.
On the contrary, SMC-based Bayesian policies can easily accommodate this setting,
by updating posterior random measures $p_M(\theta_{t}|\HH_{1:t})$ as in Algorithm~\ref{alg:sir-mab},
for both stationary (evaluated in Appendix~\ref{assec:static_bandits_experiments_logistic}) and non-stationary bandits we report here.

\begin{figure}[!h]
	\begin{subfigure}[b]{0.45\textwidth}
		\includegraphics[width=\textwidth]{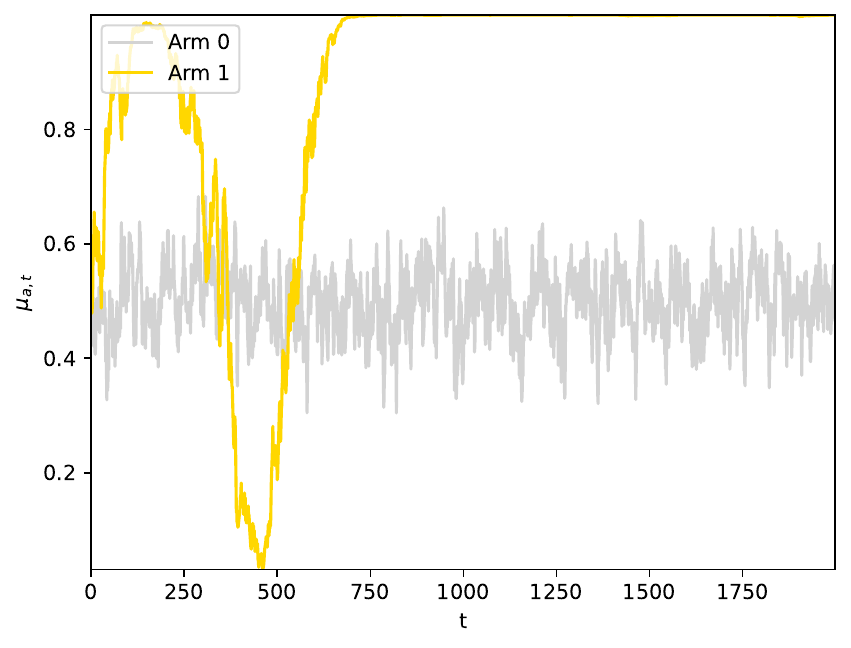}
		\caption{Expected per-arm rewards over time for Scenario C in Equation~\eqref{eq:linear_mixing_dynamics_c}.
			Notice the early optimal arm change at $t\approx600$.
		}
		\label{fig:linear_mixing_dynamics_c_logistic}
	\end{subfigure}\qquad
	\begin{subfigure}[b]{0.45\textwidth}
		\includegraphics[width=\textwidth]{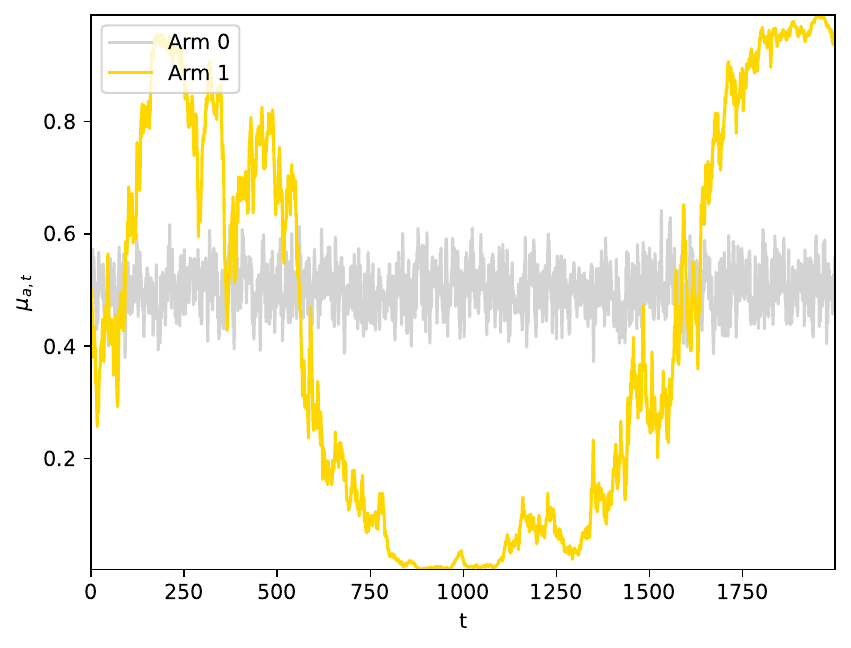}
		\caption{Expected per-arm rewards over time for Scenario D in Equation~\eqref{eq:linear_mixing_dynamics_d}.
			Notice the late optimal arm change at $t\approx1650$.
		}
		\label{fig:linear_mixing_dynamics_d_logistic}
	\end{subfigure}
	
	\begin{subfigure}[b]{0.47\textwidth}
		\includegraphics[width=\textwidth]{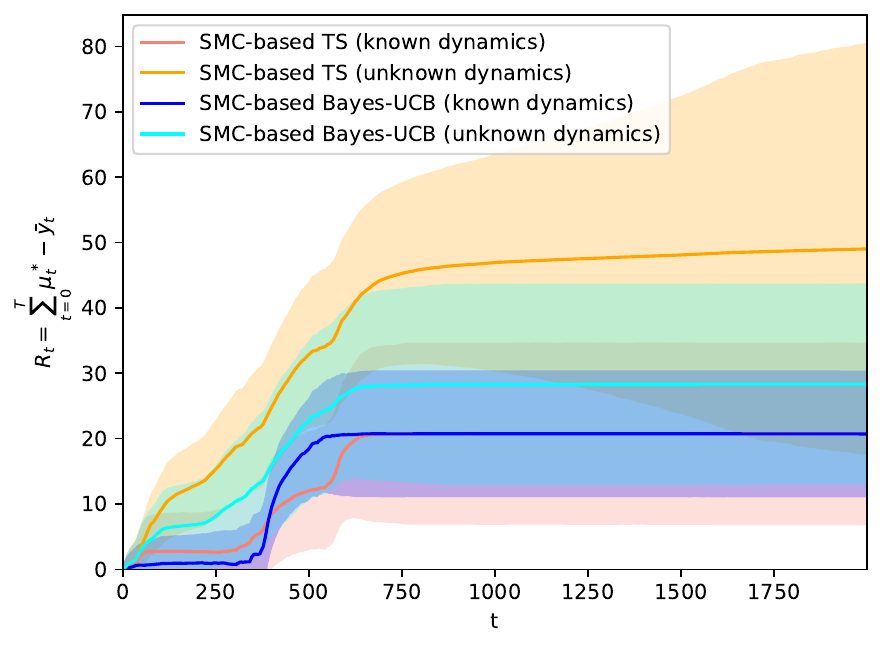}
		\caption{Cumulative regret for SMC-based Bayesian policies in scenario C: known and unknown dynamic parameters.}
		\label{fig:dynamic_bandits_c_logistic_cstatic}
	\end{subfigure}\qquad
	\begin{subfigure}[b]{0.47\textwidth}
		\includegraphics[width=\textwidth]{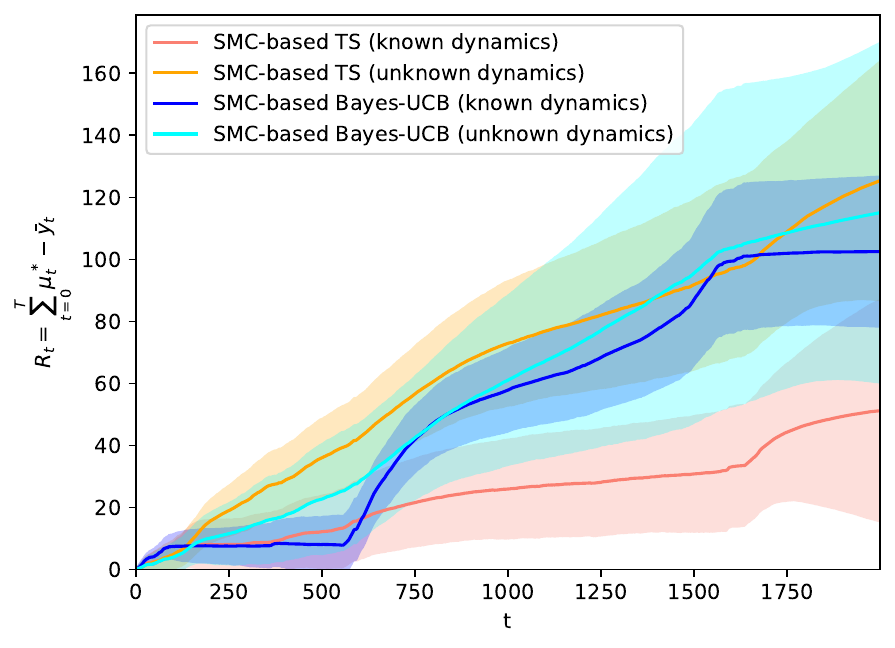}
		\caption{Cumulative regret for SMC-based Bayesian policies in scenario D: known and unknown dynamic parameters.}
		\label{fig:dynamic_bandits_d_logistic_cstatic}
	\end{subfigure}
	\caption{
		Mean regret (standard deviation shown as shaded region) in contextual linear logistic dynamic bandit Scenarios C and D
		described in Equations~\eqref{eq:linear_mixing_dynamics_c}--\eqref{eq:linear_mixing_dynamics_d}.
		Notice the difference in early ($t\approx600$ in Scenario C) and late ($t\approx1650$ in Scenario D) optimal arm changes,
			as illustrated in Figures~\ref{fig:linear_mixing_dynamics_c_logistic}--\ref{fig:linear_mixing_dynamics_d_logistic},
		and their impact in regret,
			as showcased in Figures~\ref{fig:dynamic_bandits_c_logistic_cstatic}--\ref{fig:dynamic_bandits_d_logistic_cstatic}.
		SMC-based Bayesian policies adapt and find the right exploration-exploitation tradeoff.}
	\label{fig:dynamic_bandits_logistic}
\end{figure}

Figure~\ref{fig:dynamic_bandits_logistic} illustrates how
SMC-based Bayesian policies adapt to non-stationary optimal arm switches under contextual, binary reward observations,
achieving sublinear regret.
Results in Figures~\ref{fig:linear_mixing_dynamics_d_logistic}--\ref{fig:dynamic_bandits_d_logistic_cstatic}
also showcase how a bandit agent's regret suffers when learning unknown parameters of the latent dynamics.
Even though this is a particularly challenging problem,
presented evidence suggests that
SMC-based policies do learn the underlying latent dynamics from contextual binary rewards.

Notably, proposed policies are able to successfully identify which arm to play:
\ie both \textit{SMC-based TS and SMC-based UCB} ---with no dynamic parameter knowledge---
are able to flatten their regret
for $t\geq 650$ in Figure~\ref{fig:dynamic_bandits_c_logistic_cstatic} and
$t\geq 1750$ in Figure~\ref{fig:dynamic_bandits_d_logistic_cstatic}.

\clearpage
\subsection{Non-stationary, categorical rewards}
\label{ssec:dynamic_bandits_categorical}
We evaluate SMC-based Bayesian policies in a bandit setting that remains elusive to state-of-the-art bandit algorithms:
non-stationary bandits with discrete-categorical, contextual rewards.
We simulate the following two- and three-armed categorical bandit scenarios,
where numerical rewards $Y=c$, for $c\in\{0,1,2\}$,
depend on a two-dimensional context $x_t\in\Real^2$,
with time-varying parameters $\theta_{a,c,t}$
obeying the following dynamics:

\begin{equation}
\text{Scenario E}
\begin{cases}
\vspace*{1ex}
p(\theta_{t,a=0,c}|\theta_{t-1,a=0,c})\;, \; \forall c \in \{0,1,2\}:\\ \vspace*{1ex}
\hspace*{10ex}\begin{pmatrix}
\theta_{t,a=0,c,0}\\
\theta_{t,a=0,c,1}\\
\end{pmatrix} = \begin{pmatrix}
0.9 & -0.1 \\
-0.1 & 0.9 \\
\end{pmatrix} \begin{pmatrix}
\theta_{t-1,a=0,c,0}\\
\theta_{t-1,a=0,c,1}\\
\end{pmatrix} + \epsilon_{a=0,c} \;, \\ \vspace*{1ex}
\hspace*{38ex} \text{where } \; \epsilon_{a=0,c} \sim \N{\epsilon|0,0.01 \cdot\mathrm{I}}, \\

\vspace*{1ex}
p(\theta_{t,a=1,c}|\theta_{t-1,a=1,c})\;, \; \forall c \in \{0,1,2\}:\\ \vspace*{1ex}
\hspace*{10ex}\begin{pmatrix}
\theta_{t,a=1,c,0}\\
\theta_{t,a=1,c,1}\\
\end{pmatrix} = \begin{pmatrix}
0.9 & 0.1 \\
0.1 & 0.9 \\
\end{pmatrix} \begin{pmatrix}
\theta_{t-1,a=1,c,0}\\
\theta_{t-1,a=1,c,1}\\
\end{pmatrix} + \epsilon_{a=1,c} \;, \\ \vspace*{1ex}
\hspace*{38ex} \text{where } \;  \epsilon_{a=1,c} \sim \N{\epsilon|0,0.01 \cdot\mathrm{I}},\\

p_a(Y=c|x,\theta_{t,a})=\frac{e^{(x^\top\theta_{t,a,c})}}{\sum_{c'=1}^C e^{(x^\top\theta_{t,a,c'})} } \; .
\end{cases}
\label{eq:linear_mixing_dynamics_e}
\end{equation}

\begin{equation}
\text{Scenario F}
\begin{cases}
\vspace*{1ex}
p(\theta_{t,a=0,c}|\theta_{t-1,a=0,c}) \;, \; \forall c \in \{0,1,2\}:\\ \vspace*{1ex}
\hspace*{10ex}\begin{pmatrix}
\theta_{t,a=0,c,0}\\
\theta_{t,a=0,c,1}\\
\end{pmatrix} = \begin{pmatrix}
0.9 & -0.1 \\
-0.1 & 0.9 \\
\end{pmatrix} \begin{pmatrix}
\theta_{t-1,a=0,c,0}\\
\theta_{t-1,a=0,c,1}\\
\end{pmatrix} + \epsilon_{a=0,c} \;, \\ \vspace*{1ex}
\hspace*{38ex} \text{where } \; \epsilon_{a=0,c} \sim \N{\epsilon|0,0.01 \cdot\mathrm{I}}, \\

\vspace*{1ex}
p(\theta_{t,a=1,c}|\theta_{t-1,a=1,c})\;, \; \forall c \in \{0,1,2\}:\\ \vspace*{1ex}
\hspace*{10ex}\begin{pmatrix}
\theta_{t,a=1,c,0}\\
\theta_{t,a=1,c,1}\\
\end{pmatrix} = \begin{pmatrix}
0.9 & 0.1 \\
0.1 & 0.9 \\
\end{pmatrix} \begin{pmatrix}
\theta_{t-1,a=1,c,0}\\
\theta_{t-1,a=1,c,1}\\
\end{pmatrix} + \epsilon_{a=1,c} \;, \\ \vspace*{1ex}
\hspace*{38ex} \text{where } \; \epsilon_{a=1,c} \sim \N{\epsilon|0,0.01 \cdot\mathrm{I}},\\

\vspace*{1ex}
p(\theta_{t,a=2,c}|\theta_{t-1,a=2,c})\;, \; \forall c \in \{0,1,2\}:\\ \vspace*{1ex}
\hspace*{10ex}\begin{pmatrix}
\theta_{t,a=2,c,0}\\
\theta_{t,a=2,c,1}\\
\end{pmatrix} = \begin{pmatrix}
0.9 & 0.1 \\
0.1 & 0.9 \\
\end{pmatrix} \begin{pmatrix}
\theta_{t-1,a=2,c,0}\\
\theta_{t-1,a=2,c,1}\\
\end{pmatrix} + \epsilon_{a=2,c} \;, \\ \vspace*{1ex}
\hspace*{38ex} \text{where } \; \epsilon_{a=2,c} \sim \N{\epsilon|0,0.01 \cdot\mathrm{I}},\\

p_a(Y=c|x,\theta_{t,a})=\frac{e^{(x^\top\theta_{t,a,c})}}{\sum_{c'=1}^C e^{(x^\top\theta_{t,a,c'})} } \; .
\end{cases}
\label{eq:linear_mixing_dynamics_f}
\end{equation}

\begin{figure}[!ht]
	\centering
	\begin{subfigure}[b]{0.47\textwidth}
		\includegraphics[width=\textwidth]{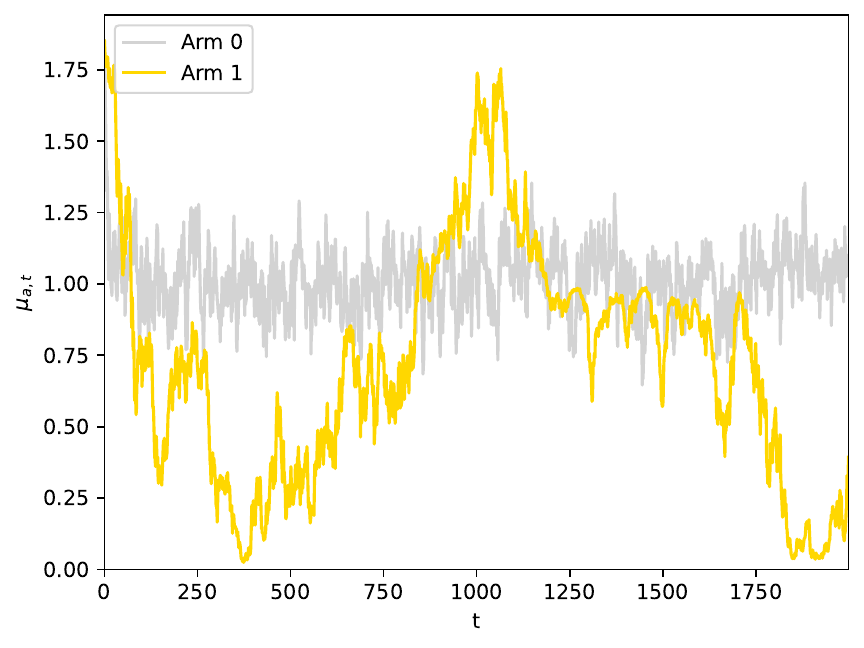}
		\caption{Expected per-arm rewards over time for Scenario E in Equation~\eqref{eq:linear_mixing_dynamics_e}.
			Notice the optimal arm changes around $t\approx1000$.}
		\label{fig:linear_mixing_dynamics_e_softmax}%
	\end{subfigure}\qquad
	\begin{subfigure}[b]{0.47\textwidth}
		\includegraphics[width=\textwidth]{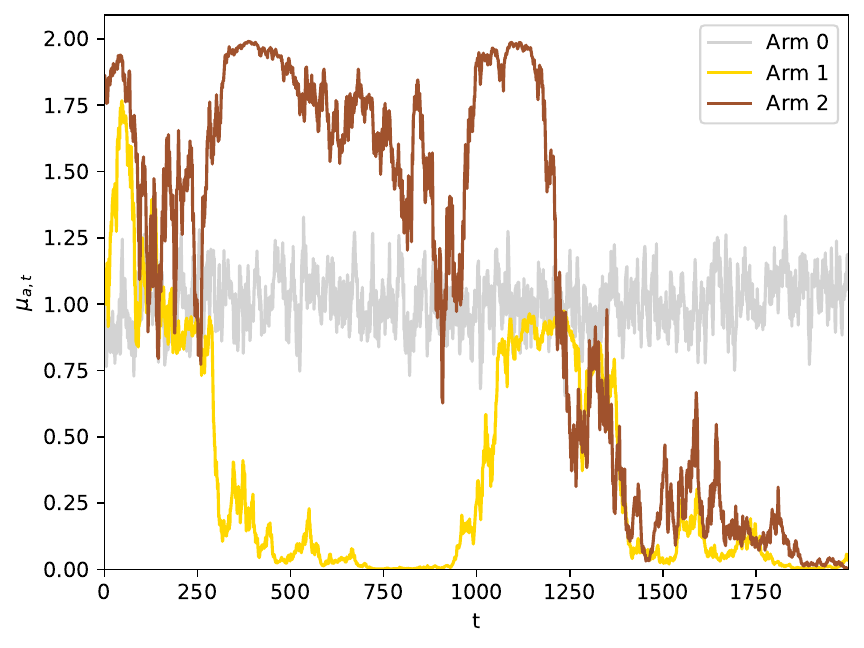}
		\caption{Expected per-arm rewards over time for Scenario F in Equation~\eqref{eq:linear_mixing_dynamics_e}.
			Notice the optimal arm change around $t\approx1250$.}
		\label{fig:linear_mixing_dynamics_f_softmax}
	\end{subfigure} %
	
	\begin{subfigure}[b]{0.47\textwidth}
		\includegraphics[width=\textwidth]{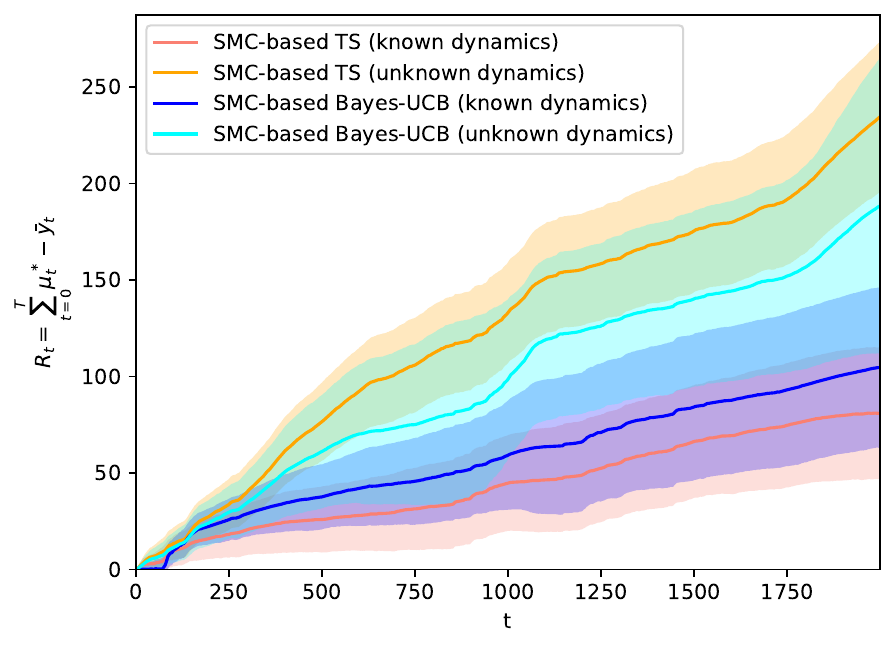}
		\caption{Cumulative regret for SMC-based Bayesian policies in scenario E: known and unknown dynamic parameters.}
		\label{fig:dynamic_bandits_e_softmax_cstatic}%
	\end{subfigure}\qquad
	\begin{subfigure}[b]{0.47\textwidth}
		\includegraphics[width=\textwidth]{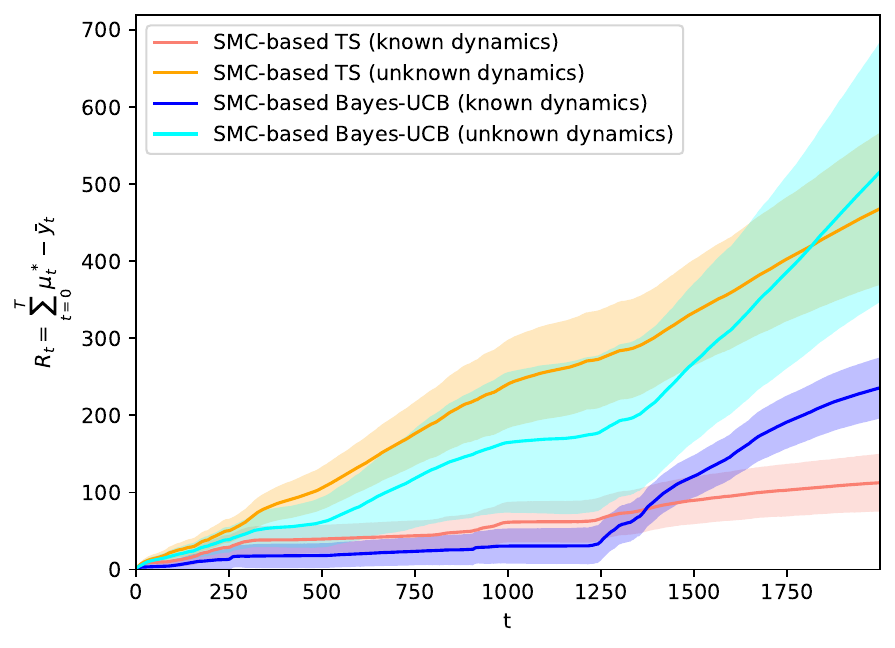}
		\caption{Cumulative regret for SMC-based Bayesian policies in scenario F: known and unknown dynamic parameters.}
		\label{fig:dynamic_bandits_f_softmax_cstatic}%
	\end{subfigure}
	
	\caption{
		Mean regret (standard deviation shown as shaded region) in contextual, non-stationary categorical bandit Scenarios E and F
		described in Equations~\eqref{eq:linear_mixing_dynamics_e}--\eqref{eq:linear_mixing_dynamics_f}.
		Notice how changes in per-arm expected rewards ($t\approx1750$ in Scenario E and $t>1250$ in Scenario F) 
		as illustrated in Figures~\ref{fig:linear_mixing_dynamics_e_softmax}--\ref{fig:linear_mixing_dynamics_f_softmax}
		impact regret
		as showcased in Figures~\ref{fig:dynamic_bandits_e_softmax_cstatic}--\ref{fig:dynamic_bandits_f_softmax_cstatic}.
		SMC-based Bayesian policies adapt to these changes and balance the exploration-exploitation tradeoff.
	}
	\label{fig:dynamic_bandits_softmax}
\vspace*{-2ex}
\end{figure}

These bandit scenarios accommodate a diverse set of expected reward dynamics,
for each realization of the noise processes $\epsilon_{a,c}, \forall a,c$,
and depending on the initialization of parameters $\theta_{0,a}$.
We illustrate per-arm, expected reward time-evolution for a realization
of the two-armed bandit Scenario E in Figure~\ref{fig:linear_mixing_dynamics_e_softmax},
and for the three-armed bandit Scenario F in Figure~\ref{fig:linear_mixing_dynamics_f_softmax}.

In all cases, expected rewards of each arm vary over time,
resulting in transient and recurrent swaps of the optimal arm's identity.
We show the corresponding cumulative regret of SMC-based Bayesian policies
in Figure~\ref{fig:dynamic_bandits_e_softmax_cstatic} for Scenario E,
and in Figure~\ref{fig:dynamic_bandits_f_softmax_cstatic} for Scenario F.

We observe that SMC-based Thompson sampling and Bayes-UCB are able to reach
satisfactory exploitation-exploration balance,
\ie the algorithms dynamically adapt their choice of which arm to play, and attain sublinear cumulative regret.

Recall the linear increase in cumulative regret (\ie exploration)
when latent parameter dynamics result in changes in the optimal arm's identity:
around $t\in (800,1000)$ in Figure~\ref{fig:dynamic_bandits_e_softmax_cstatic},
and around $t\in (1250,1500)$ in Figure~\ref{fig:dynamic_bandits_f_softmax_cstatic}.
After updating the random measure posterior over the unknown latent parameters,
and recomputing the expected rewards per-arm,
SMC-based policies are able to slowly adapt to the optimal arm changes,
reaching a new exploitation-exploration balance, \ie flattening the cumulative regret curves.

For the most interesting and challenging setting where the dynamic model's parameters are unknown,
we observe an increase in cumulative regret for both SMC-based policies.
This is a direct consequence of
the agent sequentially learning all the unknown model parameters,
per-arm $a$, and discrete value $c$: $L_{a,c}, \Sigma_{a,c}, \forall a,c$.
Only when posteriors over these 
---used by the SMC-based agents to propagate uncertainty to each bandit arms' expected reward estimates---
are improved,
can SMC-based policies make informed decisions and attain sublinear regret.

We observe that the impact of expected reward changes,
when occurring later in time
(\eg $t\approx1250$ in Figure~\ref{fig:linear_mixing_dynamics_f_softmax})
is more pronounced for SMC-based Bayes-UCB policies.
Namely, the average cumulative regret of SMC-based Bayes-UCB increases drastically, as well as its volatility,
after $t=1250$ in Figure~\ref{fig:dynamic_bandits_f_softmax_cstatic}.
We hypothesize that this deterioration over time is
due to the shrinking quantile value $\alpha_t\propto1/t$ proposed by \citet{ip-Kaufmann2012}, 
originally designed for stationary bandits.
Confidence bounds for static reward models tend to shrink proportional to the number of observations per bandit arm.
However, in non-stationary regimes, such assumption does not hold:
shrinking $\alpha_t$ over time does not capture the time-evolving parameter posteriors' uncertainty in the long run. 

More generally, the need to determine appropriate quantile values $\alpha_t$
for each reward and non-stationary bandit model is a drawback of Bayes-UCB,
as its optimal value will depend on the specific combination of underlying dynamics and reward function.
On the contrary, Thompson sampling relies on samples from the posterior,
which we here show SMC is able to approximate accurately enough
for SMC-based Thompson sampling to operate successfully in all studied cases,
without any hyperparameter selection.

\subsection{Bandits for personalized news article recommendation}
\label{ssec:logged_data_bandits}
We evaluate the application of SMC-based policies in a real-life application of bandits:
the recommendation of personalized news articles, as previously done by \citet{ic-Chapelle2011}.
%

We use a dataset\footnote{
	Available at \href{https://webscope.sandbox.yahoo.com/catalog.php?datatype=r\&did=49}{R6A - Yahoo! Front Page Today Module User Click Log Dataset.}
} that contains a fraction of user click logs for news articles displayed in the Featured Tab of the Today Module on the Yahoo! Front Page during the first ten days in May 2009. The articles to be displayed were originally chosen uniformly at random from a hand-picked pool of high-quality articles.
From these pool of original candidates,
we pick a subset of 20 articles shown at different times within May 6th,
and collect all user interactions logged with these articles,
for a total of 500,354 events.
In the dataset,
each user is associated with six features:
a bias term and 5 features that correspond to the membership features constructed via the conjoint analysis with a bilinear model described in~\citep{ip-Chu2009}.

The goal is to identify the most interesting article for each user,
or in bandit terms,
to maximize the total number of clicks on the recommended articles over all user interactions,
\ie the average click-through rate (CTR).

We treat each article as a bandit arm ($|\A|=20$),
and define whether the article is clicked or not by the user as a binary reward: $y_t=\{1,0\}$.
Hence, we pose the problem as a MAB with logistic rewards,
where we incorporate the user features as context, $x_t\in \Real^6$.

We implement SMC-based Thompson sampling only, due to the flexibility shown in simulated scenarios,
and its lack of hyperparameter tuning.

We argue that a news recommendation system should evolve over time,
as the relevance of news might change during the course of the day.
We evaluate both stationary and non-stationary bandits with logistic rewards.

\begin{figure}[!h]
	\centering
	\vspace*{-2ex}
	\includegraphics[width=0.85\textwidth]{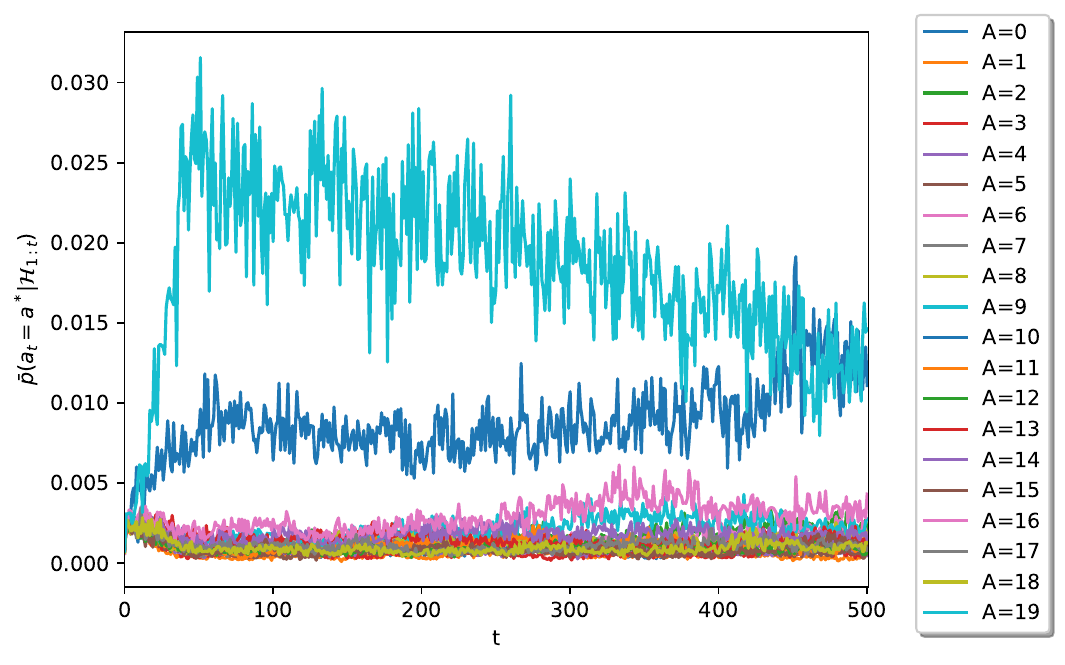}
	\vspace*{-2ex}
	\caption{Empirical probability of playing each bandit arm over time, for SMC-based dynamic logistic Thompson sampling.
		The proposed dynamic bandit policy captures the changing popularity of articles over time.}
	\label{fig:yahoo_logistic_dynamic}
\end{figure}

As shown in Figure~\ref{fig:yahoo_logistic_dynamic},
we observe the flexibility of a non-stationary logistic bandit model,
where we notice how the SMC-based TS agent is able to pick up the dynamic popularity of certain articles over time
---averaged CTR results are provided in Table \ref{tab:yahoo_logistic_crt}.

\begin{table}[!ht]
	\begin{center}
		\resizebox*{\textwidth}{!}{
			\begin{tabular}{*{3}{|c}|}
				\hline
				Model \cellcolor[gray]{0.6} & CTR\cellcolor[gray]{0.6} & Normalized CTR\cellcolor[gray]{0.6} \\ \hline
				\cellcolor[gray]{0.8} Logistic rewards, static arms & 0.0670 +/- 0.0088 & 1.6095 +/- 0.2115  \\ \hline
				\cellcolor[gray]{0.8} Logistic rewards, time-evolving arms & 0.0655 +/- 0.0082 & 1.5745 +/- 0.2064 \\ \hline
			\end{tabular}
		}
		\caption{CTR results for SMC-based policies on the news article recommendation dataset.
			The normalized CTR is with respect to a random recommendation baseline.}
		\label{tab:yahoo_logistic_crt}
	\end{center}
	\vspace*{-2ex}
\end{table}

\section{Conclusion and discussion}
\label{sec:conclusion}
We presented a sequential Monte Carlo (SMC)-based framework for multi-armed bandits (MABs),
where we combine SMC inference with state-of-the-art Bayesian bandit policies.
We extend the applicability of Bayesian MAB policies ---Thompson sampling and Bayes-UCB---
to previously elusive bandit environments,
by accommodating nonlinear and time-varying models of the world,
via SMC-based inference of the sufficient statistics of interest.

The proposed SMC-based Bayesian MAB framework allows for interpretable modeling of nonlinear and time-evolving reward functions,
as it sequentially learns the sufficient statistics and dynamics of the bandit from online data,
to find the right exploration-exploitation balance.
Empirical results show good cumulative regret performance of the proposed policies
in simulated MAB environments that previous algorithms can not address,
and in practical scenarios (personalized news article recommendation)
where time-varying models of data are required.

We show that SMC-based posterior random measures 
are accurate enough for Bayesian bandit policies to find satisfactory exploration-exploitation tradeoffs.
The proposed SMC-based Bayesian agents do not only estimate the evolving latent parameters,
but also quantify how their uncertainty maps to the uncertainty over the optimality of each arm,
adjusting to non-stationary environments.
Careful computation of SMC random measures
is fundamental for the accuracy of the sequential approximation to the posteriors of interest,
and the downstream performance of the proposed SMC-based MAB policies.
The time-varying uncertainty of the sequentially updated parameter posterior encourages exploration of arms
that have not been played recently, but may have reached new exploitable rewards.
Namely, as the posteriors of unobserved arms 
result in broader SMC posteriors,
SMC-based Bayesian MAB policies are more likely to explore such arm,
reduce their posterior's uncertainty, and in turn, update the exploration-exploitation balance.

Important future work remains
on the theoretical understanding of Thompson sampling and Bayes-UCB within the proposed SMC-based MAB framework.
Given that SMC posteriors converge to the true posterior under suitable conditions~\citep{b-Liu2001,j-Crisan2002,j-Chopin2004},
we hypothesize that the proposed SMC-based bandit policies can achieve sub-linear regret, under appropriate assumptions on the latent dynamics.

On the one hand, \citet{ip-Gopalan2014} have shown that a logarithmic regret bound holds for Thompson sampling in complex problems,
for bandits with discretely-supported priors over the parameter space without additional structural properties,
such as conjugate prior structure or independence across arms.
On the other, regret of a non-stationary bandit agent is linear if optimal arm changes occur continuously or adversarially.
However, as long as the bandit's latent dynamics incur in a controlled number of optimal arm changes,
SMC can provide accurate enough posteriors to find the right exploration-exploitation tradeoff, as we show empirically here.

A theoretical analysis of the dependency between
the dynamic bandit model's characteristics,
the resulting rate of optimal arm changes,
and SMC posterior convergence guarantees,
leading to formal regret bounds for the proposed SMC-based Bayesian policies, is an open research direction.

\section*{Acknowledgments}
We thank Luke Bornn for bringing~\citep{j-Cherkassky2013} to our attention,
and the reviewers of previous versions of the work, for their valuable insights and suggestions.
This research was supported in part by NSF grant SCH-1344668.
I\~{n}igo Urteaga acknowledges
this research is supported by ``la Caixa'' foundation fellowship LCF/BQ/PI22/11910028,
and also by the Basque Government through the BERC 2022-2025 program
and by the Ministry of Science and Innovation: BCAM Severo Ochoa accreditation
CEX2021-001142-S / MICIN / AEI / 10.13039/501100011033.


\begin{thebibliography}{103}
\providecommand{\natexlab}[1]{#1}
\providecommand{\url}[1]{\texttt{#1}}
\expandafter\ifx\csname urlstyle\endcsname\relax
  \providecommand{\doi}[1]{doi: #1}\else
  \providecommand{\doi}{doi: \begingroup \urlstyle{rm}\Url}\fi

\bibitem[Abbasi-Yadkori et~al.(2011)Abbasi-Yadkori, P\'{a}l, and
  Szepesv\'{a}ri]{ic-Abbasi-Yadkori2011}
Y.~Abbasi-Yadkori, D.~P\'{a}l, and C.~Szepesv\'{a}ri.
\newblock {Improved Algorithms for Linear Stochastic Bandits}.
\newblock In J.~Shawe-Taylor, R.~S. Zemel, P.~L. Bartlett, F.~Pereira, and
  K.~Q. Weinberger, editors, \emph{Advances in Neural Information Processing
  Systems 24}, pages 2312--2320. Curran Associates, Inc., 2011.
\newblock URL
  \url{https://papers.nips.cc/paper/4417-improved-algorithms-for-linear-stochastic-bandits}.

\bibitem[Agarwal(2013)]{ip-Agarwal2013}
D.~Agarwal.
\newblock {Computational Advertising: The Linkedin Way}.
\newblock In \emph{Proceedings of the 22Nd ACM International Conference on
  Information \& Knowledge Management}, CIKM '13, pages 1585--1586, New York,
  NY, USA, 2013. ACM.
\newblock ISBN 978-1-4503-2263-8.
\newblock \doi{10.1145/2505515.2514690}.
\newblock URL \url{http://doi.acm.org/10.1145/2505515.2514690}.

\bibitem[Agrawal and Goyal(2012)]{ip-Agrawal2012}
S.~Agrawal and N.~Goyal.
\newblock {Analysis of Thompson Sampling for the multi-armed bandit problem}.
\newblock In \emph{{Conference on Learning Theory}}, pages 39--1, 2012.

\bibitem[Agrawal and Goyal(2013{\natexlab{a}})]{ip-Agrawal2013}
S.~Agrawal and N.~Goyal.
\newblock {Further Optimal Regret Bounds for Thompson Sampling}.
\newblock In \emph{{Artificial Intelligence and Statistics}}, pages 99--107,
  2013{\natexlab{a}}.

\bibitem[Agrawal and Goyal(2013{\natexlab{b}})]{ip-Agrawal2013a}
S.~Agrawal and N.~Goyal.
\newblock {Thompson Sampling for Contextual Bandits with Linear Payoffs}.
\newblock In \emph{{International Conference on Machine Learning}}, pages
  127--135, 2013{\natexlab{b}}.

\bibitem[Andrieu et~al.(2010)Andrieu, Doucet, and Holenstein]{j-Andrieu2010}
C.~Andrieu, A.~Doucet, and R.~Holenstein.
\newblock {Particle markov chain monte carlo methods}.
\newblock \emph{Journal of the Royal Statistical Society: Series B (Statistical
  Methodology)}, 72\penalty0 (3):\penalty0 269--342, 2010.

\bibitem[Arulampalam et~al.(2002)Arulampalam, Maskell, Gordon, and
  Clapp]{j-Arulampalam2002}
M.~S. Arulampalam, S.~Maskell, N.~Gordon, and T.~Clapp.
\newblock {A tutorial on particle filters for online nonlinear/non-{G}aussian
  {B}ayesian tracking}.
\newblock \emph{Signal Processing, IEEE Transactions on}, 50\penalty0
  (2):\penalty0 174--188, 2 2002.
\newblock ISSN 1053-587X.

\bibitem[Auer et~al.(2002{\natexlab{a}})Auer, Cesa-Bianchi, and
  Fischer]{j-Auer2002}
P.~Auer, N.~Cesa-Bianchi, and P.~Fischer.
\newblock {Finite-time Analysis of the Multiarmed Bandit Problem}.
\newblock \emph{Machine Learning}, 47\penalty0 (2-3):\penalty0 235--256, May
  2002{\natexlab{a}}.
\newblock ISSN 0885-6125.
\newblock \doi{10.1023/A:1013689704352}.

\bibitem[Auer et~al.(2002{\natexlab{b}})Auer, Cesa-Bianchi, Freund, and
  Schapire]{j-Auer2002a}
P.~Auer, N.~Cesa-Bianchi, Y.~Freund, and R.~E. Schapire.
\newblock {The Nonstochastic Multiarmed Bandit Problem}.
\newblock \emph{SIAM Journal on Computing}, 32\penalty0 (1):\penalty0 48--77,
  2002{\natexlab{b}}.
\newblock \doi{10.1137/S0097539701398375}.
\newblock URL \url{https://doi.org/10.1137/S0097539701398375}.

\bibitem[Bai et~al.(2013)Bai, Wu, and Chen]{ic-Bai2013}
A.~Bai, F.~Wu, and X.~Chen.
\newblock {Bayesian Mixture Modelling and Inference based Thompson Sampling in
  Monte-Carlo Tree Search}.
\newblock In C.~J.~C. Burges, L.~Bottou, M.~Welling, Z.~Ghahramani, and K.~Q.
  Weinberger, editors, \emph{Advances in Neural Information Processing Systems
  26}, pages 1646--1654. Curran Associates, Inc., 2013.
\newblock URL
  \url{https://proceedings.neurips.cc/paper_files/paper/2013/hash/846c260d715e5b854ffad5f70a516c88-Abstract.html}.

\bibitem[Besbes et~al.(2014)Besbes, Gur, and Zeevi]{ic-Besbes2014}
O.~Besbes, Y.~Gur, and A.~Zeevi.
\newblock Stochastic multi-armed-bandit problem with non-stationary rewards.
\newblock In Z.~Ghahramani, M.~Welling, C.~Cortes, N.~D. Lawrence, and K.~Q.
  Weinberger, editors, \emph{Advances in Neural Information Processing Systems
  27}, pages 199--207. Curran Associates, Inc., 2014.
\newblock URL
  \url{http://papers.nips.cc/paper/5378-stochastic-multi-armed-bandit-problem-with-non-stationary-rewards}.

\bibitem[Blundell et~al.(2015)Blundell, Cornebise, Kavukcuoglu, and
  Wierstra]{ip-Blundell2015}
C.~Blundell, J.~Cornebise, K.~Kavukcuoglu, and D.~Wierstra.
\newblock {Weight Uncertainty in Neural Networks}.
\newblock In \emph{Proceedings of the 32Nd International Conference on
  International Conference on Machine Learning - Volume 37}, ICML'15, pages
  1613--1622. JMLR.org, 2015.

\bibitem[Bogunovic et~al.(2016)Bogunovic, Scarlett, and
  Cevher]{ip-Bogunovic2016}
I.~Bogunovic, J.~Scarlett, and V.~Cevher.
\newblock {Time-Varying Gaussian Process Bandit Optimization}.
\newblock In A.~Gretton and C.~C. Robert, editors, \emph{Proceedings of the
  19th International Conference on Artificial Intelligence and Statistics},
  volume~51 of \emph{Proceedings of Machine Learning Research}, pages 314--323,
  Cadiz, Spain, 09--11 May 2016. PMLR.
\newblock URL \url{https://proceedings.mlr.press/v51/bogunovic16.html}.

\bibitem[Box and Jenkins(1976)]{b-Box1976}
G.~Box and G.~Jenkins.
\newblock \emph{{Time Series Analysis: Forecasting and Control}}.
\newblock Holden-Day series in time series analysis and digital processing.
  Holden-Day, 1976.
\newblock ISBN 9780816211043.

\bibitem[Brezzi and Lai(2002)]{j-Brezzi2002}
M.~Brezzi and T.~L. Lai.
\newblock {Optimal learning and experimentation in bandit problems}.
\newblock \emph{Journal of Economic Dynamics and Control}, 27\penalty0
  (1):\penalty0 87 -- 108, 2002.
\newblock ISSN 0165-1889.
\newblock \doi{https://doi.org/10.1016/S0165-1889(01)00028-8}.

\bibitem[Brockwell and Davis(1991)]{b-Brockwell1991}
P.~J. Brockwell and R.~A. Davis.
\newblock \emph{{Time Series: Theory and Methods}}.
\newblock Springer Series in Statistics. Springer, 2nd edition, 1991.
\newblock ISBN 1441903194,9781441903198.

\bibitem[Bubeck and Cesa-Bianchi(2012)]{j-Bubeck2012}
S.~Bubeck and N.~Cesa-Bianchi.
\newblock {Regret analysis of stochastic and nonstochastic multi-armed bandit
  problems}.
\newblock \emph{Foundations and Trends{\textregistered} in Machine Learning},
  5\penalty0 (1):\penalty0 1--122, 2012.

\bibitem[Carvalho et~al.(2010)Carvalho, Johannes, Lopes, and
  Polson]{j-Carvalho2010}
C.~M. Carvalho, M.~S. Johannes, H.~F. Lopes, and N.~G. Polson.
\newblock {Particle Learning and Smoothing}.
\newblock \emph{Statist. Sci.}, 25\penalty0 (1):\penalty0 88--106, 02 2010.

\bibitem[Chapelle and Li(2011)]{ic-Chapelle2011}
O.~Chapelle and L.~Li.
\newblock {An Empirical Evaluation of Thompson Sampling}.
\newblock In J.~Shawe-Taylor, R.~S. Zemel, P.~L. Bartlett, F.~Pereira, and
  K.~Q. Weinberger, editors, \emph{Advances in Neural Information Processing
  Systems 24}, pages 2249--2257. Curran Associates, Inc., 2011.
\newblock URL
  \url{https://papers.nips.cc/paper/4321-an-empirical-evaluation-of-thompson-sampling}.

\bibitem[Cherkassky and Bornn(2013)]{j-Cherkassky2013}
M.~Cherkassky and L.~Bornn.
\newblock {Sequential Monte Carlo Bandits}.
\newblock \emph{ArXiv e-prints}, Oct. 2013.

\bibitem[Chopin(2004)]{j-Chopin2004}
N.~Chopin.
\newblock {Central Limit Theorem for Sequential Monte Carlo Methods and Its
  Application to Bayesian Inference}.
\newblock \emph{The Annals of Statistics}, 32\penalty0 (6):\penalty0
  2385--2411, 2004.
\newblock ISSN 00905364.

\bibitem[Chopin et~al.(2011)Chopin, Jacob, and Papaspiliopoulos]{j-Chopin2011}
N.~Chopin, P.~E. Jacob, and O.~Papaspiliopoulos.
\newblock {SMC$^2$: an efficient algorithm for sequential analysis of
  state-space models}.
\newblock \emph{arXiv preprint arXiv:1101.1528}, Jan. 2011.
\newblock URL \url{https://arxiv.org/abs/1101.1528}.

\bibitem[Chu et~al.(2009)Chu, Park, Beaupre, Motgi, Phadke, Chakraborty, and
  Zachariah]{ip-Chu2009}
W.~Chu, S.-T. Park, T.~Beaupre, N.~Motgi, A.~Phadke, S.~Chakraborty, and
  J.~Zachariah.
\newblock {A Case Study of Behavior-driven Conjoint Analysis on Yahoo!: Front
  Page Today Module}.
\newblock In \emph{Proceedings of the 15th ACM SIGKDD International Conference
  on Knowledge Discovery and Data Mining}, KDD '09, pages 1097--1104, New York,
  NY, USA, 2009. ACM.
\newblock ISBN 978-1-60558-495-9.
\newblock \doi{10.1145/1557019.1557138}.

\bibitem[Chu et~al.(2011)Chu, Li, Reyzin, and Schapire]{ip-Chu2011}
W.~Chu, L.~Li, L.~Reyzin, and R.~Schapire.
\newblock {Contextual Bandits with Linear Payoff Functions}.
\newblock In G.~Gordon, D.~Dunson, and M.~Dud\'ik, editors, \emph{Proceedings
  of the Fourteenth International Conference on Artificial Intelligence and
  Statistics}, volume~15 of \emph{Proceedings of Machine Learning Research},
  pages 208--214, Fort Lauderdale, FL, USA, 11--13 Apr 2011. PMLR.
\newblock URL \url{http://proceedings.mlr.press/v15/chu11a.html}.

\bibitem[Creal(2012)]{j-Creal2012}
D.~Creal.
\newblock {A Survey of Sequential Monte Carlo Methods for Economics and
  Finance}.
\newblock \emph{Econometric Reviews}, 31\penalty0 (3):\penalty0 245--296, 2012.

\bibitem[Crisan and Doucet(2002)]{j-Crisan2002}
D.~Crisan and A.~Doucet.
\newblock {A survey of convergence results on particle filtering methods for
  practitioners}.
\newblock \emph{IEEE Transactions on Signal Processing}, 50\penalty0
  (3):\penalty0 736--746, Mar 2002.
\newblock ISSN 1053-587X.
\newblock \doi{10.1109/78.984773}.

\bibitem[Crisan and M\'{i}guez(2013)]{j-Crisan2013}
D.~Crisan and J.~M\'{i}guez.
\newblock {Nested particle filters for online parameter estimation in
  discrete-time state-space Markov models}.
\newblock \emph{arXiv preprint arXiv:1308.1883}, Aug 2013.
\newblock URL \url{https://arxiv.org/abs/1308.1883}.

\bibitem[Djuri\'{c} and Bugallo(2010)]{ib-Djuric2010}
P.~M. Djuri\'{c} and M.~F. Bugallo.
\newblock \emph{{Particle Filtering}}, chapter~5, pages 271--331.
\newblock Wiley-Blackwell, 2010.
\newblock ISBN 9780470575758.
\newblock \doi{10.1002/9780470575758.ch5}.

\bibitem[Djuri\'{c} et~al.(2003)Djuri\'{c}, Kotecha, Zhang, Huang, Ghirmai,
  Bugallo, and M\'{i}guez]{j-Djuric2003}
P.~M. Djuri\'{c}, J.~H. Kotecha, J.~Zhang, Y.~Huang, T.~Ghirmai, M.~F. Bugallo,
  and J.~M\'{i}guez.
\newblock {Particle Filtering}.
\newblock \emph{IEEE Signal Processing Magazine}, 20(5):\penalty0 19--38, 9
  2003.

\bibitem[Djuri\'{c} et~al.(2004)Djuri\'{c}, Bugallo, and
  M\'{i}guez]{ip-Djuric2004}
P.~M. Djuri\'{c}, M.~F. Bugallo, and J.~M\'{i}guez.
\newblock {Density assisted particle filters for state and parameter
  estimation}.
\newblock In \emph{2004 IEEE International Conference on Acoustics, Speech, and
  Signal Processing, (ICASSP)}, volume~2, pages ii -- 701--704, 5 2004.
\newblock \doi{10.1109/ICASSP.2004.1326354}.

\bibitem[Dong et~al.(2019)Dong, Ma, and Roy]{ip-Dong2019}
S.~Dong, T.~Ma, and B.~V. Roy.
\newblock {On the Performance of Thompson Sampling on Logistic Bandits}.
\newblock In A.~Beygelzimer and D.~Hsu, editors, \emph{Proceedings of the
  Thirty-Second Conference on Learning Theory}, volume~99 of \emph{Proceedings
  of Machine Learning Research}, pages 1158--1160. PMLR, 25--28 Jun 2019.
\newblock URL \url{https://proceedings.mlr.press/v99/dong19a.html}.

\bibitem[Doucet et~al.(2000)Doucet, de~Freitas, Murphy, and
  Russell]{ip-Doucet2000}
A.~Doucet, N.~de~Freitas, K.~P. Murphy, and S.~J. Russell.
\newblock {Rao-Blackwellised Particle Filtering for Dynamic Bayesian Networks}.
\newblock In \emph{Proceedings of the 16th Conference on Uncertainty in
  Artificial Intelligence}, UAI '00, pages 176--183, San Francisco, CA, USA,
  2000. Morgan Kaufmann Publishers Inc.
\newblock ISBN 1-55860-709-9.

\bibitem[Doucet et~al.(2001)Doucet, Freitas, and Gordon]{b-Doucet2001}
A.~Doucet, N.~D. Freitas, and N.~Gordon, editors.
\newblock \emph{{Sequential Monte Carlo Methods in Practice}}.
\newblock Springer, 2001.

\bibitem[Dumitrascu et~al.(2018)Dumitrascu, Feng, and
  Engelhardt]{ic-Dumitrascu2018}
B.~Dumitrascu, K.~Feng, and B.~Engelhardt.
\newblock {PG-TS: Improved Thompson Sampling for Logistic Contextual Bandits}.
\newblock In S.~Bengio, H.~Wallach, H.~Larochelle, K.~Grauman, N.~Cesa-Bianchi,
  and R.~Garnett, editors, \emph{Advances in Neural Information Processing
  Systems 31}, pages 4629--4638. Curran Associates, Inc., 2018.
\newblock URL
  \url{https://proceedings.neurips.cc/paper_files/paper/2018/hash/ce6c92303f38d297e263c7180f03d402-Abstract.html}.

\bibitem[Durbin and Koopman(2001)]{b-Durbin2001}
J.~Durbin and S.~J. Koopman.
\newblock \emph{{Time Series Analysis by State-Space Methods}}.
\newblock Oxford Statistical Science Series. Oxford University Press, 2001.

\bibitem[Durbin and Koopman(2012)]{b-Durbin2012}
J.~Durbin and S.~J. Koopman.
\newblock \emph{{Time Series Analysis by State-Space Methods: Second Edition}}.
\newblock Oxford Statistical Science Series. Oxford University Press, 2
  edition, 2012.
\newblock ISBN 9780199641178.

\bibitem[Eckles and Kaptein(2019)]{j-Eckles2019}
D.~Eckles and M.~Kaptein.
\newblock {Bootstrap Thompson Sampling and Sequential Decision Problems in the
  Behavioral Sciences}.
\newblock \emph{SAGE Open}, 9\penalty0 (2):\penalty0 2158244019851675, 2019.
\newblock \doi{10.1177/2158244019851675}.
\newblock URL \url{https://doi.org/10.1177/2158244019851675}.

\bibitem[Faury et~al.(2020)Faury, Abeille, Calauz{\`e}nes, and
  Fercoq]{ip-Faury2020}
L.~Faury, M.~Abeille, C.~Calauz{\`e}nes, and O.~Fercoq.
\newblock {Improved optimistic algorithms for logistic bandits}.
\newblock In \emph{International Conference on Machine Learning}, pages
  3052--3060. PMLR, 2020.

\bibitem[Ferreira et~al.(2018)Ferreira, Simchi-Levi, and Wang]{j-Ferreira2018}
K.~J. Ferreira, D.~Simchi-Levi, and H.~Wang.
\newblock {Online Network Revenue Management Using Thompson Sampling}.
\newblock \emph{Operations Research}, 66\penalty0 (6):\penalty0 1586--1602,
  2018.
\newblock \doi{10.1287/opre.2018.1755}.
\newblock URL \url{https://doi.org/10.1287/opre.2018.1755}.

\bibitem[Garivier and Capp\'{e}(2011)]{ip-Garivier2011a}
A.~Garivier and O.~Capp\'{e}.
\newblock {The KL-UCB Algorithm for Bounded Stochastic Bandits and Beyond}.
\newblock In S.~M. Kakade and U.~von Luxburg, editors, \emph{Proceedings of the
  24th Annual Conference on Learning Theory}, volume~19 of \emph{Proceedings of
  Machine Learning Research}, pages 359--376, Budapest, Hungary, 09--11 Jun
  2011. PMLR.
\newblock URL \url{http://proceedings.mlr.press/v19/garivier11a.html}.

\bibitem[Garivier and Moulines(2011)]{ip-Garivier2011}
A.~Garivier and E.~Moulines.
\newblock On upper-confidence bound policies for switching bandit problems.
\newblock In \emph{Proceedings of the 22Nd International Conference on
  Algorithmic Learning Theory}, ALT'11, pages 174--188, Berlin, Heidelberg,
  2011. Springer-Verlag.
\newblock ISBN 978-3-642-24411-7.
\newblock URL \url{http://dl.acm.org/citation.cfm?id=2050345.2050365}.

\bibitem[Geisser(1965)]{j-Geisser1965}
S.~Geisser.
\newblock {Bayesian-Estimation in multivariate analysis}.
\newblock \emph{Annals of American Statistics}, 36\penalty0 (1):\penalty0
  150--159, 1965.
\newblock ISSN {0003-4851}.
\newblock \doi{10.1214/aoms/1177700279}.

\bibitem[Geisser and Cornfield(1963)]{j-Geisser1963}
S.~Geisser and J.~Cornfield.
\newblock {Posterior Distributions for Multivariate Normal Parameters}.
\newblock \emph{Journal of the Royal Statistical Society. Series B
  (Methodological)}, 25\penalty0 (2):\penalty0 368--376, 1963.
\newblock ISSN 00359246.
\newblock URL \url{http://www.jstor.org/stable/2984304}.

\bibitem[Gittins(1979)]{j-Gittins1979}
J.~C. Gittins.
\newblock {Bandit Processes and Dynamic Allocation Indices}.
\newblock \emph{Journal of the Royal Statistical Society. Series B
  (Methodological)}, 41\penalty0 (2):\penalty0 148--177, 1979.
\newblock ISSN 00359246.

\bibitem[Gopalan et~al.(2014)Gopalan, Mannor, and Mansour]{ip-Gopalan2014}
A.~Gopalan, S.~Mannor, and Y.~Mansour.
\newblock {Thompson Sampling for Complex Online Problems}.
\newblock In E.~P. Xing and T.~Jebara, editors, \emph{Proceedings of the 31st
  International Conference on Machine Learning}, volume~32 of \emph{Proceedings
  of Machine Learning Research}, pages 100--108, Bejing, China, 22--24 Jun
  2014. PMLR.
\newblock URL \url{http://proceedings.mlr.press/v32/gopalan14.html}.

\bibitem[Gordon et~al.(1993)Gordon, Salmond, and Smith]{j-Gordon1993}
N.~J. Gordon, D.~Salmond, and A.~Smith.
\newblock {Novel approach to nonlinear/non-Gaussian Bayesian state estimation}.
\newblock \emph{Radar and Signal Processing, IEEE Proceedings}, 140\penalty0
  (2):\penalty0 107 --113, 4 1993.
\newblock ISSN 0956-375X.

\bibitem[Hill et~al.(2017)Hill, Nassif, Liu, Iyer, and
  Vishwanathan]{ip-Hill2017}
D.~N. Hill, H.~Nassif, Y.~Liu, A.~Iyer, and S.~Vishwanathan.
\newblock {An efficient bandit algorithm for realtime multivariate
  optimization}.
\newblock In \emph{Proceedings of the 23rd ACM SIGKDD International Conference
  on Knowledge Discovery and Data Mining}, pages 1813--1821. ACM, 2017.

\bibitem[Ionides et~al.(2006)Ionides, Bret\'{o}, and King]{j-Ionides2006}
E.~L. Ionides, C.~Bret\'{o}, and A.~A. King.
\newblock {Inference for nonlinear dynamical systems}.
\newblock \emph{Proceedings of the National Academy of Sciences}, 103\penalty0
  (49):\penalty0 18438--18443, 2006.

\bibitem[Jung and Tewari(2019)]{ip-Jung2019}
Y.~H. Jung and A.~Tewari.
\newblock {Regret Bounds for Thompson Sampling in Episodic Restless Bandit
  Problems}.
\newblock In H.~Wallach, H.~Larochelle, A.~Beygelzimer, F.~d\textquotesingle
  Alch\'{e}-Buc, E.~Fox, and R.~Garnett, editors, \emph{Advances in Neural
  Information Processing Systems}, volume~32. Curran Associates, Inc., 2019.
\newblock URL
  \url{https://proceedings.neurips.cc/paper_files/paper/2019/hash/2edfeadfe636973b42d7b6ac315b896c-Abstract.html}.

\bibitem[Jung et~al.(2019)Jung, Abeille, and Tewari]{j-Jung2019}
Y.~H. Jung, M.~Abeille, and A.~Tewari.
\newblock {Thompson sampling in non-episodic restless bandits}.
\newblock \emph{arXiv preprint arXiv:1910.05654}, 2019.
\newblock URL \url{https://arxiv.org/abs/1910.05654}.

\bibitem[Kalman(1960)]{j-Kalman1960}
R.~E. Kalman.
\newblock {A New Approach to Linear Filtering and Prediction Problems}.
\newblock \emph{Transactions of the ASME--Journal of Basic Engineering},
  82\penalty0 (Series D):\penalty0 35--45, 1960.

\bibitem[Kandasamy et~al.(2018)Kandasamy, Krishnamurthy, Schneider, and
  Poczos]{ip-Kandasamy2018}
K.~Kandasamy, A.~Krishnamurthy, J.~Schneider, and B.~Poczos.
\newblock {Parallelised Bayesian Optimisation via Thompson Sampling}.
\newblock In A.~Storkey and F.~Perez-Cruz, editors, \emph{Proceedings of the
  Twenty-First International Conference on Artificial Intelligence and
  Statistics}, volume~84 of \emph{Proceedings of Machine Learning Research},
  pages 133--142, Playa Blanca, Lanzarote, Canary Islands, 09--11 Apr 2018.
  PMLR.
\newblock URL \url{http://proceedings.mlr.press/v84/kandasamy18a.html}.

\bibitem[Kantas et~al.(2015)Kantas, Doucet, Singh, Maciejowski, and
  Chopin]{j-Kantas2015}
N.~Kantas, A.~Doucet, S.~S. Singh, J.~Maciejowski, and N.~Chopin.
\newblock {On particle methods for parameter estimation in state-space models}.
\newblock \emph{Statistical science}, 30\penalty0 (3):\penalty0 328--351, 2015.

\bibitem[Kaufmann et~al.(2012)Kaufmann, Cappe, and Garivier]{ip-Kaufmann2012}
E.~Kaufmann, O.~Cappe, and A.~Garivier.
\newblock {On Bayesian Upper Confidence Bounds for Bandit Problems}.
\newblock In N.~D. Lawrence and M.~Girolami, editors, \emph{Proceedings of the
  Fifteenth International Conference on Artificial Intelligence and
  Statistics}, volume~22 of \emph{Proceedings of Machine Learning Research},
  pages 592--600, La Palma, Canary Islands, 21--23 Apr 2012. PMLR.

\bibitem[Kawale et~al.(2015)Kawale, Bui, Kveton, Tran-Thanh, and
  Chawla]{ic-Kawale2015}
J.~Kawale, H.~H. Bui, B.~Kveton, L.~Tran-Thanh, and S.~Chawla.
\newblock {Efficient Thompson Sampling for Online Matrix-Factorization
  Recommendation}.
\newblock In C.~Cortes, N.~D. Lawrence, D.~D. Lee, M.~Sugiyama, and R.~Garnett,
  editors, \emph{Advances in Neural Information Processing Systems 28}, pages
  1297--1305. Curran Associates, Inc., 2015.
\newblock URL
  \url{https://proceedings.neurips.cc/paper_files/paper/2015/hash/846c260d715e5b854ffad5f70a516c88-Abstract.html}.

\bibitem[Kingma et~al.(2015)Kingma, Salimans, and Welling]{ic-Kingma2015}
D.~P. Kingma, T.~Salimans, and M.~Welling.
\newblock {Variational Dropout and the Local Reparameterization Trick}.
\newblock In C.~Cortes, N.~D. Lawrence, D.~D. Lee, M.~Sugiyama, and R.~Garnett,
  editors, \emph{Advances in Neural Information Processing Systems 28}, pages
  2575--2583. Curran Associates, Inc., 2015.

\bibitem[Korda et~al.(2013)Korda, Kaufmann, and Munos]{ic-Korda2013}
N.~Korda, E.~Kaufmann, and R.~Munos.
\newblock {Thompson Sampling for 1-Dimensional Exponential Family Bandits}.
\newblock In C.~J.~C. Burges, L.~Bottou, M.~Welling, Z.~Ghahramani, and K.~Q.
  Weinberger, editors, \emph{Advances in Neural Information Processing Systems
  26}, pages 1448--1456. Curran Associates, Inc., 2013.

\bibitem[Lai(1987)]{j-Lai1987}
T.~L. Lai.
\newblock {Adaptive Treatment Allocation and the Multi-Armed Bandit Problem}.
\newblock \emph{The Annals of Statistics}, 15\penalty0 (3):\penalty0
  1091--1114, 1987.
\newblock ISSN 00905364.

\bibitem[Lai and Robbins(1985)]{j-Lai1985}
T.~L. Lai and H.~Robbins.
\newblock {Asymptotically Efficient Adaptive Allocation Rules}.
\newblock \emph{Advances in Applied Mathematics}, 6\penalty0 (1):\penalty0
  4--22, mar 1985.
\newblock ISSN 0196-8858.
\newblock \doi{10.1016/0196-8858(85)90002-8}.

\bibitem[Lattimore and Szepesv{\'a}ri(2020)]{b-Lattimore2020}
T.~Lattimore and C.~Szepesv{\'a}ri.
\newblock \emph{Bandit algorithms}.
\newblock Cambridge University Press, 2020.

\bibitem[Li et~al.(2016)Li, Chen, Carlson, and Carin]{ip-Li2016}
C.~Li, C.~Chen, D.~Carlson, and L.~Carin.
\newblock {Preconditioned Stochastic Gradient Langevin Dynamics for Deep Neural
  Networks}.
\newblock In \emph{Proceedings of the Thirtieth AAAI Conference on Artificial
  Intelligence}, AAAI'16, pages 1788--1794. AAAI Press, 2016.

\bibitem[Li(2013)]{j-Li2013}
L.~Li.
\newblock {Generalized Thompson Sampling for Contextual Bandits}.
\newblock \emph{arXiv preprint arXiv:1310.7163}, Oct. 2013.
\newblock URL \url{https://arxiv.org/abs/1310.7163}.

\bibitem[Li et~al.(2010)Li, Chu, Langford, and Schapire]{ip-Li2010}
L.~Li, W.~Chu, J.~Langford, and R.~E. Schapire.
\newblock {A Contextual-Bandit Approach to Personalized News Article
  Recommendation}.
\newblock In \emph{Proceedings of the 19th international conference on World
  wide web}, volume abs/1003.0146, pages 661--670, 2010.

\bibitem[Li et~al.(2015)Li, Boli\'{c}, and Djuri\'{c}]{j-Li2015}
T.~Li, M.~Boli\'{c}, and P.~M. Djuri\'{c}.
\newblock {Resampling Methods for Particle Filtering: Classification,
  implementation, and strategies}.
\newblock \emph{Signal Processing Magazine, IEEE}, 32\penalty0 (3):\penalty0
  70--86, 5 2015.
\newblock ISSN 1053-5888.

\bibitem[Liu and West(2001)]{ib-Liu2001}
J.~Liu and M.~West.
\newblock \emph{{Combined Parameter and State Estimation in Simulation-Based
  Filtering}}, chapter~10, pages 197--223.
\newblock Springer New York, New York, NY, 2001.
\newblock ISBN 978-1-4757-3437-9.
\newblock \doi{10.1007/978-1-4757-3437-9_10}.

\bibitem[Liu(2001)]{b-Liu2001}
J.~S. Liu.
\newblock \emph{{Monte Carlo Strategies in Scientific Computing}}.
\newblock Springer Series in Statistics. Springer, 2001.

\bibitem[Lu and Roy(2017)]{ip-Lu2017}
X.~Lu and B.~V. Roy.
\newblock {Ensemble sampling}.
\newblock In \emph{Advances in Neural Information Processing Systems}, pages
  3258--3266, 2017.

\bibitem[Luo et~al.(2018)Luo, Wei, Agarwal, and Langford]{ip-Luo2018}
H.~Luo, C.-Y. Wei, A.~Agarwal, and J.~Langford.
\newblock {Efficient Contextual Bandits in Non-stationary Worlds}.
\newblock In S.~Bubeck, V.~Perchet, and P.~Rigollet, editors, \emph{Proceedings
  of the 31st Conference On Learning Theory}, volume~75 of \emph{Proceedings of
  Machine Learning Research}, pages 1739--1776. PMLR, 06--09 Jul 2018.
\newblock URL \url{https://proceedings.mlr.press/v75/luo18a.html}.

\bibitem[Maiz et~al.(2012)Maiz, Molanes-Lopez, Miguez, and Djuric]{j-Maiz2012}
C.~S. Maiz, E.~M. Molanes-Lopez, J.~Miguez, and P.~M. Djuric.
\newblock {A Particle Filtering Scheme for Processing Time Series Corrupted by
  Outliers}.
\newblock \emph{IEEE Transactions on Signal Processing}, 60\penalty0
  (9):\penalty0 4611--4627, 2012.
\newblock \doi{10.1109/TSP.2012.2200480}.

\bibitem[Mellor and Shapiro(2013)]{ip-Mellor2013}
J.~Mellor and J.~Shapiro.
\newblock {Thompson Sampling in Switching Environments with Bayesian Online
  Change Detection}.
\newblock In C.~M. Carvalho and P.~Ravikumar, editors, \emph{Proceedings of the
  Sixteenth International Conference on Artificial Intelligence and
  Statistics}, volume~31 of \emph{Proceedings of Machine Learning Research},
  pages 442--450, Scottsdale, Arizona, USA, 29 Apr--01 May 2013. PMLR.
\newblock URL \url{http://proceedings.mlr.press/v31/mellor13a.html}.

\bibitem[Merwe et~al.(2001)Merwe, Doucet, Freitas, and Wan]{ip-Merwe2001}
R.~V.~D. Merwe, A.~Doucet, N.~D. Freitas, and E.~A. Wan.
\newblock {The unscented particle filter}.
\newblock In \emph{Advances in neural information processing systems}, pages
  584--590, 2001.

\bibitem[Mnih et~al.(2015)Mnih, Kavukcuoglu, Silver, Rusu, Veness, Bellemare,
  Graves, Riedmiller, Fidjeland, Ostrovski, Petersen, Beattie, Sadik,
  Antonoglou, King, Kumaran, Wierstra, Legg, and Hassabis]{j-Mnih2015}
V.~Mnih, K.~Kavukcuoglu, D.~Silver, A.~A. Rusu, J.~Veness, M.~G. Bellemare,
  A.~Graves, M.~Riedmiller, A.~K. Fidjeland, G.~Ostrovski, S.~Petersen,
  C.~Beattie, A.~Sadik, I.~Antonoglou, H.~King, D.~Kumaran, D.~Wierstra,
  S.~Legg, and D.~Hassabis.
\newblock Human-level control through deep reinforcement learning.
\newblock \emph{nature}, 518\penalty0 (7540):\penalty0 529--533, 2015.

\bibitem[Netflix(2017)]{Netflix2017}
Netflix.
\newblock {Artwork Personalization at Netflix}.
\newblock
  {\href{https://medium.com/netflix-techblog/artwork-personalization-c589f074ad76}{medium.com}},
  December 2017.

\bibitem[{Olsson} and {Westerborn}(2014)]{j-Olsson2014}
J.~{Olsson} and J.~{Westerborn}.
\newblock {Efficient particle-based online smoothing in general hidden Markov
  models: the PaRIS algorithm}.
\newblock \emph{arXiv preprint arXiv:1412.7550}, Dec 2014.
\newblock URL \url{https://arxiv.org/abs/1412.7550}.

\bibitem[{Olsson} et~al.(2006){Olsson}, {Capp{\'e}}, {Douc}, and
  {Moulines}]{j-Olsson2006}
J.~{Olsson}, O.~{Capp{\'e}}, R.~{Douc}, and E.~{Moulines}.
\newblock {Sequential Monte Carlo smoothing with application to parameter
  estimation in non-linear state space models}.
\newblock \emph{arXiv preprint arXiv:math/0609514}, Sep 2006.
\newblock URL \url{https://arxiv.org/abs/math/0609514}.

\bibitem[Osband and Roy(2015)]{j-Osband2015}
I.~Osband and B.~V. Roy.
\newblock {Bootstrapped Thompson sampling and deep exploration}.
\newblock \emph{arXiv preprint arXiv:1507.00300}, 2015.
\newblock URL \url{https://arxiv.org/abs/1507.00300}.

\bibitem[Osband et~al.(2016)Osband, Blundell, Pritzel, and Roy]{ic-Osband2016}
I.~Osband, C.~Blundell, A.~Pritzel, and B.~V. Roy.
\newblock {Deep Exploration via Bootstrapped DQN}.
\newblock In D.~D. Lee, M.~Sugiyama, U.~V. Luxburg, I.~Guyon, and R.~Garnett,
  editors, \emph{Advances in Neural Information Processing Systems 29}, pages
  4026--4034. Curran Associates, Inc., 2016.

\bibitem[Raj and Kalyani(2017)]{j-Raj2017}
V.~Raj and S.~Kalyani.
\newblock Taming non-stationary bandits: A bayesian approach.
\newblock \emph{arXiv preprint arXiv:1707.09727}, 2017.
\newblock URL \url{https://arxiv.org/abs/1707.09727}.

\bibitem[Riquelme et~al.(2018)Riquelme, Tucker, and Snoek]{ip-Riquelme2018}
C.~Riquelme, G.~Tucker, and J.~Snoek.
\newblock {Deep Bayesian Bandits Showdown: An Empirical Comparison of Bayesian
  Deep Networks for Thompson Sampling}.
\newblock In \emph{International Conference on Learning Representations}, 2018.

\bibitem[Ristic et~al.(2004)Ristic, Arulampalam, and Gordon]{b-Ristic2004}
B.~Ristic, S.~Arulampalam, and N.~Gordon.
\newblock \emph{{Beyond the Kalman Filter: Particle Filters for Tracking
  Applications}}.
\newblock Artech House, 2004.
\newblock ISBN 9781580538510.

\bibitem[Robbins(1952)]{j-Robbins1952}
H.~Robbins.
\newblock {Some aspects of the sequential design of experiments}.
\newblock \emph{Bulletin of the American Mathematical Society}, \penalty0
  (58):\penalty0 527--535, 1952.

\bibitem[Russo and Roy(2014)]{j-Russo2014}
D.~Russo and B.~V. Roy.
\newblock {Learning to optimize via posterior sampling}.
\newblock \emph{Mathematics of Operations Research}, 39\penalty0 (4):\penalty0
  1221--1243, 2014.

\bibitem[Russo and Roy(2016)]{j-Russo2016}
D.~Russo and B.~V. Roy.
\newblock {An information-theoretic analysis of Thompson sampling}.
\newblock \emph{The Journal of Machine Learning Research}, 17\penalty0
  (1):\penalty0 2442--2471, 2016.

\bibitem[Russo et~al.(2018)Russo, Roy, Kazerouni, Osband, and Wen]{j-Russo2018}
D.~J. Russo, B.~V. Roy, A.~Kazerouni, I.~Osband, and Z.~Wen.
\newblock {A Tutorial on Thompson Sampling}.
\newblock \emph{{Foundations and Trends\textsuperscript{\textregistered} in
  Machine Learning}}, 11\penalty0 (1):\penalty0 1--96, 2018.
\newblock ISSN 1935-8237.
\newblock \doi{10.1561/2200000070}.
\newblock URL \url{http://dx.doi.org/10.1561/2200000070}.

\bibitem[Schwartz et~al.(2017)Schwartz, Bradlow, and Fader]{j-Schwartz2017}
E.~M. Schwartz, E.~T. Bradlow, and P.~S. Fader.
\newblock {Customer Acquisition via Display Advertising Using Multi-Armed
  Bandit Experiments}.
\newblock \emph{Marketing Science}, 36\penalty0 (4):\penalty0 500--522, 2017.
\newblock \doi{10.1287/mksc.2016.1023}.
\newblock URL \url{https://doi.org/10.1287/mksc.2016.1023}.

\bibitem[Scott(2015)]{j-Scott2015}
S.~L. Scott.
\newblock {Multi-armed bandit experiments in the online service economy}.
\newblock \emph{Applied Stochastic Models in Business and Industry},
  31:\penalty0 37--49, 2015.
\newblock Special issue on actual impact and future perspectives on stochastic
  modelling in business and industry.

\bibitem[Shumway and Stoffer(2010)]{b-Shumway2010}
R.~H. Shumway and D.~S. Stoffer.
\newblock \emph{{Time Series Analysis and Its Applications: With R Examples
  (Springer Texts in Statistics)}}.
\newblock Springer, 3rd edition, 2010.
\newblock ISBN 144197864X,9781441978646.

\bibitem[Silver et~al.(2017)Silver, Schrittwieser, Simonyan, Antonoglou, Huang,
  Guez, Hubert, Baker, Lai, Bolton, Chen, Lillicrap, Hui, Sifre, van~den
  Driessche, Graepel, and Hassabis]{j-Silver2017}
D.~Silver, J.~Schrittwieser, K.~Simonyan, I.~Antonoglou, A.~Huang, A.~Guez,
  T.~Hubert, L.~Baker, M.~Lai, A.~Bolton, Y.~Chen, T.~Lillicrap, F.~Hui,
  L.~Sifre, G.~van~den Driessche, T.~Graepel, and D.~Hassabis.
\newblock {Mastering the game of go without human knowledge}.
\newblock \emph{nature}, 550\penalty0 (7676):\penalty0 354--359, 2017.

\bibitem[Slivkins and Upfal(2008)]{ip-Slivkins2008}
A.~Slivkins and E.~Upfal.
\newblock {Adapting to a Changing Environment: the Brownian Restless Bandits}.
\newblock In \emph{COLT}, pages 343--354, 2008.

\bibitem[Sutton and Barto(1998)]{b-Sutton1998}
R.~S. Sutton and A.~G. Barto.
\newblock \emph{{Reinforcement Learning: An Introduction}}.
\newblock MIT Press: Cambridge, MA, 1998.

\bibitem[Thompson(1935)]{j-Thompson1935}
W.~R. Thompson.
\newblock {On the Theory of Apportionment}.
\newblock \emph{American Journal of Mathematics}, 57\penalty0 (2):\penalty0
  450--456, 1935.
\newblock ISSN 00029327, 10806377.

\bibitem[Tiao and Zellner(1964)]{j-Tiao1964}
G.~C. Tiao and A.~Zellner.
\newblock {On the Bayesian Estimation of Multivariate Regression}.
\newblock \emph{Journal of the Royal Statistical Society. Series B
  (Methodological)}, 26\penalty0 (2):\penalty0 277--285, 1964.
\newblock ISSN 00359246.
\newblock URL \url{http://www.jstor.org/stable/2984424}.

\bibitem[Urteaga and Djuri\'{c}(2016{\natexlab{a}})]{j-Urteaga2016}
I.~Urteaga and P.~M. Djuri\'{c}.
\newblock {Sequential Estimation of Hidden {ARMA} Processes by Particle
  Filtering - {P}art {I}}.
\newblock \emph{IEEE Transactions on Signal Processing}, PP\penalty0
  (99):\penalty0 1--1, 2016{\natexlab{a}}.
\newblock ISSN 1053-587X.
\newblock \doi{10.1109/TSP.2016.2598309}.

\bibitem[Urteaga and Djuri\'{c}(2016{\natexlab{b}})]{j-Urteaga2016a}
I.~Urteaga and P.~M. Djuri\'{c}.
\newblock {Sequential Estimation of Hidden {ARMA} Processes by Particle
  Filtering - {P}art {II}}.
\newblock \emph{IEEE Transactions on Signal Processing}, PP\penalty0
  (99):\penalty0 1--1, 2016{\natexlab{b}}.
\newblock ISSN 1053-587X.
\newblock \doi{10.1109/TSP.2016.2598324}.

\bibitem[Urteaga and Wiggins(2018{\natexlab{a}})]{ip-Urteaga2018}
I.~Urteaga and C.~Wiggins.
\newblock {Variational inference for the multi-armed contextual bandit}.
\newblock In A.~Storkey and F.~Perez-Cruz, editors, \emph{Proceedings of the
  Twenty-First International Conference on Artificial Intelligence and
  Statistics}, volume~84 of \emph{Proceedings of Machine Learning Research},
  pages 698--706, Playa Blanca, Lanzarote, Canary Islands, 09--11 Apr
  2018{\natexlab{a}}. PMLR.

\bibitem[Urteaga and Wiggins(2018{\natexlab{b}})]{j-Urteaga2018a}
I.~Urteaga and C.~Wiggins.
\newblock {Nonparametric Gaussian mixture models for the multi-armed contextual
  bandit}.
\newblock \emph{arXiv preprint arXiv:1808.02932}, Sept. 2018{\natexlab{b}}.
\newblock URL \url{https://arxiv.org/abs/1808.02932}.

\bibitem[Urteaga and Wiggins(2018{\natexlab{c}})]{Urteaga2018}
I.~Urteaga and C.~H. Wiggins.
\newblock {Sequential Monte Carlo for dynamic softmax bandits}.
\newblock In \emph{{1st Symposium on Advances in Approximate Bayesian Inference
  (AABI)}}, 2018{\natexlab{c}}.

\bibitem[Urteaga et~al.(2017)Urteaga, Bugallo, and Djuri\'{c}]{j-Urteaga2017b}
I.~Urteaga, M.~F. Bugallo, and P.~M. Djuri\'{c}.
\newblock {Sequential Monte Carlo for inference of latent ARMA time-series with
  innovations correlated in time}.
\newblock \emph{EURASIP Journal on Advances in Signal Processing},
  2017\penalty0 (1), Dec 2017.
\newblock \doi{10.1186/s13634-017-0518-4}.
\newblock URL \url{https://doi.org/10.1186/s13634-017-0518-4}.

\bibitem[Urteaga et~al.(2023)Urteaga, Draidia, Lancewicki, and
  Khadivi]{Urteaga2023}
I.~Urteaga, M.~Z. Draidia, T.~Lancewicki, and S.~Khadivi.
\newblock {Multi-armed bandits for resource efficient, online optimization of
  language model pre-training: the use case of dynamic masking}.
\newblock In A.~Rogers, J.~Boyd-Graber, and N.~Okazaki, editors, \emph{Findings
  of the Association for Computational Linguistics: ACL 2023}, pages
  10609--10627, Toronto, Canada, July 2023. Association for Computational
  Linguistics.
\newblock \doi{10.18653/v1/2023.findings-acl.675}.
\newblock URL \url{https://aclanthology.org/2023.findings-acl.675}.

\bibitem[van Leeuwen(2009)]{j-Leeuwen2009}
P.~J. van Leeuwen.
\newblock {Particle Filtering in Geophysical Systems}.
\newblock \emph{Monthly Weather Review}, 12\penalty0 (137):\penalty0
  4089--4114., 2009.

\bibitem[Whittle(1951)]{b-Whittle1951}
P.~Whittle.
\newblock \emph{{Hypothesis Testing in Time Series Analysis}}.
\newblock Almquist and Wicksell, 1951.

\bibitem[Whittle(1988)]{j-Whittle1988}
P.~Whittle.
\newblock {Restless bandits: activity allocation in a changing world}.
\newblock \emph{Journal of Applied Probability}, 25\penalty0 (A):\penalty0
  287–298, 1988.
\newblock \doi{10.2307/3214163}.

\bibitem[Yu and Mannor(2009)]{ip-Yu2009}
J.~Y. Yu and S.~Mannor.
\newblock {Piecewise-Stationary Bandit Problems with Side Observations}.
\newblock In \emph{Proceedings of the 26th Annual International Conference on
  Machine Learning}, ICML '09, page 1177–1184, New York, NY, USA, 2009.
  Association for Computing Machinery.
\newblock ISBN 9781605585161.
\newblock \doi{10.1145/1553374.1553524}.
\newblock URL \url{https://doi.org/10.1145/1553374.1553524}.

\end{thebibliography}
\bibliographystyle{abbrvnat}

\clearpage
\appendix

\section{SMC-based policies for stationary bandits}
\label{asec:static_bandits}

We here apply the proposed SMC-based Bayesian policies as in Algorithm~\ref{alg:sir-mab}
to the original settings where Thompson sampling and Bayes-UCB were derived,
\ie for stationary bandits with Bernoulli and contextual, linear Gaussian reward functions~\cite{ip-Kaufmann2012,ip-Garivier2011a,ic-Korda2013,ip-Agrawal2013a}.

Empirical results for these bandits is provided in Section~\ref{assec:static_bandits_experiments_analytical},
while the stationary logistic bandit case is evaluated in Section~\ref{assec:static_bandits_experiments_logistic},
where we also evaluate the impact of sample size $M$ in the SMC-based bandit algorithms.

\subsection{SMC-based policies for stationary bandits}
\label{assec:static_bandits_smc}
In stationary bandits, 
there are no time-varying parameters, \ie $\theta_t=\theta, \ \forall t$.
For these cases, SIR-based parameter propagation becomes troublesome \cite{b-Liu2001}.
To mitigate such issues, several alternatives have been proposed in the SMC community:
\eg artificial parameter evolution \cite{j-Gordon1993},
kernel smoothing \cite{b-Liu2001}, and density assisted techniques \cite{ip-Djuric2004}.

We implement density assisted SMC,
rather than kernel based particle filters as in~\cite{j-Cherkassky2013},
where one approximates the posterior of the unknown parameters with a density of choice.
Density assisted importance sampling is a well studied SMC approach
that extends random-walking and kernel-based alternatives~\cite{j-Gordon1993, ib-Liu2001, ip-Djuric2004},
with its asymptotic correctness guaranteed for the static parameter case.
We acknowledge that any of the SMC techniques that further mitigate
the challenges of estimating constant parameters
(\eg parameter smoothing~\cite{j-Carvalho2010,j-Olsson2006,j-Olsson2014}
or nested SMC methods~\cite{j-Chopin2011,j-Crisan2013})
can only improve the accuracy of the implemented SMC-based policies. 

More precisely,
we approximate the posterior of the unknown parameters,
given the current state of knowledge,
with a Gaussian distribution
\begin{equation}
p(\theta_a|\HH_{1:t}) \approx \N{\theta_a|\hat{\theta}_{t,a}, \hat{\Sigma}_{\theta_{t,a}}},
\end{equation}
based on the updated random measure.
Namely, the sufficient statistics of the distribution 
are estimated based on samples and weights of the SMC random measure $p_M(\theta_{t,a})=\sum_{m=1}^M w_{t}^{(m)} \delta\left(\theta_{t,a}-\theta_{t,a}^{(m)}\right)$,
available at each bandit interaction; \ie
\begin{equation}
\begin{split}
\hat{\theta}_{t,a} &= \sum_{i=1}^{M} w_{t,a}^{(m)} \theta^{(m)}_{t,a} \;,  \\
\hat{\Sigma}_{\theta_{t,a}} &= \sum_{i=1}^{M} w_{t,a}^{(m)}(\theta^{(m)}_{t,a} - \hat{\theta}_{t,a})(\theta^{(m)}_{t,a} - \hat{\theta}_{t,a})^\top \;.
\end{split}
\label{eq:proposedMethod_unknownAB_DA_estSuffStatistics}
\end{equation}
Hence, when addressing static bandits,
we slightly modify Algorithm~\ref{alg:sir-mab},
and propagate parameters in Steps 5 and 9-b,
by drawing from
\begin{equation}
p(\theta_{t+1,a}|\theta_{t,a})=p(\theta_{t,a}|\HH_{1:t}) \approx \N{\theta_{t,a}|\hat{\theta}_{t,a}, \hat{\Sigma}_{\theta_{t,a}}} \;.
\end{equation}

\subsection{Experiments with SMC-based Bayesian policies for Bernoulli and Gaussian stationary bandits}
\label{assec:static_bandits_experiments_analytical}

We provide results for stationary bandits with 2 and 5 arms,
for Bernoulli rewards in Sections~\ref{asssec:static_bandits_bernoulli_2} \& \ref{asssec:static_bandits_bernoulli_5},
and contextual-Gaussian rewards in Sections~\ref{asssec:static_bandits_linearGaussian_2} \& \ref{asssec:static_bandits_linearGaussian_5}, respectively.

Provided empirical evidence showcases
how the proposed SMC-based Bayesian policies perform satisfactorily in both stationary bandit settings.
The more realistic assumption of unknown reward variance $\sigma^2$ for the contextual, linear Gaussian case is also evaluated,
where SMC-based policies are shown to be equally competitive.

We observe that, as the posterior random measure $p_M(\theta_{t,a})$ becomes more accurate, \eg for $M\geq 500$,
SMC-based TS and UCB perform similarly to their counterpart benchmark policies that make use of analytical posteriors.

We note a increased performance uncertainty due to the SMC posterior random measure,
which is empirically reduced by increasing the number $M$ of Monte Carlo samples:
we illustrate the impact of sample size $M$ in the provided figures.

In general, $M=1000$ samples suffice in our static bandit experiments for accurate estimation of parameter posteriors.
Advanced and dynamic determination of SMC sample size is an active research area, out of the scope of this paper.


\subsubsection{Bernoulli bandits, A=2}
\label{asssec:static_bandits_bernoulli_2}

We present below cumulative regret results for different parameterizations of 2-armed Bernoulli bandits.

\begin{figure}[!h]
	\centering
	\begin{subfigure}[b]{\textwidth}
		\centering
		\includegraphics[width=0.75\textwidth]{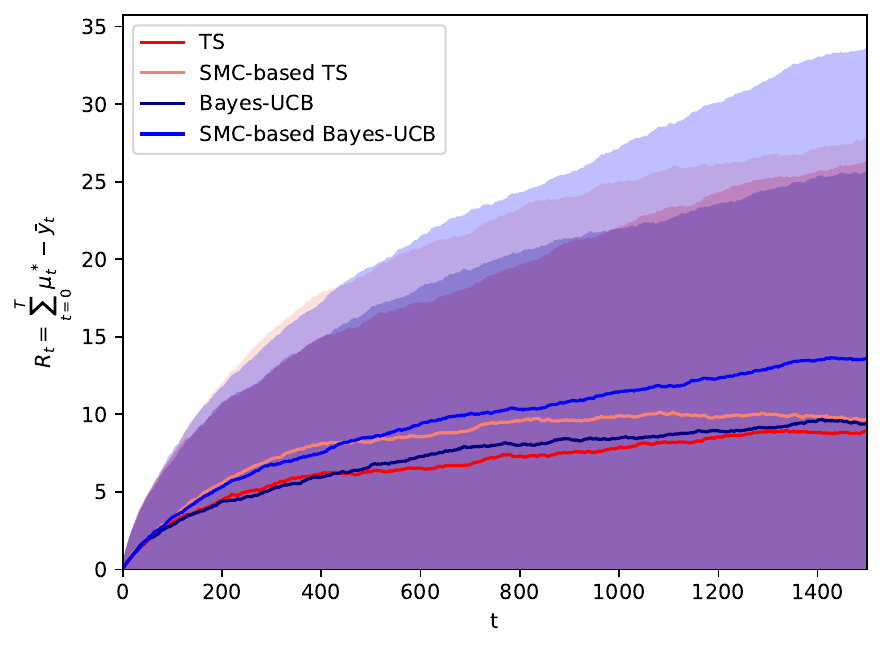}
		\caption{Analytical and SMC-based ($M=1000$) TS and Bayes-UCB.}
	\end{subfigure}
	
	\begin{subfigure}[b]{0.46\textwidth}
		\centering
		\includegraphics[width=\textwidth]{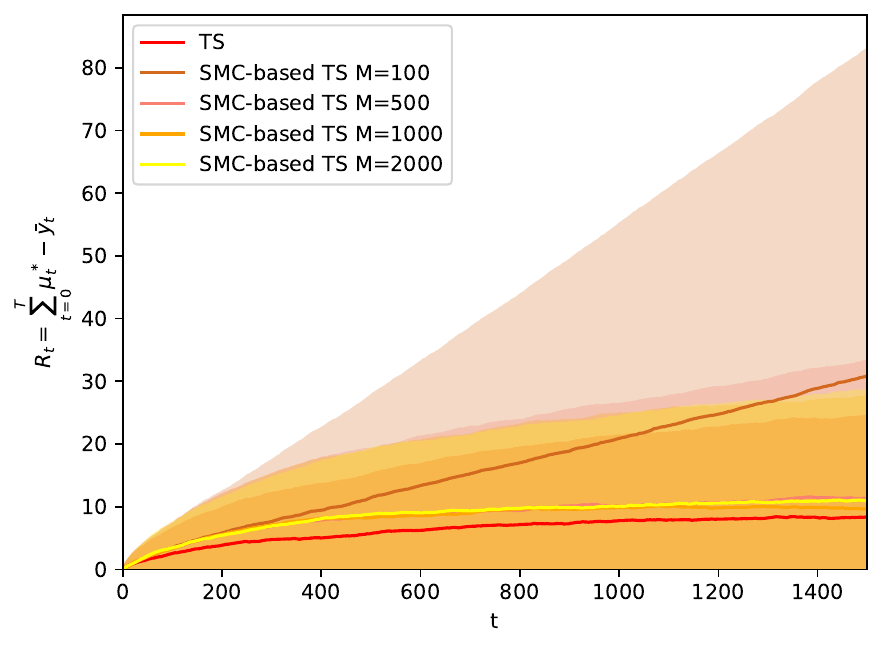}
		\caption{SMC-based TS: impact of $M$.}
	\end{subfigure}
	\begin{subfigure}[b]{0.46\textwidth}
		\centering
		\includegraphics[width=\textwidth]{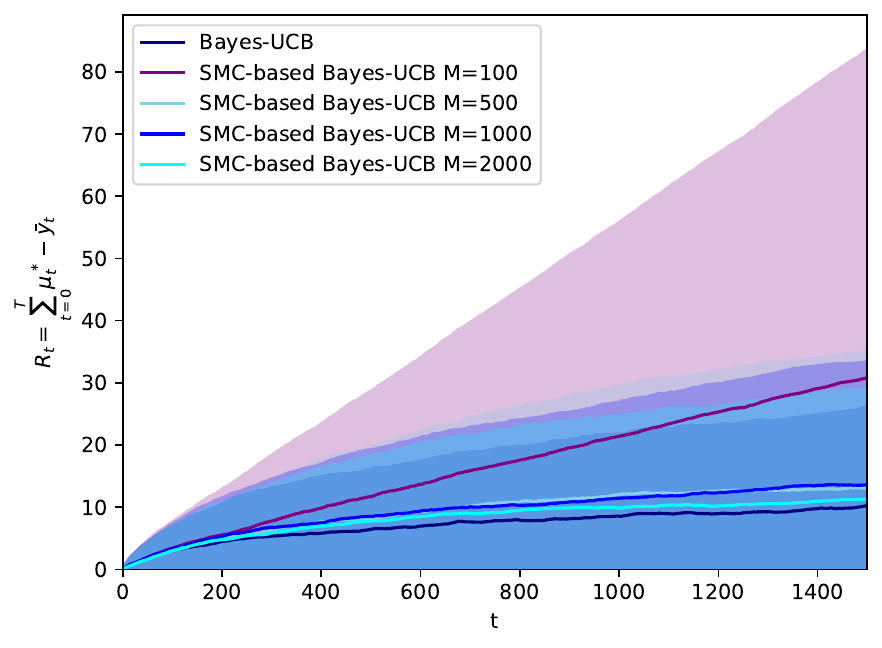}
		\caption{SMC-based Bayes-UCB: impact of $M$}
	\end{subfigure}
	
	\caption{Mean cumulative regret (standard deviation shown as the shaded region) of Bayesian policies in a stationary two-armed Bernoulli bandit, with $\theta_0=0.1, \ \theta_1=0.2$.}
\end{figure}

\begin{figure}[!h]
	\centering
	\begin{subfigure}[b]{\textwidth}
		\centering
		\includegraphics[width=0.75\textwidth]{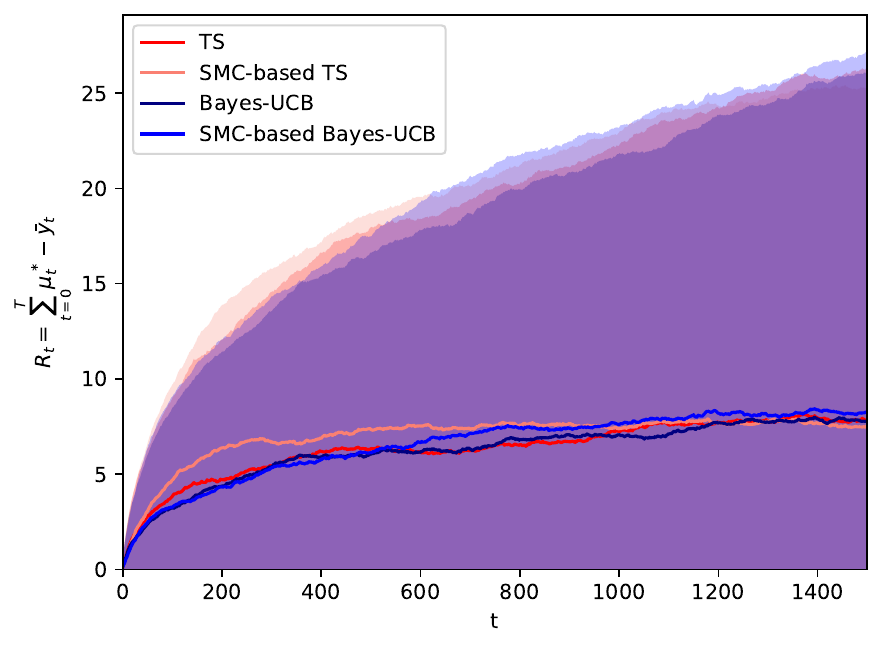}
		\caption{Analytical and SMC-based ($M=1000$) TS and Bayes-UCB.}
	\end{subfigure}
	
	\begin{subfigure}[b]{0.46\textwidth}
		\centering
		\includegraphics[width=\textwidth]{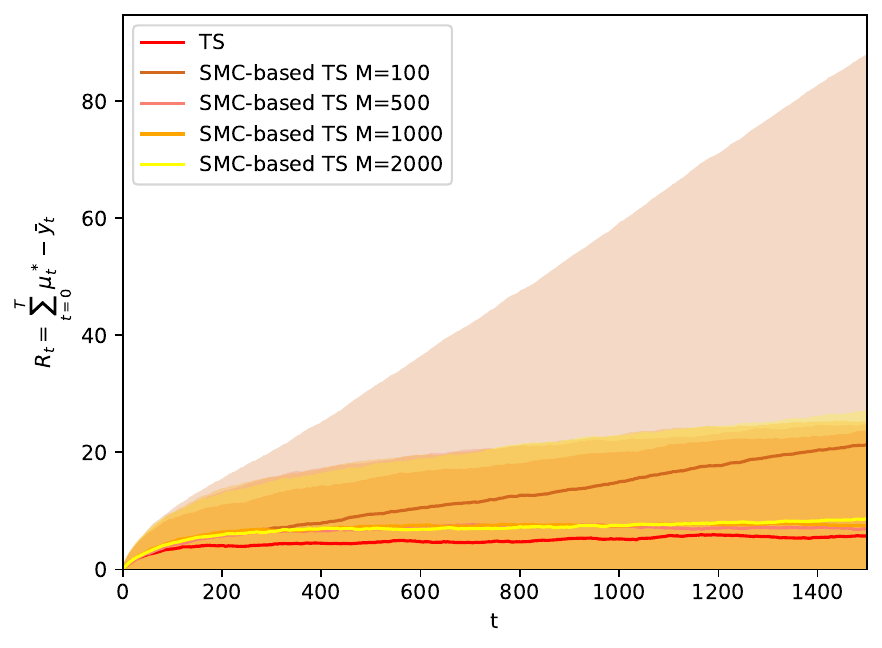}
		\caption{SMC-based TS: impact of $M$.}
	\end{subfigure}
	\begin{subfigure}[b]{0.46\textwidth}
		\centering
		\includegraphics[width=\textwidth]{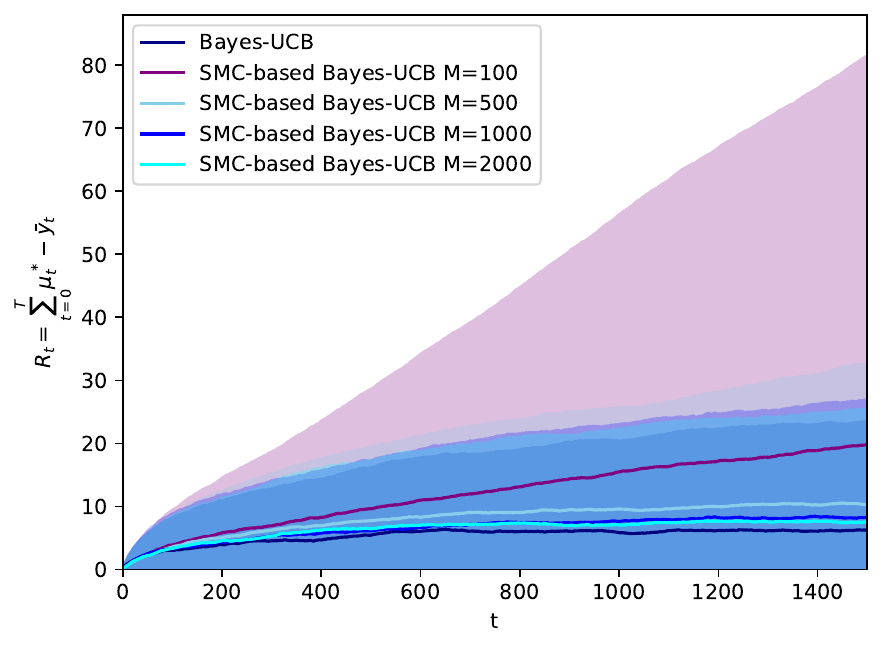}
		\caption{SMC-based Bayes-UCB: impact of $M$}
	\end{subfigure}
	
	\caption{Mean cumulative regret (standard deviation shown as the shaded region) of Bayesian policies in a stationary two-armed Bernoulli bandit, with $\theta_0=0.1, \ \theta_1=0.3$.}
\end{figure}

\begin{figure}[!h]
	\centering
	\begin{subfigure}[b]{\textwidth}
		\centering
		\includegraphics[width=0.75\textwidth]{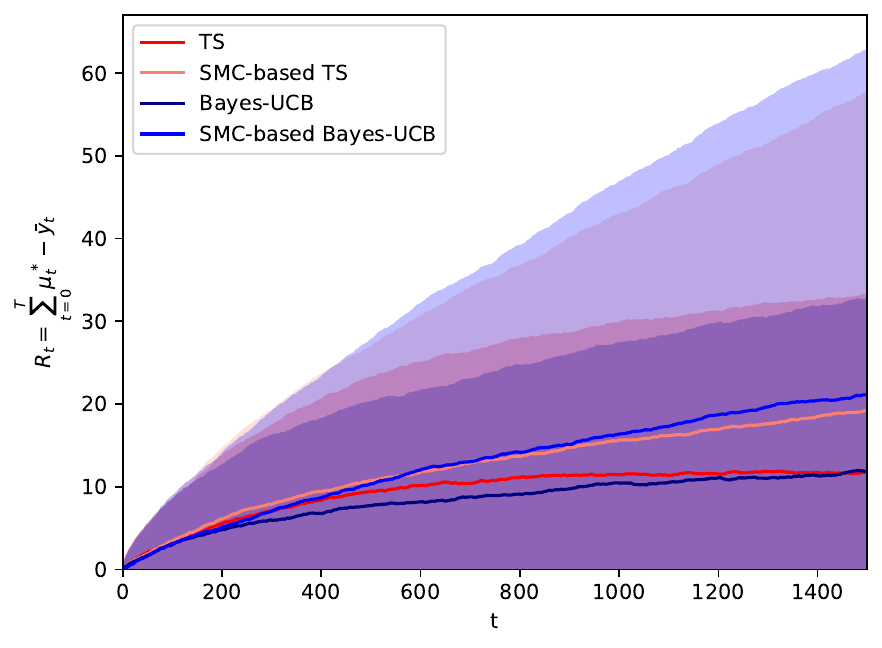}
		\caption{Analytical and SMC-based ($M=1000$) TS and Bayes-UCB.}
	\end{subfigure}
	
	\begin{subfigure}[b]{0.46\textwidth}
		\centering
		\includegraphics[width=\textwidth]{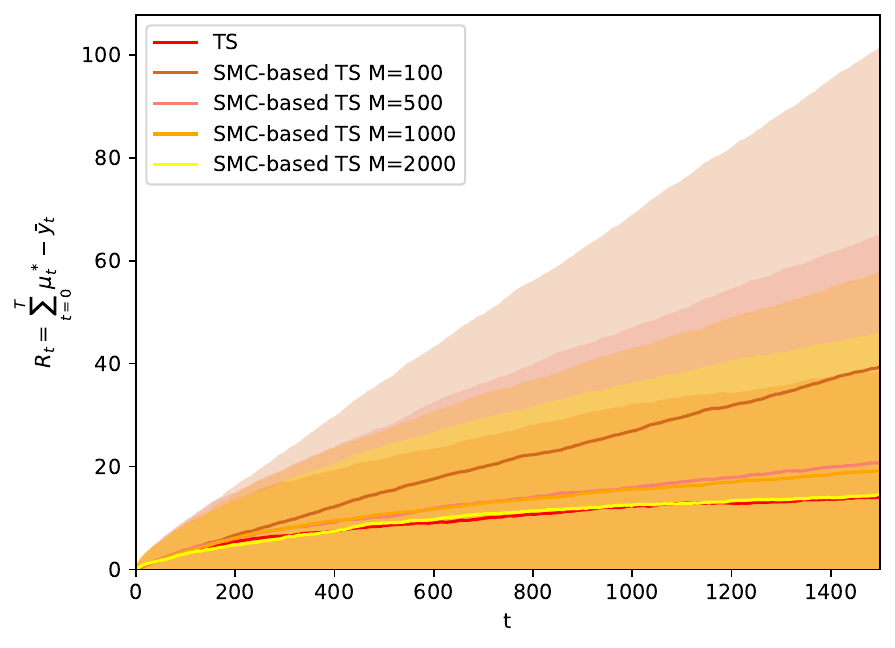}
		\caption{SMC-based TS: impact of $M$.}
	\end{subfigure}
	\begin{subfigure}[b]{0.46\textwidth}
		\centering
		\includegraphics[width=\textwidth]{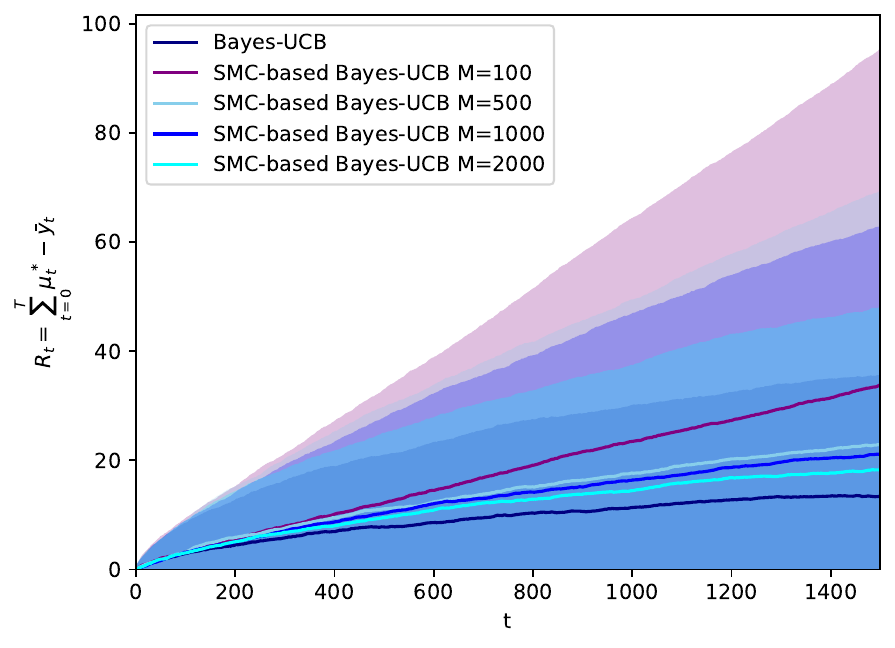}
		\caption{SMC-based Bayes-UCB: impact of $M$}
	\end{subfigure}
	
	\caption{Mean cumulative regret (standard deviation shown as the shaded region) of Bayesian policies in a stationary two-armed Bernoulli bandit, with $\theta_0=0.5, \ \theta_1=0.6$.}
\end{figure}

\begin{figure}[!h]
	\centering
	\begin{subfigure}[b]{\textwidth}
		\centering
		\includegraphics[width=0.75\textwidth]{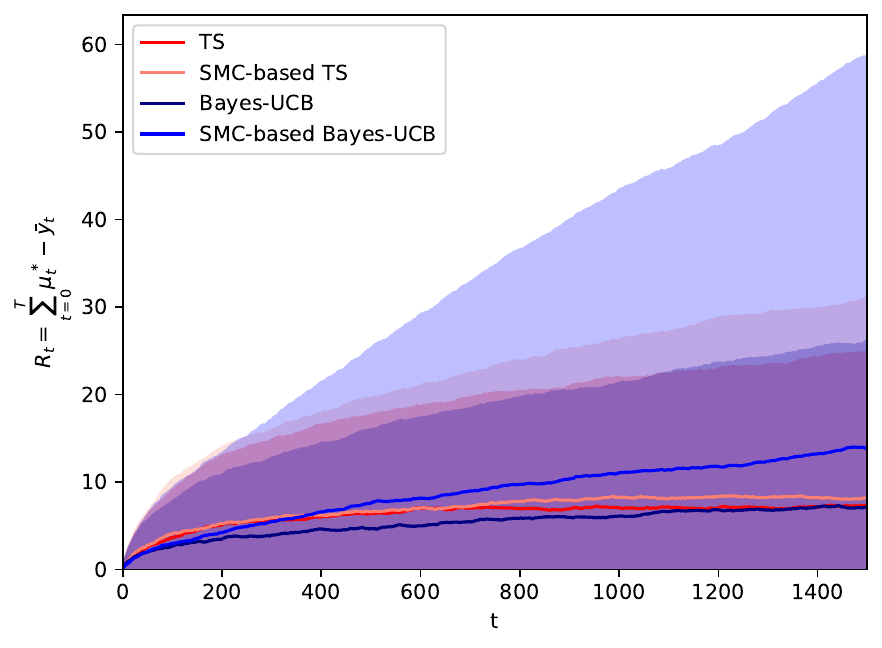}
		\caption{Analytical and SMC-based ($M=1000$) TS and Bayes-UCB.}
	\end{subfigure}
	
	\begin{subfigure}[b]{0.46\textwidth}
		\centering
		\includegraphics[width=\textwidth]{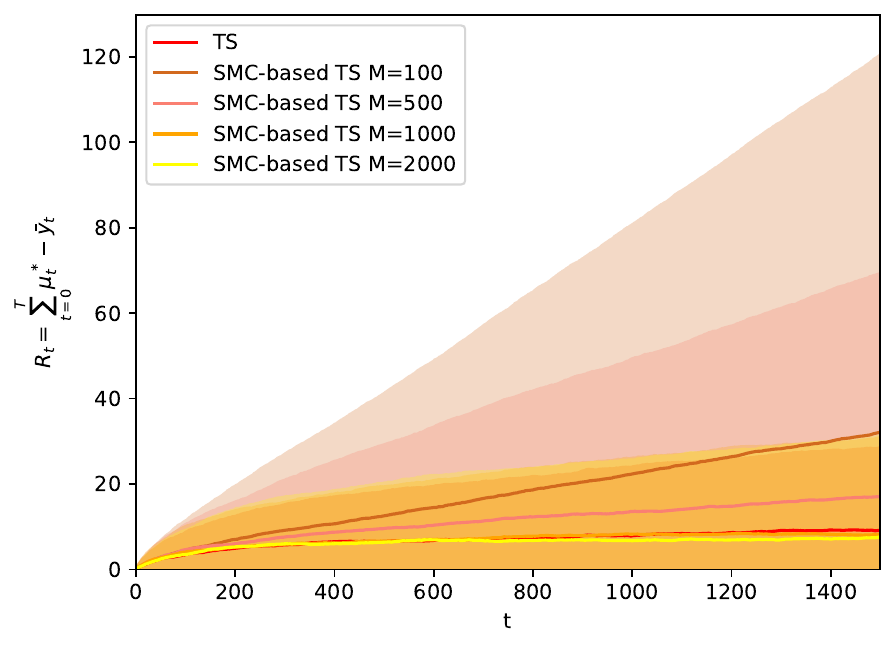}
		\caption{SMC-based TS: impact of $M$.}
	\end{subfigure}
	\begin{subfigure}[b]{0.46\textwidth}
		\centering
		\includegraphics[width=\textwidth]{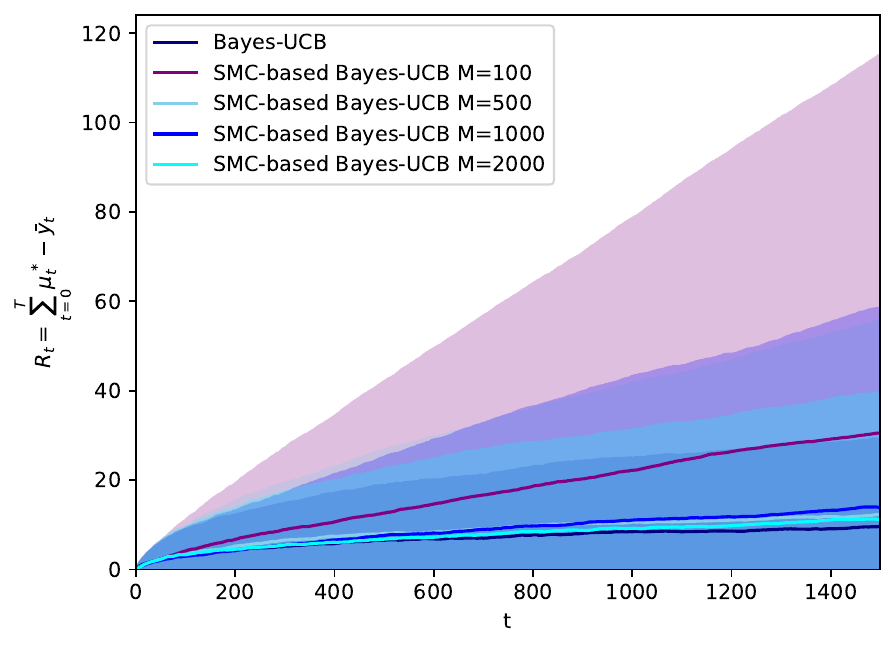}
		\caption{SMC-based Bayes-UCB: impact of $M$}
	\end{subfigure}
	
	\caption{Mean cumulative regret (standard deviation shown as the shaded region) of Bayesian policies in a stationary two-armed Bernoulli bandit, with $\theta_0=0.5, \ \theta_1=0.7$.}
\end{figure}

\begin{figure}[!h]
	\centering
	\begin{subfigure}[b]{\textwidth}
		\centering
		\includegraphics[width=0.75\textwidth]{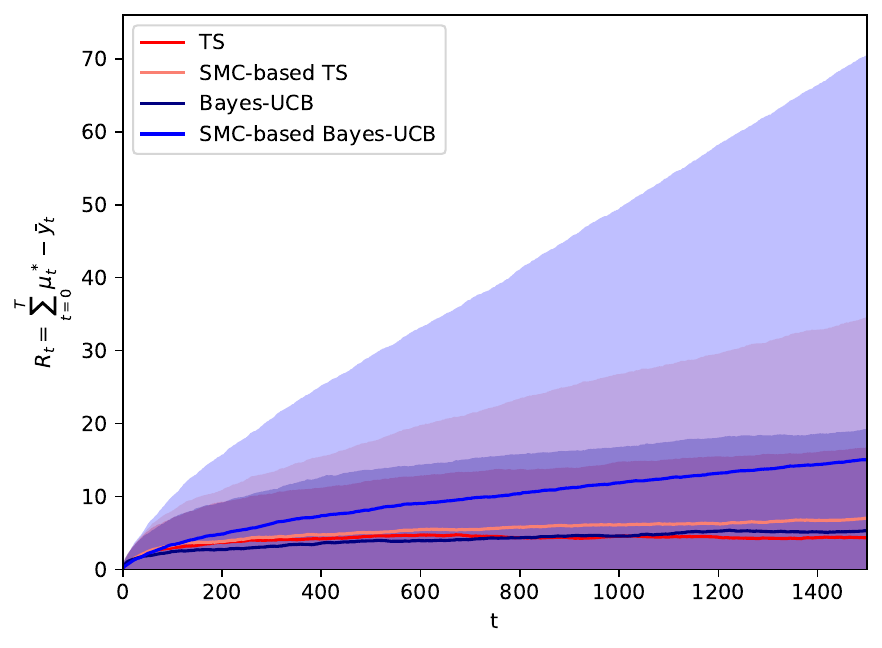}
		\caption{Analytical and SMC-based ($M=1000$) TS and Bayes-UCB.}
	\end{subfigure}
	
	\begin{subfigure}[b]{0.46\textwidth}
		\centering
		\includegraphics[width=\textwidth]{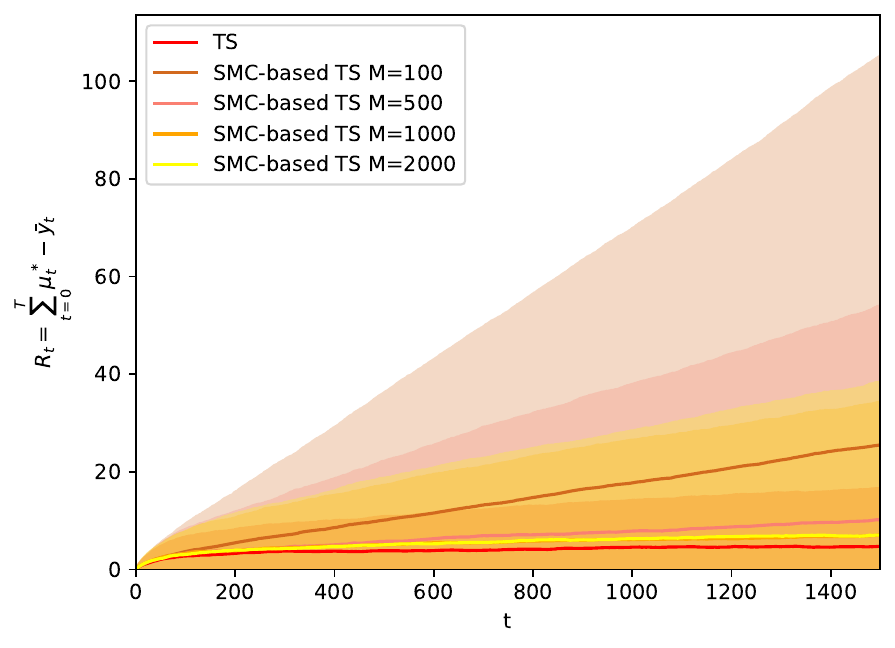}
		\caption{SMC-based TS: impact of $M$.}
	\end{subfigure}
	\begin{subfigure}[b]{0.46\textwidth}
		\centering
		\includegraphics[width=\textwidth]{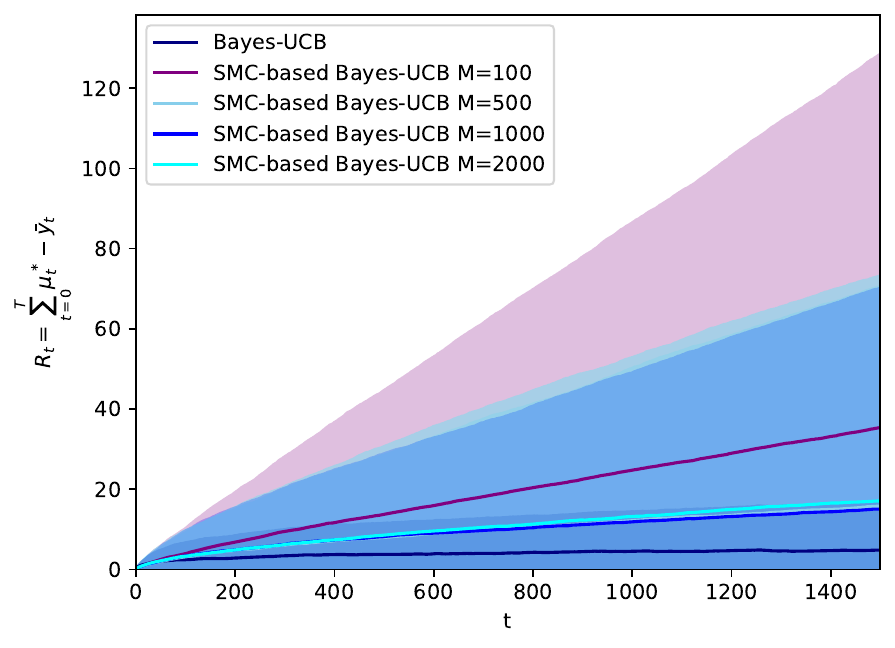}
		\caption{SMC-based Bayes-UCB: impact of $M$}
	\end{subfigure}
	
	\caption{Mean cumulative regret (standard deviation shown as the shaded region) of Bayesian policies in a stationary two-armed Bernoulli bandit, with $\theta_0=0.7, \ \theta_1=0.9$.}
\end{figure}

\begin{figure}[!h]
	\centering
	\begin{subfigure}[b]{\textwidth}
		\centering
		\includegraphics[width=0.75\textwidth]{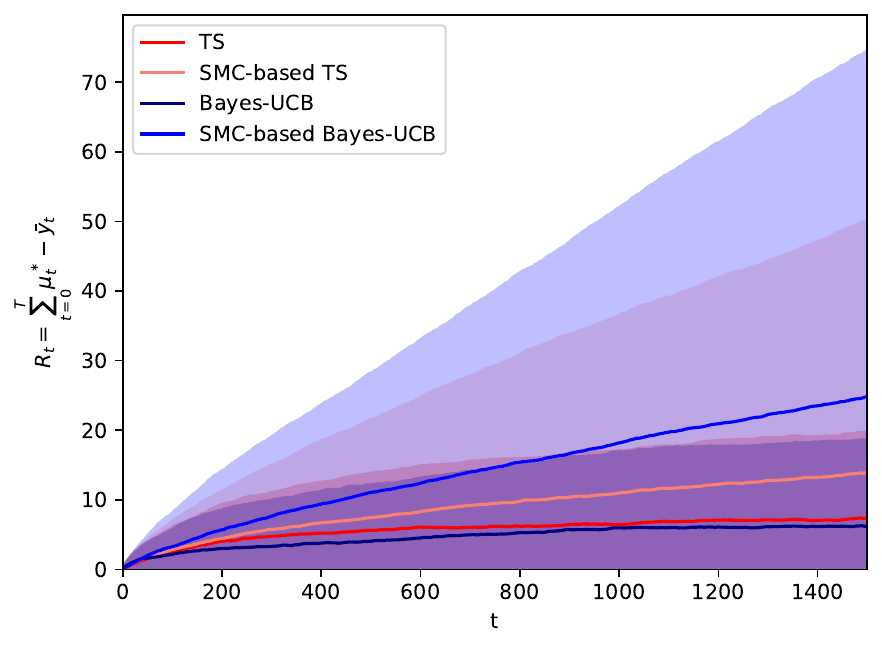}
		\caption{Analytical and SMC-based ($M=1000$) TS and Bayes-UCB.}
	\end{subfigure}
	
	\begin{subfigure}[b]{0.46\textwidth}
		\centering
		\includegraphics[width=\textwidth]{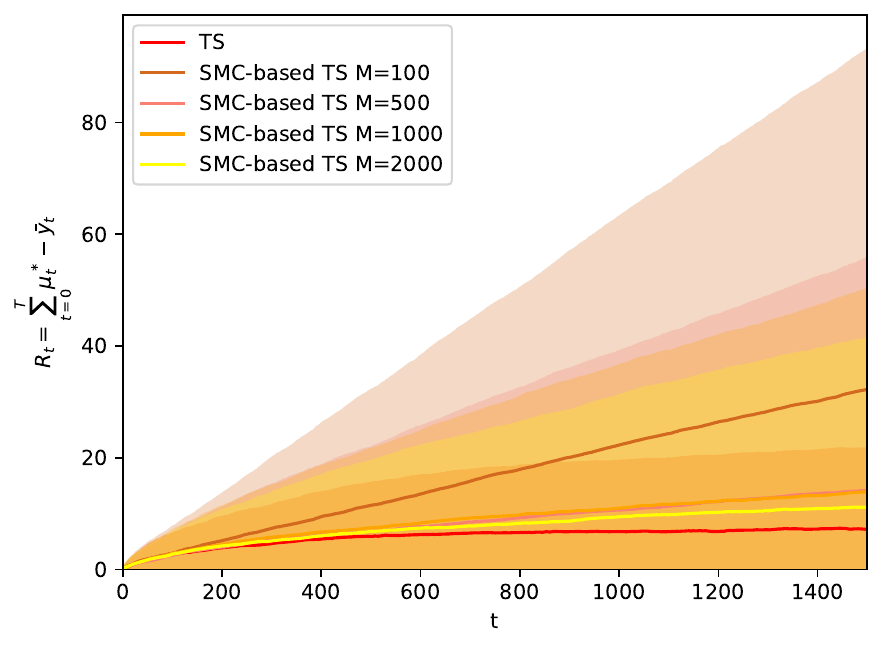}
		\caption{SMC-based TS: impact of $M$.}
	\end{subfigure}
	\begin{subfigure}[b]{0.46\textwidth}
		\centering
		\includegraphics[width=\textwidth]{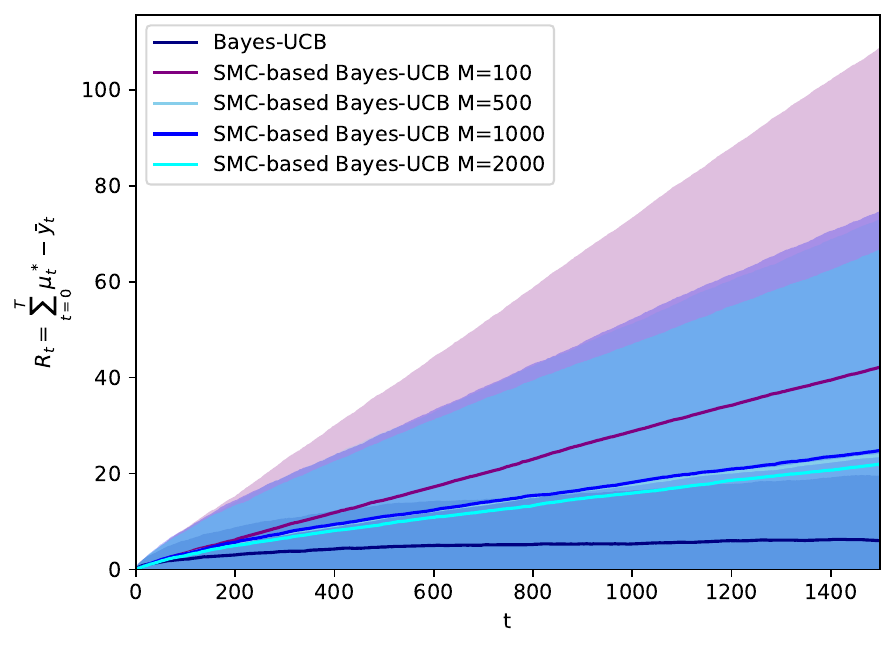}
		\caption{SMC-based Bayes-UCB: impact of $M$}
	\end{subfigure}
	
	\caption{Mean cumulative regret (standard deviation shown as the shaded region) of Bayesian policies in a stationary two-armed Bernoulli bandit, with $\theta_0=0.8, \ \theta_1=0.9$.}
\end{figure}

\clearpage
\subsubsection{Bernoulli bandits, A=5}
\label{asssec:static_bandits_bernoulli_5}

We present below cumulative regret results for different parameterizations of 5-armed Bernoulli bandits.

\begin{figure}[!h]
	\centering
	\begin{subfigure}[b]{\textwidth}
		\centering
		\includegraphics[width=0.75\textwidth]{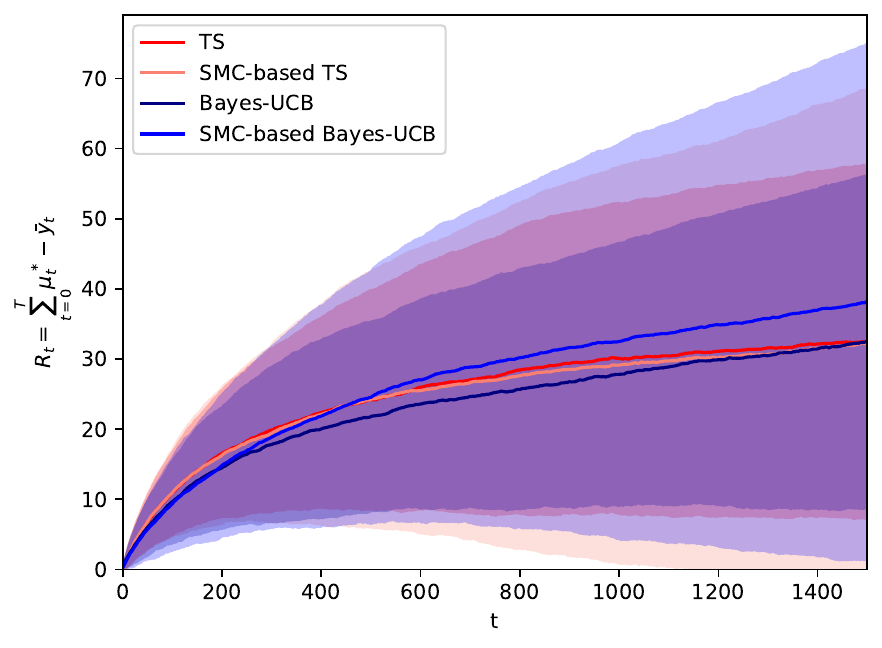}
		\caption{Analytical and SMC-based ($M=1000$) TS and Bayes-UCB.}
	\end{subfigure}
	
	\begin{subfigure}[b]{0.46\textwidth}
		\centering
		\includegraphics[width=\textwidth]{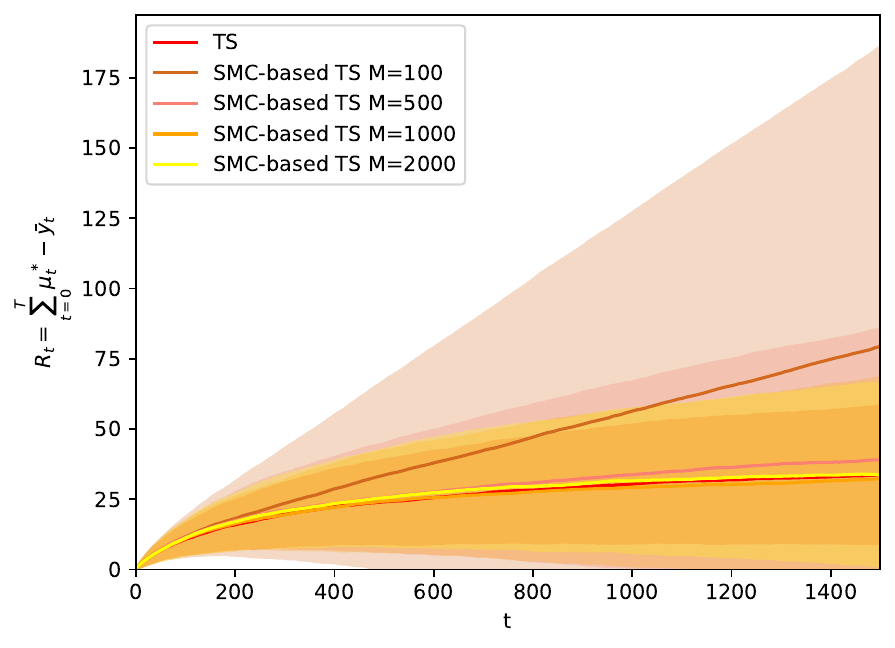}
		\caption{SMC-based TS: impact of $M$.}
	\end{subfigure}
	\begin{subfigure}[b]{0.46\textwidth}
		\centering
		\includegraphics[width=\textwidth]{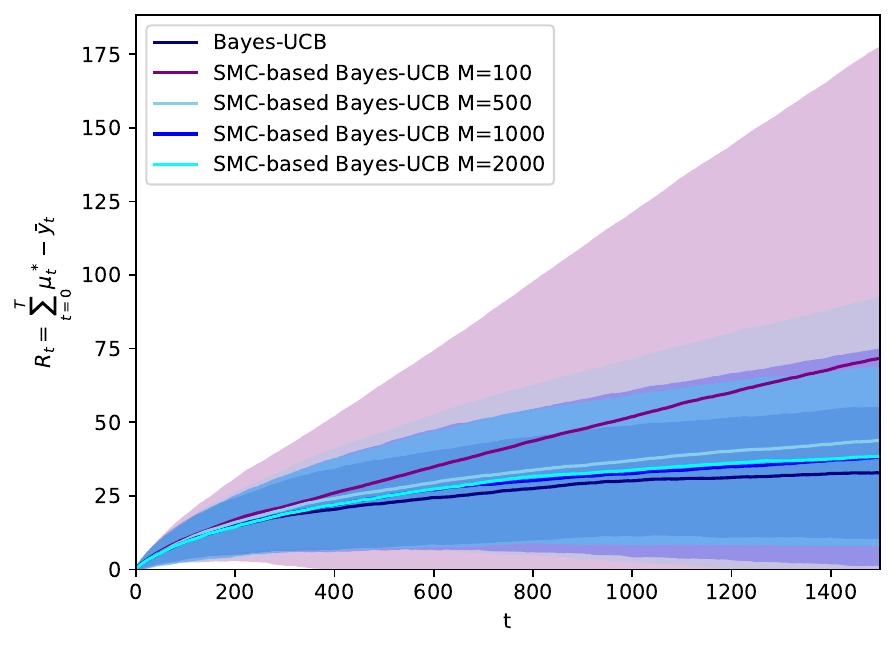}
		\caption{SMC-based Bayes-UCB: impact of $M$}
	\end{subfigure}
	
	\caption{Mean cumulative regret (standard deviation shown as the shaded region) of Bayesian policies in a stationary five-armed Bernoulli bandit:
		$\theta_0=0.1, \ \theta_1=0.2, \ \theta_3=0.3, \ \theta_4=0.4, \ \theta_5=0.5$.
	}
\end{figure}

\begin{figure}[!h]
	\centering
	\begin{subfigure}[b]{\textwidth}
		\centering
		\includegraphics[width=0.75\textwidth]{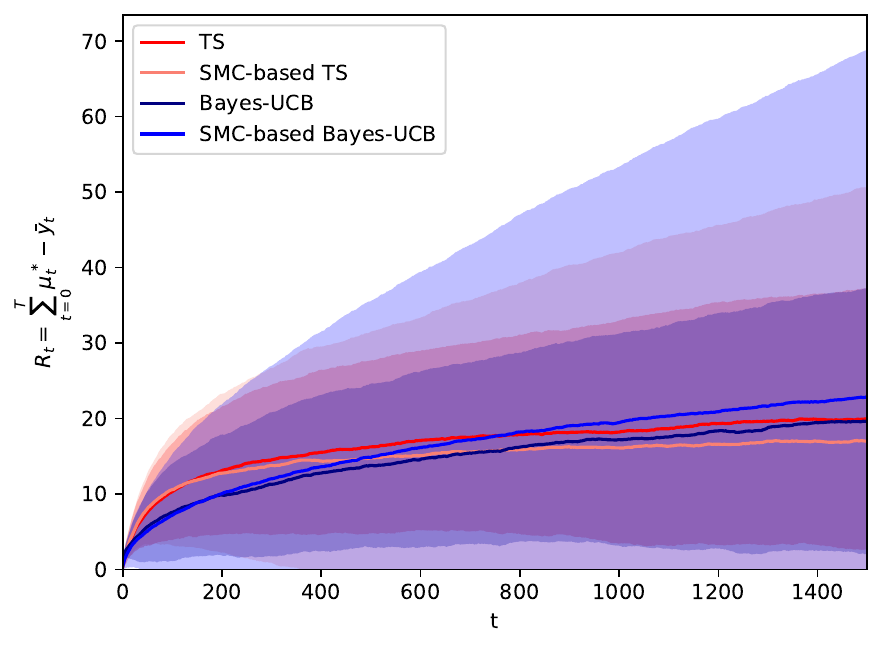}
		\caption{Analytical and SMC-based ($M=1000$) TS and Bayes-UCB.}
	\end{subfigure}
	
	\begin{subfigure}[b]{0.46\textwidth}
		\centering
		\includegraphics[width=\textwidth]{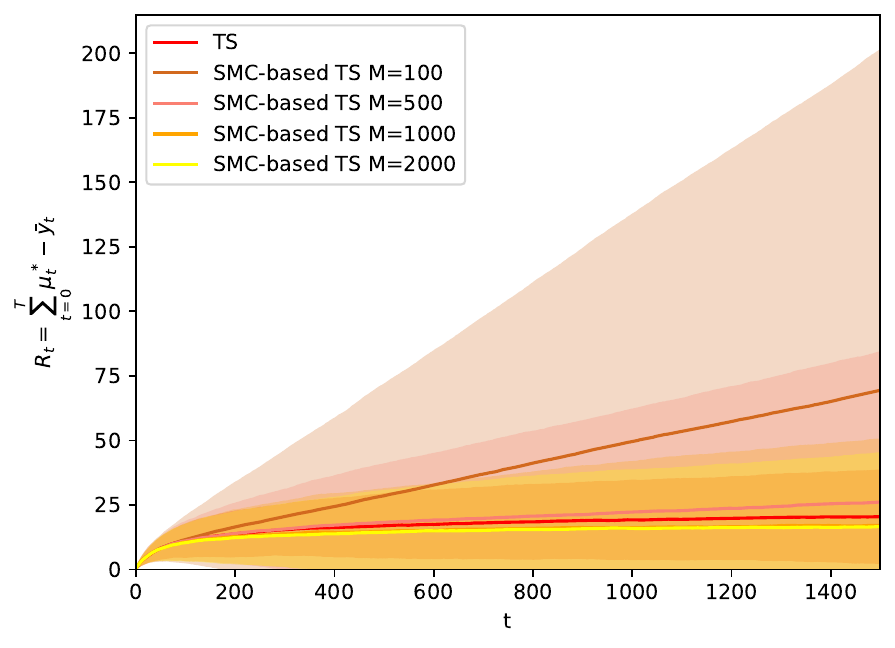}
		\caption{SMC-based TS: impact of $M$.}
	\end{subfigure}
	\begin{subfigure}[b]{0.46\textwidth}
		\centering
		\includegraphics[width=\textwidth]{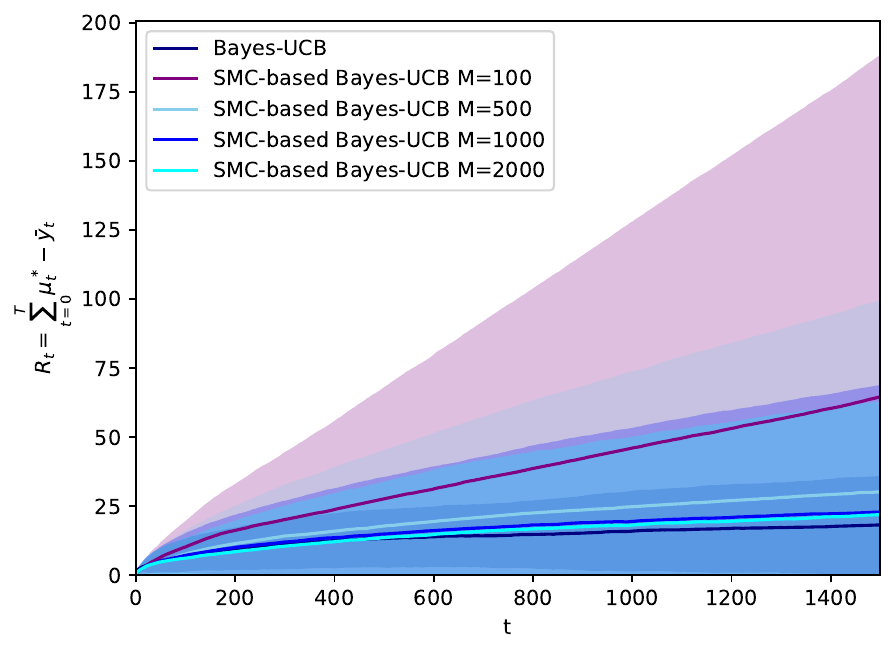}
		\caption{SMC-based Bayes-UCB: impact of $M$}
	\end{subfigure}
	
	\caption{Mean cumulative regret (standard deviation shown as the shaded region) of Bayesian policies in a stationary five-armed Bernoulli bandit:
		 $\theta_0=0.1, \ \theta_1=0.3, \ \theta_3=0.5, \ \theta_4=0.6, \ \theta_5=0.8$.
	}
\end{figure}

\begin{figure}[!h]
	\centering
	\begin{subfigure}[b]{\textwidth}
		\centering
		\includegraphics[width=0.75\textwidth]{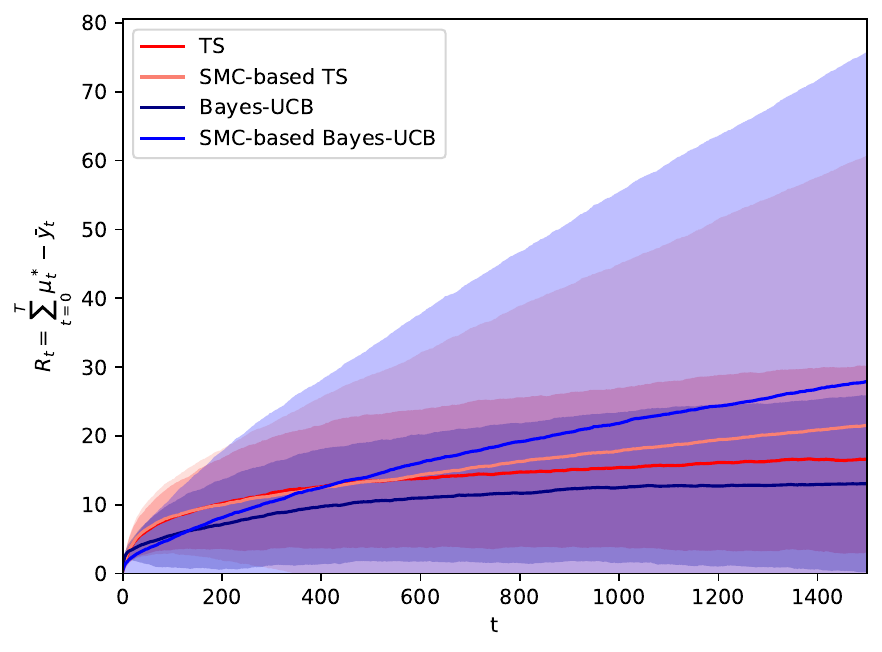}
		\caption{Analytical and SMC-based ($M=1000$) TS and Bayes-UCB.}
	\end{subfigure}
	
	\begin{subfigure}[b]{0.46\textwidth}
		\centering
		\includegraphics[width=\textwidth]{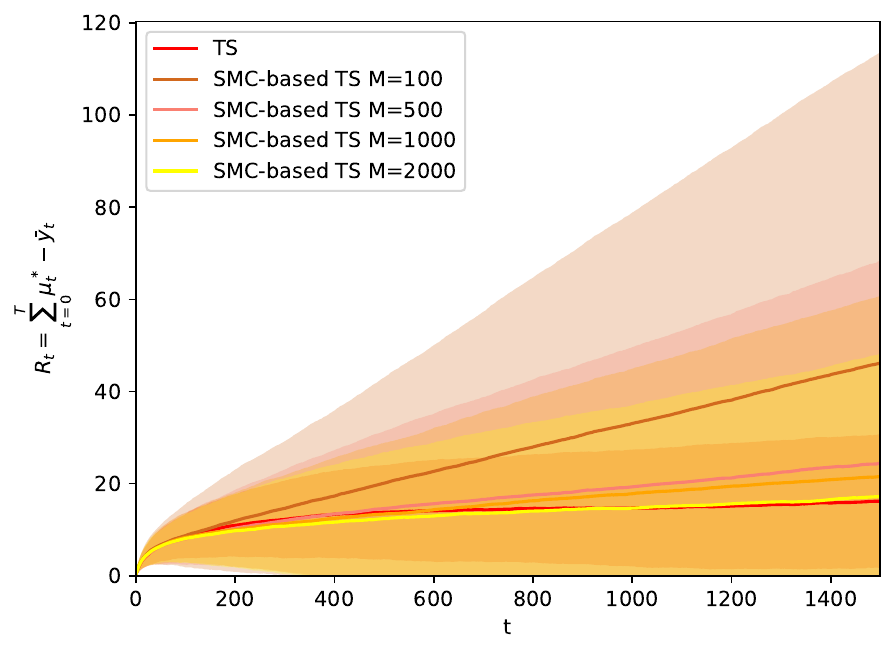}
		\caption{SMC-based TS: impact of $M$.}
	\end{subfigure}
	\begin{subfigure}[b]{0.46\textwidth}
		\centering
		\includegraphics[width=\textwidth]{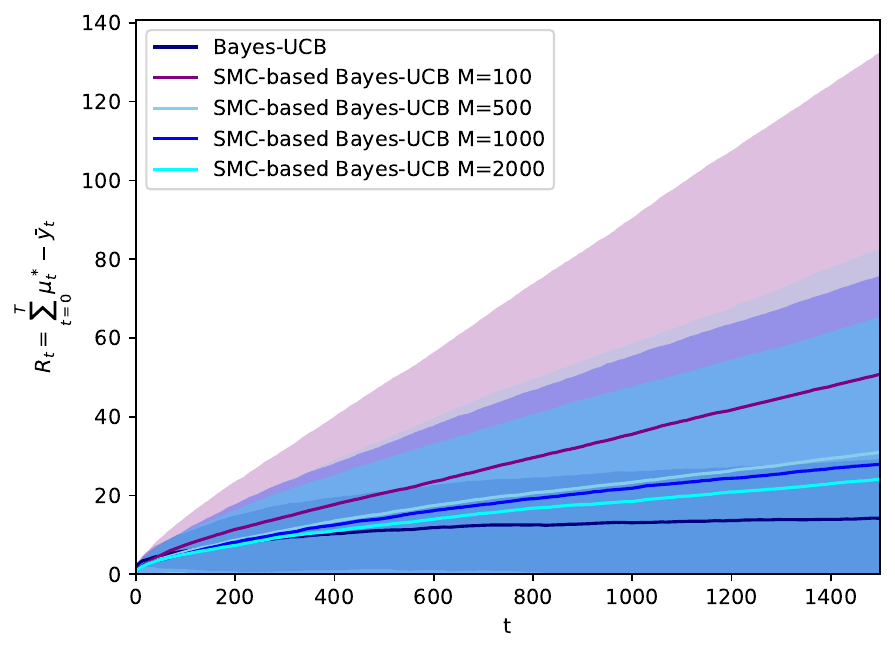}
		\caption{SMC-based Bayes-UCB: impact of $M$}
	\end{subfigure}
	
	\caption{Mean cumulative regret (standard deviation shown as the shaded region) of Bayesian policies in a stationary five-armed Bernoulli bandit:
		$\theta_0=0.1, \ \theta_1=0.3, \ \theta_3=0.5, \ \theta_4=0.8, \ \theta_5=0.9$.
	}
\end{figure}

\begin{figure}[!h]
	\centering
	\begin{subfigure}[b]{\textwidth}
		\centering
		\includegraphics[width=0.75\textwidth]{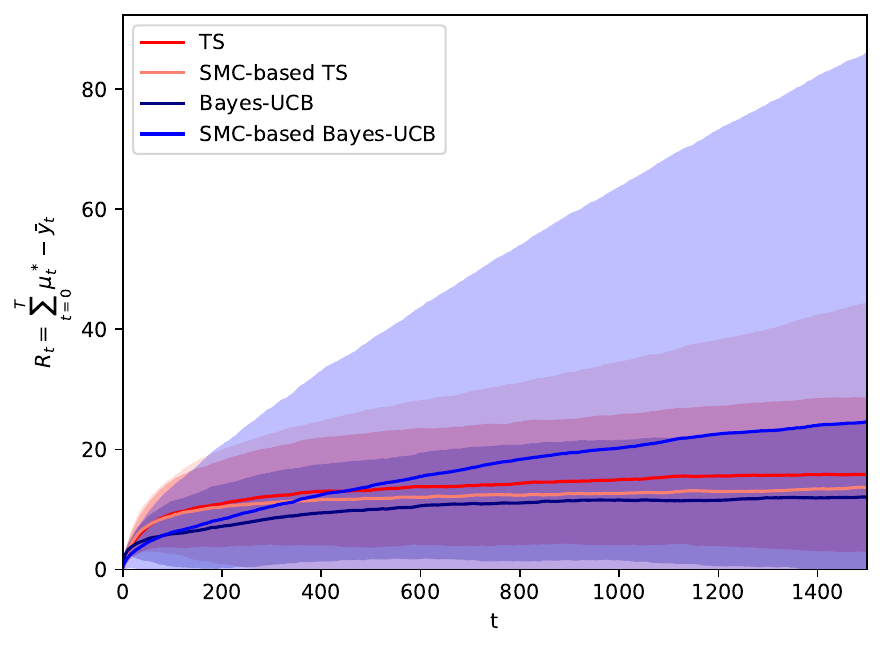}
		\caption{Analytical and SMC-based ($M=1000$) TS and Bayes-UCB.}
	\end{subfigure}
	
	\begin{subfigure}[b]{0.46\textwidth}
		\centering
		\includegraphics[width=\textwidth]{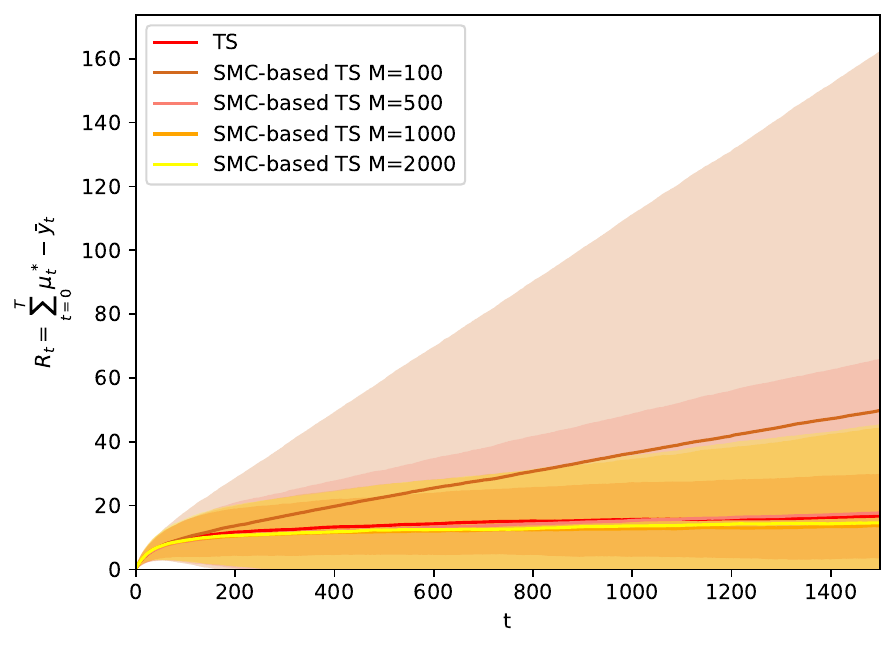}
		\caption{SMC-based TS: impact of $M$.}
	\end{subfigure}
	\begin{subfigure}[b]{0.46\textwidth}
		\centering
		\includegraphics[width=\textwidth]{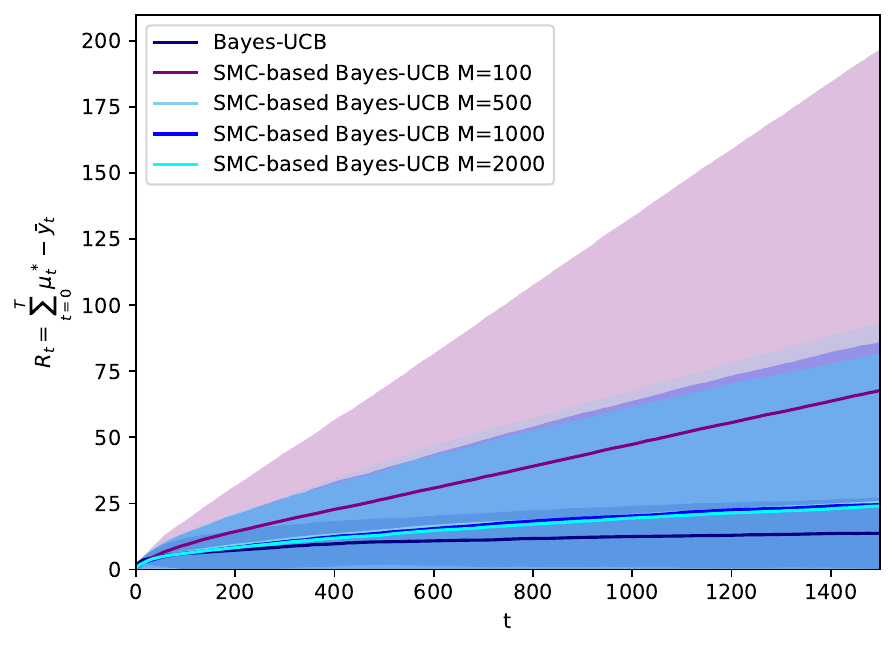}
		\caption{SMC-based Bayes-UCB: impact of $M$}
	\end{subfigure}
	
	\caption{Mean cumulative regret (standard deviation shown as the shaded region) of Bayesian policies in a stationary five-armed Bernoulli bandit:
		 $\theta_0=0.1, \ \theta_1=0.5, \ \theta_3=0.6, \ \theta_4=0.7, \ \theta_5=0.9$.
	}
\end{figure}

\begin{figure}[!h]
	\centering
	\begin{subfigure}[b]{\textwidth}
		\centering
		\includegraphics[width=0.75\textwidth]{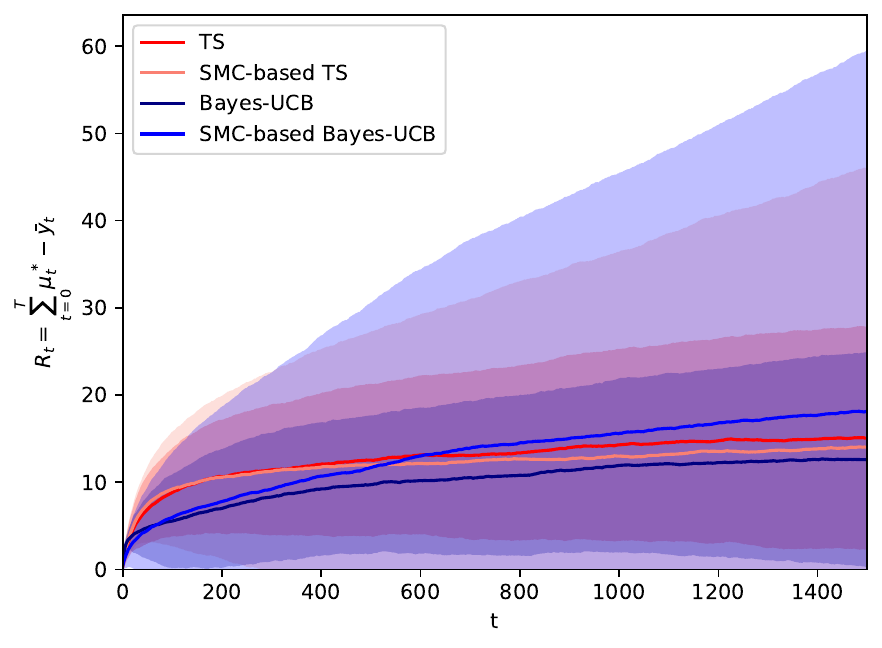}
		\caption{Analytical and SMC-based ($M=1000$) TS and Bayes-UCB.}
	\end{subfigure}
	
	\begin{subfigure}[b]{0.46\textwidth}
		\centering
		\includegraphics[width=\textwidth]{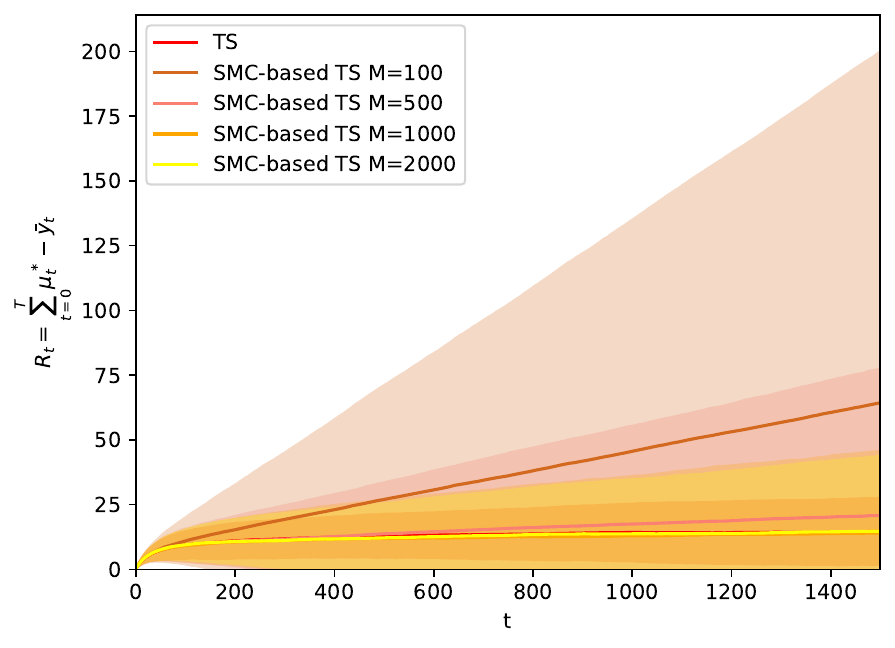}
		\caption{SMC-based TS: impact of $M$.}
	\end{subfigure}
	\begin{subfigure}[b]{0.46\textwidth}
		\centering
		\includegraphics[width=\textwidth]{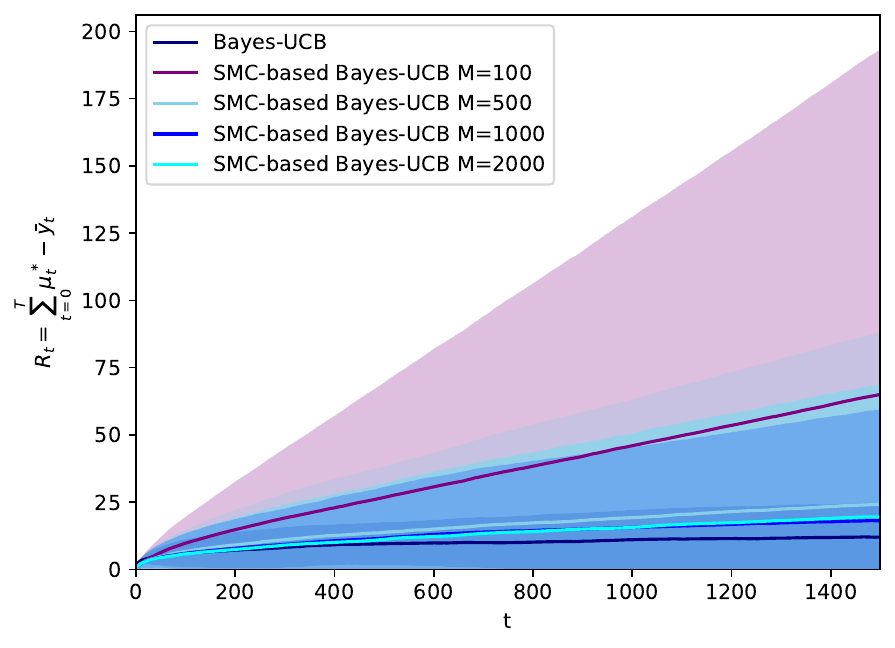}
		\caption{SMC-based Bayes-UCB: impact of $M$}
	\end{subfigure}
	
	\caption{Mean cumulative regret (standard deviation shown as the shaded region) of Bayesian policies in a stationary five-armed Bernoulli bandit:
		 $\theta_0=0.3, \ \theta_1=0.4, \ \theta_3=0.5, \ \theta_4=0.7, \ \theta_5=0.9$.
	}
\end{figure}

\begin{figure}[!h]
	\centering
	\begin{subfigure}[b]{\textwidth}
		\centering
		\includegraphics[width=0.75\textwidth]{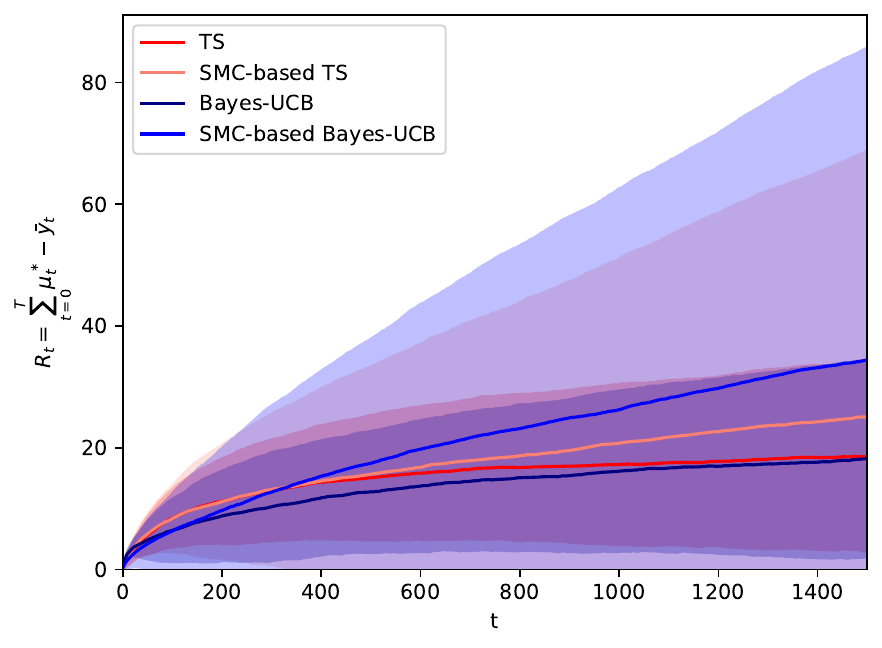}
		\caption{Analytical and SMC-based ($M=1000$) TS and Bayes-UCB.}
	\end{subfigure}
	
	\begin{subfigure}[b]{0.46\textwidth}
		\centering
		\includegraphics[width=\textwidth]{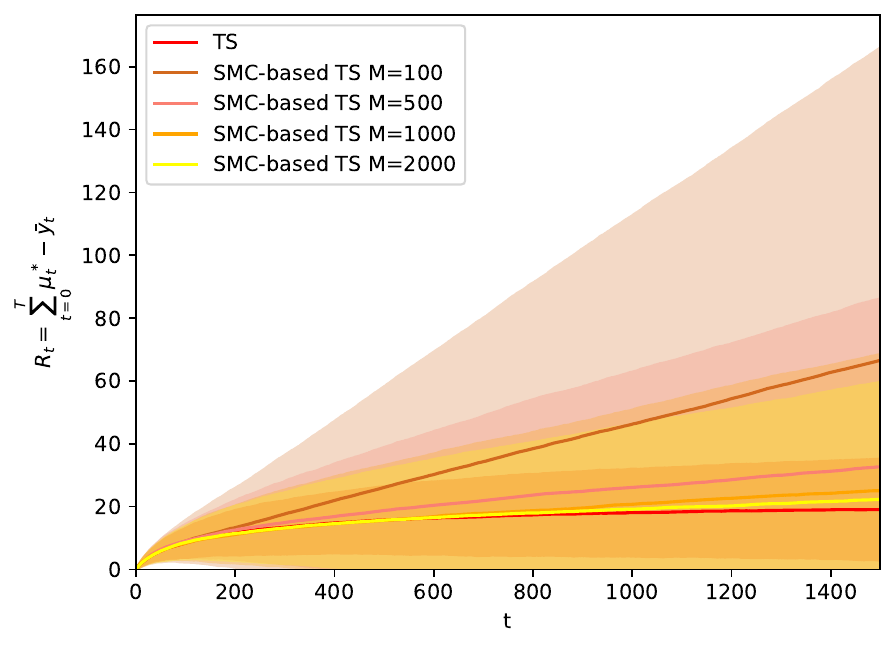}
		\caption{SMC-based TS: impact of $M$.}
	\end{subfigure}
	\begin{subfigure}[b]{0.46\textwidth}
		\centering
		\includegraphics[width=\textwidth]{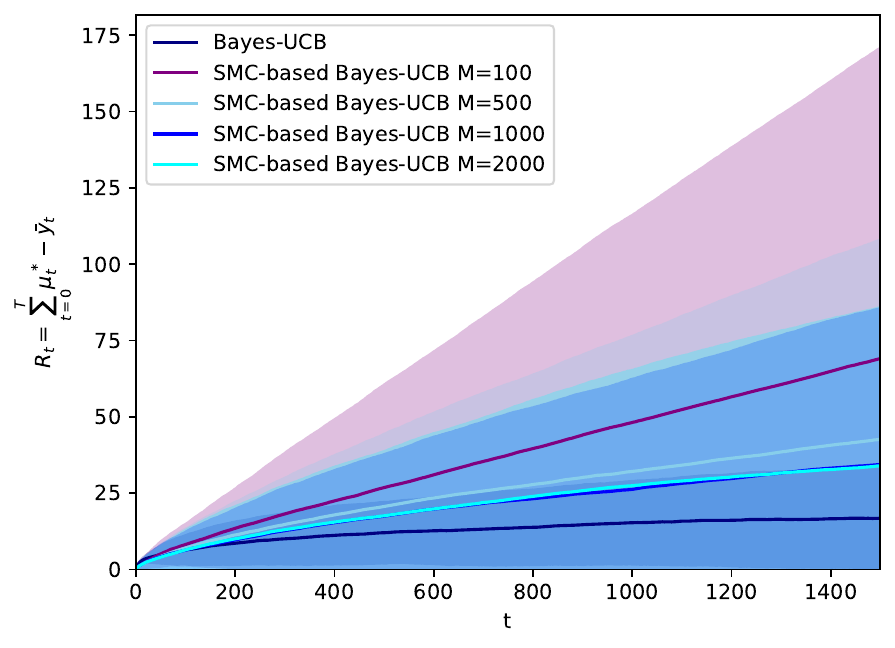}
		\caption{SMC-based Bayes-UCB: impact of $M$}
	\end{subfigure}
	
	\caption{Mean cumulative regret (standard deviation shown as the shaded region) of Bayesian policies in a stationary five-armed Bernoulli bandit:
		 $\theta_0=0.5, \ \theta_1=0.6, \ \theta_3=0.7, \ \theta_4=0.8, \ \theta_5=0.9$.
	}
\end{figure}

\clearpage

\subsubsection{Contextual Linear Gaussian bandits, A=2}
\label{asssec:static_bandits_linearGaussian_2}

We present below cumulative regret results for different parameterizations of 2-armed, contextual linear Gaussian bandits.

\begin{figure}[!h]
	\centering
	\begin{subfigure}[b]{0.46\textwidth}
		\centering
		\includegraphics[width=\textwidth]{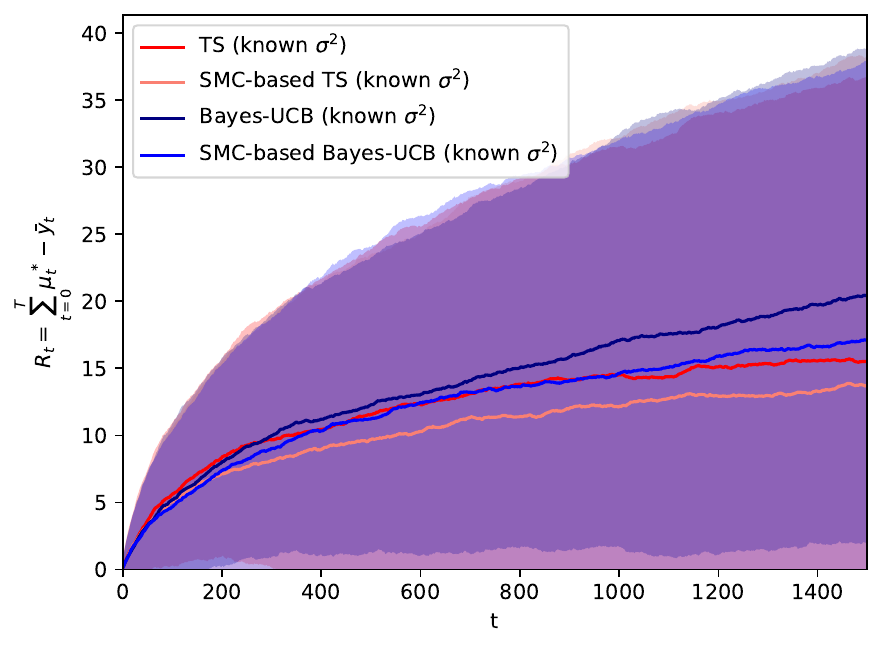}
		\caption{Known $\sigma$: Analytical and SMC-based ($M=1000$) TS and Bayes-UCB.}
	\end{subfigure}
	\begin{subfigure}[b]{0.46\textwidth}
		\centering
		\includegraphics[width=\textwidth]{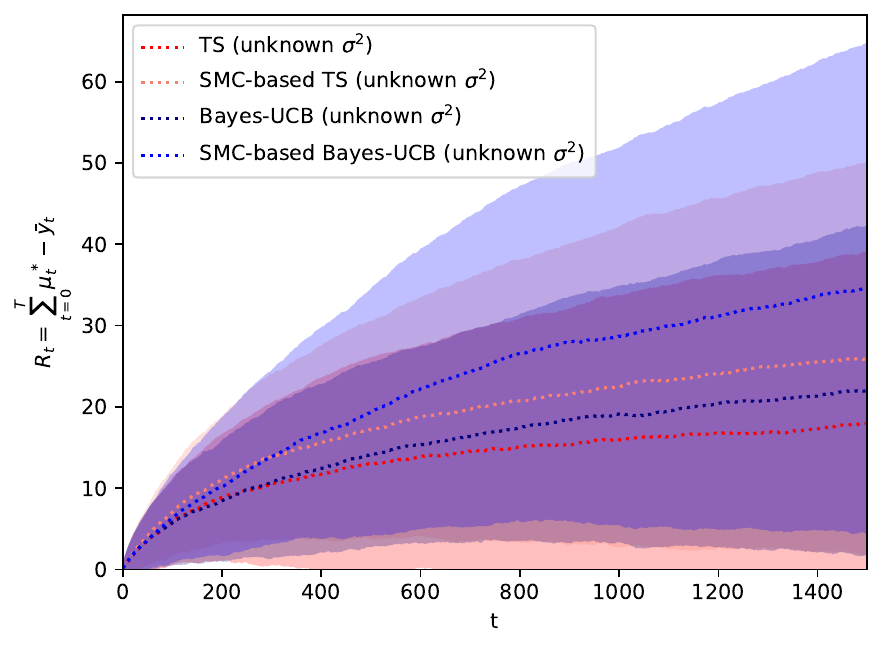}
		\caption{Unknown $\sigma$: Analytical and SMC-based ($M=1000$) TS and Bayes-UCB.}
	\end{subfigure}
	
	\begin{subfigure}[b]{0.46\textwidth}
		\centering
		\includegraphics[width=\textwidth]{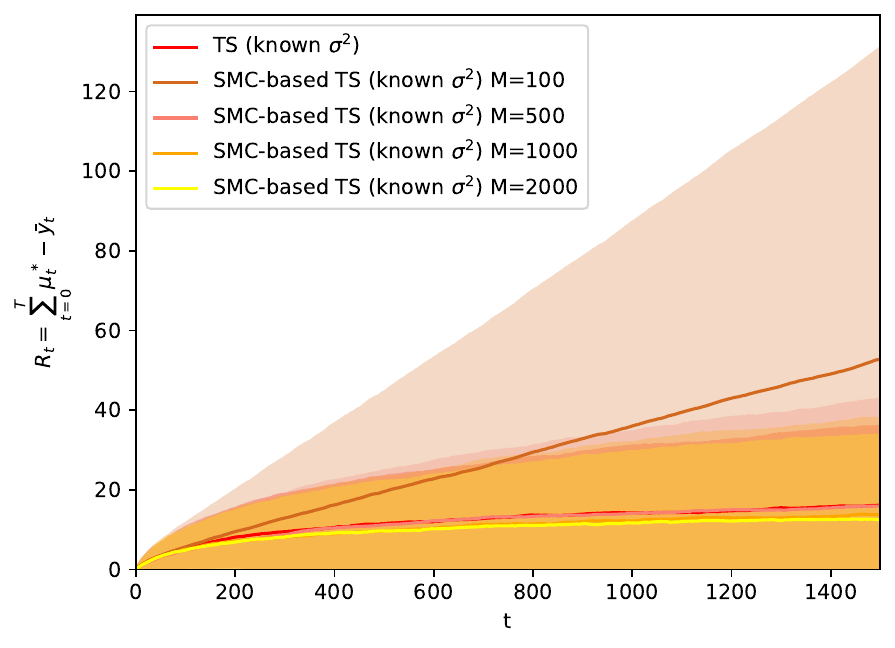}
		\caption{Known $\sigma$: SMC-based TS, \\ impact of $M$.}
	\end{subfigure}
	\begin{subfigure}[b]{0.46\textwidth}
		\centering
		\includegraphics[width=\textwidth]{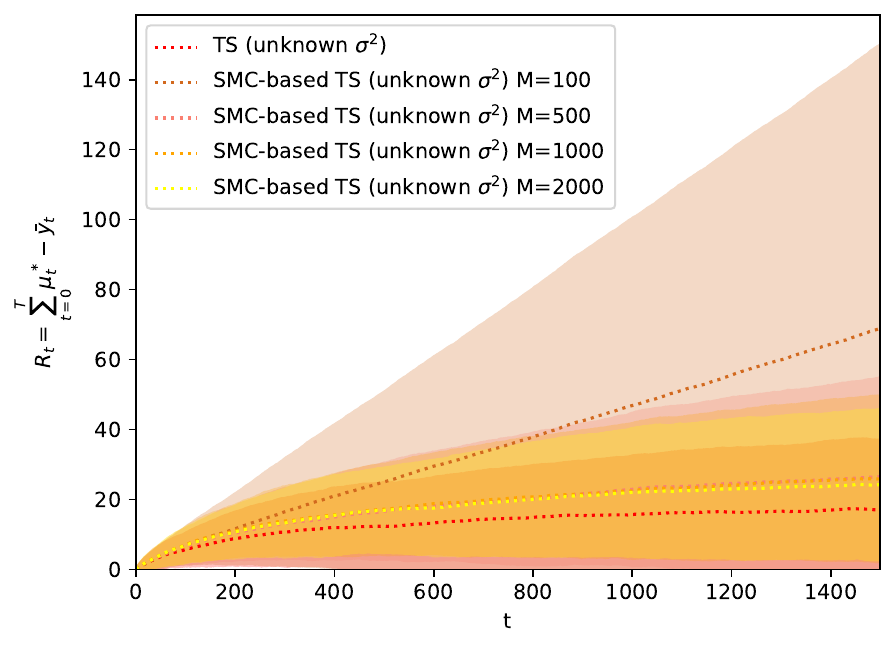}
		\caption{Unknown $\sigma$: SMC-based TS, \\ impact of $M$.}
	\end{subfigure}

	\begin{subfigure}[b]{0.46\textwidth}
		\centering
		\includegraphics[width=\textwidth]{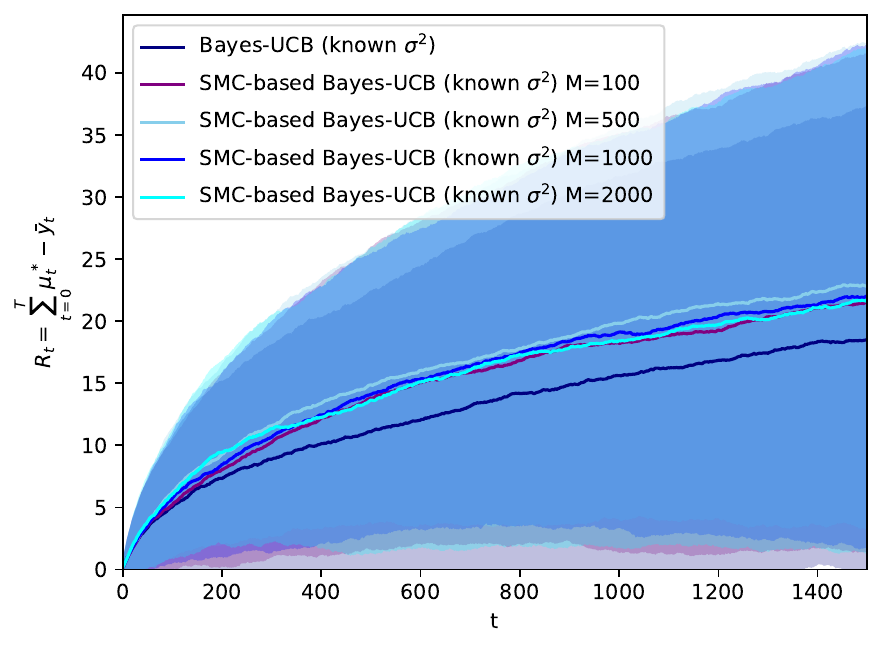}
		\caption{Known $\sigma$: SMC-based Bayes-UCB, impact of $M$.}
	\end{subfigure}
	\begin{subfigure}[b]{0.46\textwidth}
		\centering
		\includegraphics[width=\textwidth]{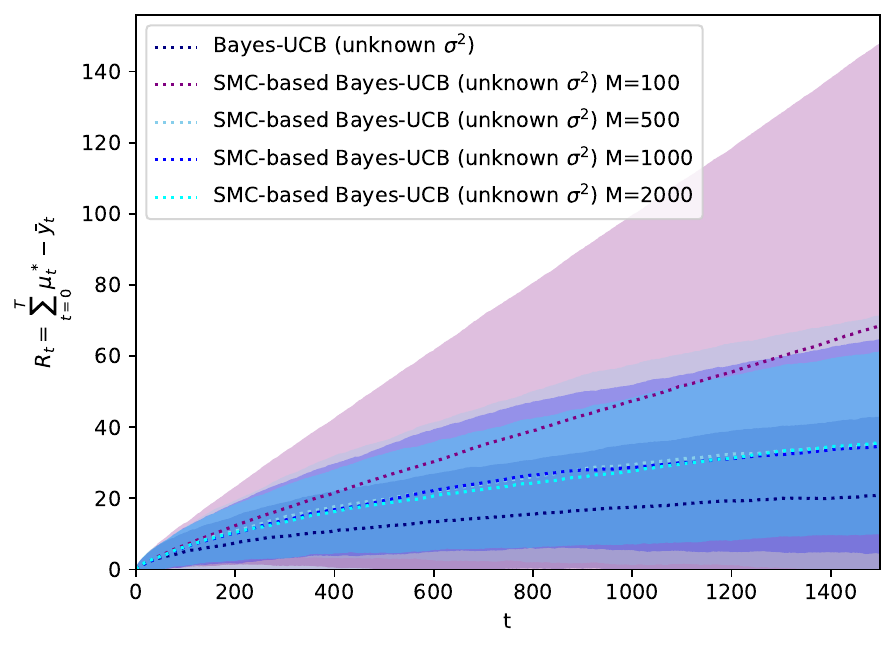}
		\caption{Unknown $\sigma$: SMC-based Bayes-UCB, impact of $M$.}
	\end{subfigure}
	
	\caption{Mean cumulative regret (standard deviation shown as the shaded region) of Bayesian policies in a stationary, two-armed contextual Gaussian bandit:
		$\theta_0=(-0.1,-0.1), \ \theta_1=(0.1,0.1), \sigma^2=0.5$.}
\end{figure}

\clearpage
\subsubsection{Contextual Linear Gaussian bandits, A=5}
\label{asssec:static_bandits_linearGaussian_5}

We present below cumulative regret results for different parameterizations of 5-armed, contextual linear Gaussian bandits.

\begin{figure}[!h]
	\centering
	\begin{subfigure}[b]{0.46\textwidth}
		\centering
		\includegraphics[width=\textwidth]{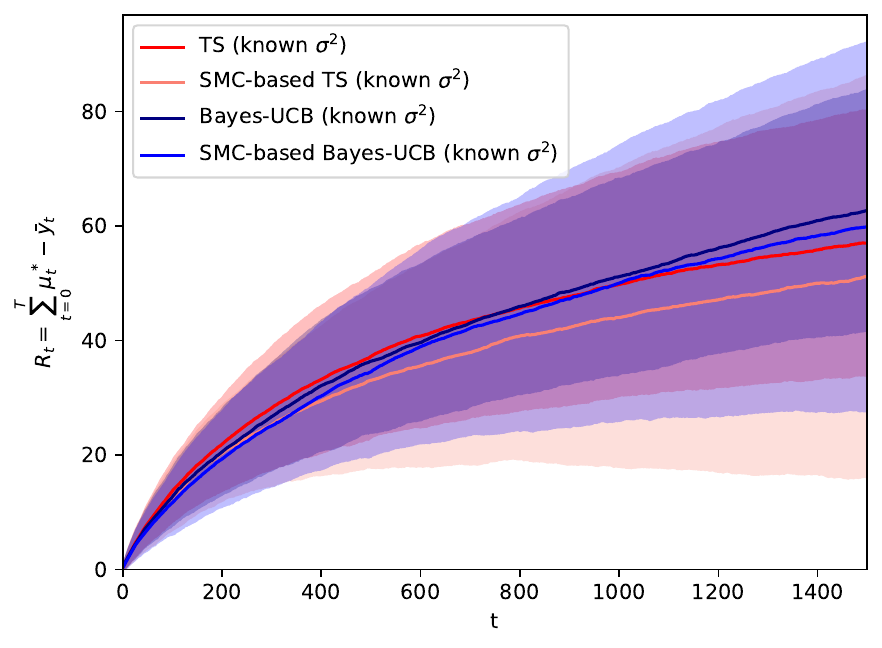}
		\caption{Known $\sigma$: Analytical and SMC-based ($M=1000$) TS and Bayes-UCB.}
	\end{subfigure}
	\begin{subfigure}[b]{0.46\textwidth}
		\centering
		\includegraphics[width=\textwidth]{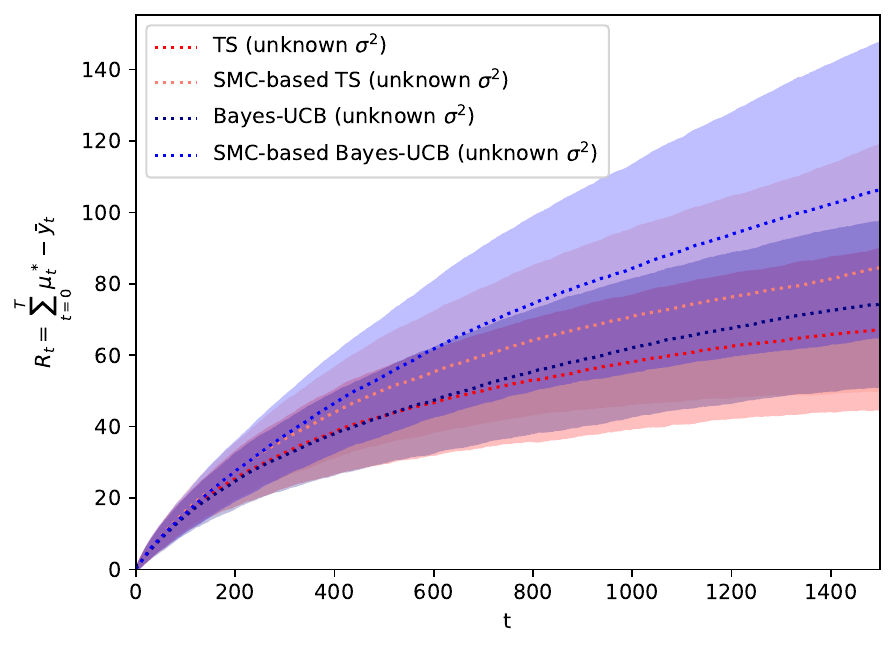}
		\caption{Unknown $\sigma$: Analytical and SMC-based ($M=1000$) TS and Bayes-UCB.}
	\end{subfigure}
	
	\begin{subfigure}[b]{0.46\textwidth}
		\centering
		\includegraphics[width=\textwidth]{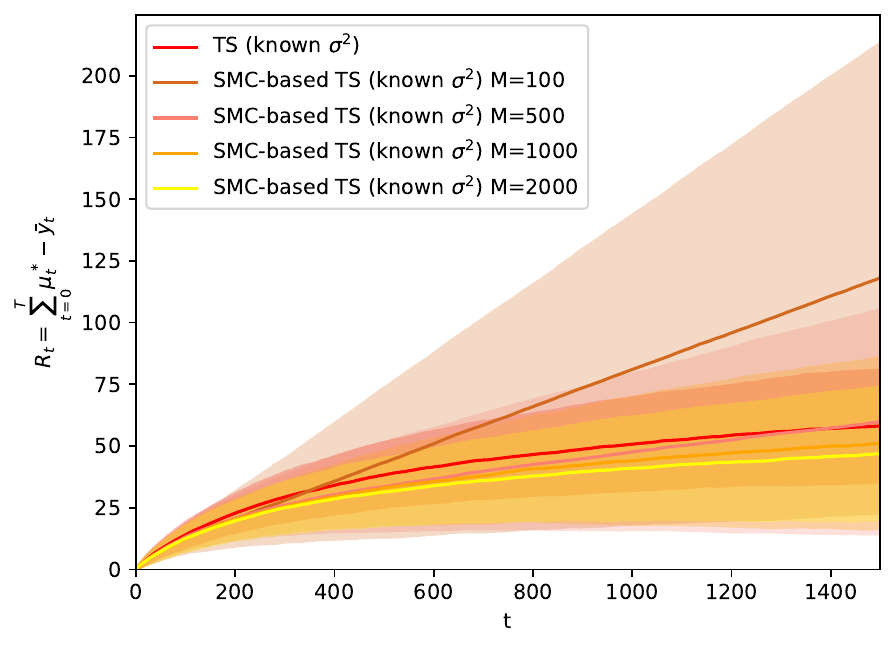}
		\caption{Known $\sigma$: SMC-based TS, \\ impact of $M$.}
	\end{subfigure}
	\begin{subfigure}[b]{0.46\textwidth}
		\centering
		\includegraphics[width=\textwidth]{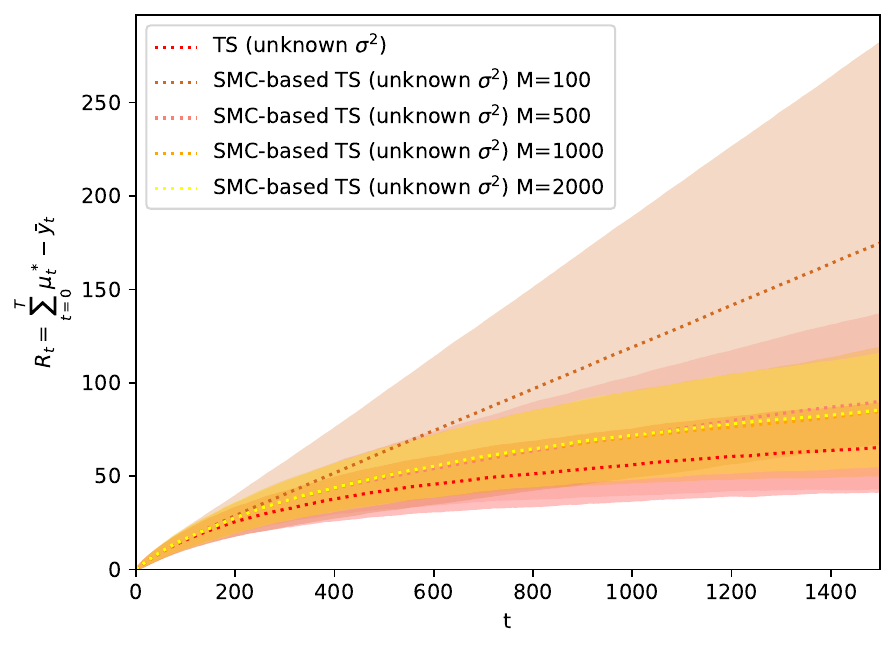}
		\caption{Unknown $\sigma$: SMC-based TS, \\ impact of $M$.}
	\end{subfigure}
	
	\begin{subfigure}[b]{0.46\textwidth}
		\centering
		\includegraphics[width=\textwidth]{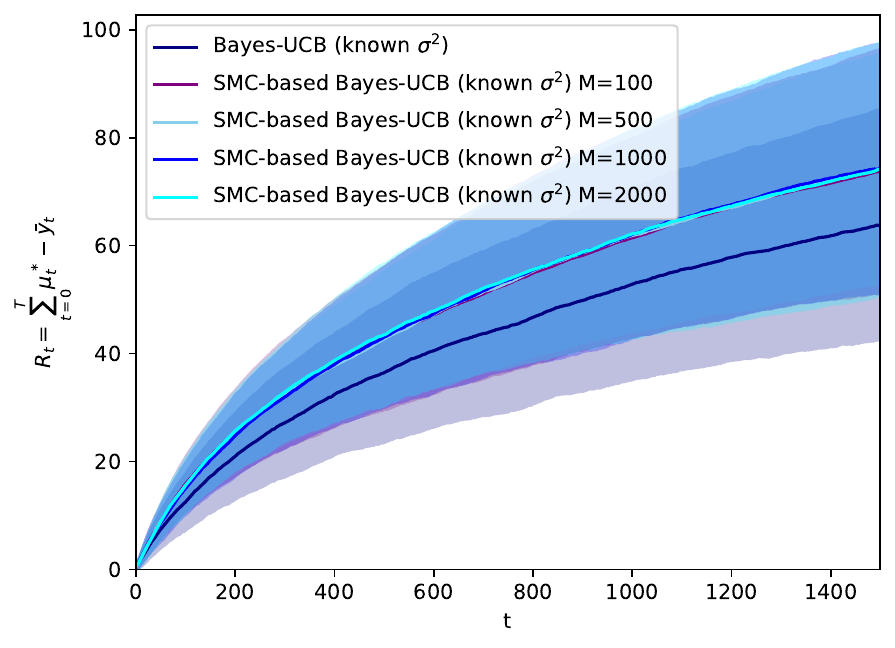}
		\caption{Known $\sigma$: SMC-based Bayes-UCB, impact of $M$.}
	\end{subfigure}
	\begin{subfigure}[b]{0.46\textwidth}
		\centering
		\includegraphics[width=\textwidth]{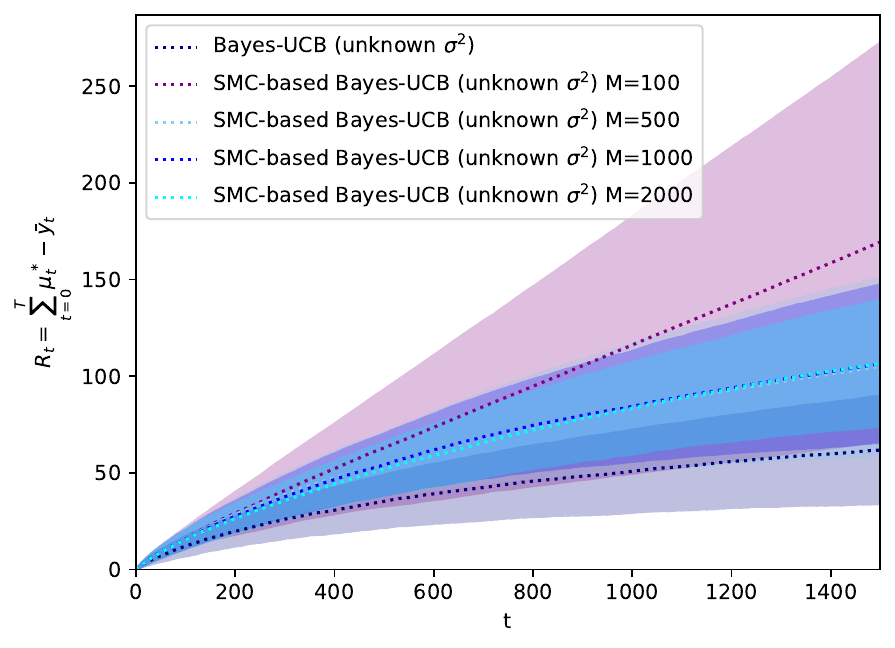}
		\caption{Unknown $\sigma$: SMC-based Bayes-UCB, impact of $M$.}
	\end{subfigure}
	
	\caption{Mean cumulative regret (standard deviation shown as the shaded region) of Bayesian policies in a stationary, five-armed contextual Gaussian bandit: $\theta_0=(-0.2,-0.2), \ \theta_1=(-0.1,-0.1), \ \theta_2=(0,0), \ \theta_3=(0.1,0.1), \ \theta_4=(0.2,0.2), \sigma^2=0.5$.}
\end{figure}

\clearpage

\subsection{Experiments with SMC-based Bayesian policies for logistic stationary bandits}
\label{assec:static_bandits_experiments_logistic}

Results in Sections~\ref{asssec:static_bandits_logistic_2} and \ref{asssec:static_bandits_logistic_5}
demonstrate how SMC-based Thompson sampling and Bayes-UCB achieve
successful exploration-exploitation tradeoff, 
for different parameterizations of stationary logistic bandits.
This evidence indicates that
the impact of observing context-dependent binary rewards of the played arms is minimal for the proposed SMC-based policies.

The parameter posterior uncertainty associated with SMC-based estimation
is automatically accounted for by both algorithms,
as they explore rarely-played arms if the uncertainty is high.
However, we observe a slight performance deterioration for SMC-based Bayes-UCB,
which we hypothesize is related to the quantile value used ($\alpha_t\propto1/t$).
This decay rate was justified by Kaufmann~\cite{ip-Kaufmann2012} for Bernoulli rewards,
but might not be optimal for other reward functions and,
more importantly, for the SMC-based parameter posterior random measures.

On the contrary, Thompson sampling automatically adjusts
to the uncertainty of the posterior random measure without extra hyperparameter search or tuning,
and attains reduced regret.

\subsubsection{Contextual logistic bandits, A=2}
\label{asssec:static_bandits_logistic_2}

We present below cumulative regret results for different parameterizations of 2-armed, contextual logistic bandits.

\begin{figure}[!h]
	\centering
	\begin{subfigure}[b]{\textwidth}
		\centering
		\includegraphics[width=0.75\textwidth]{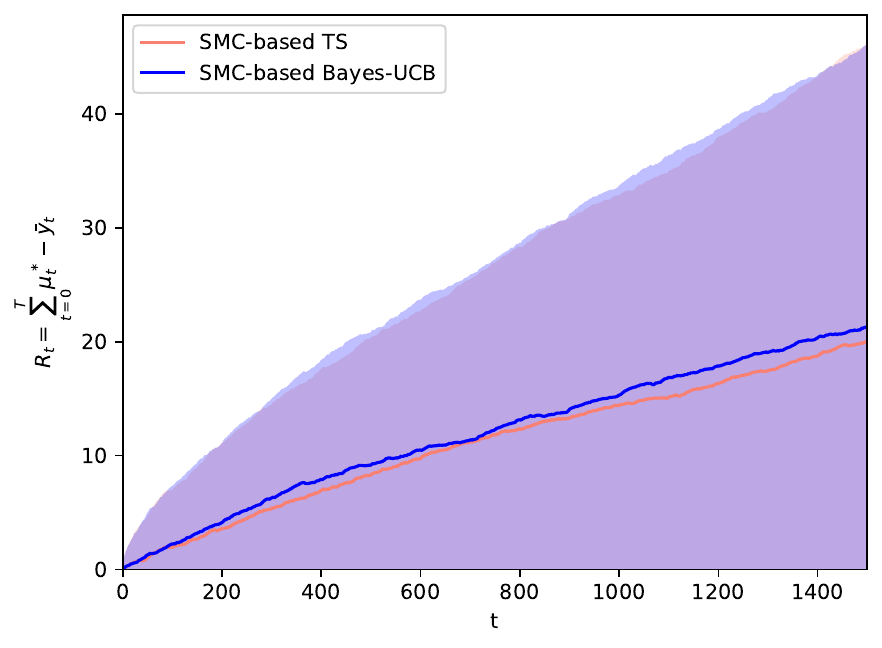}
		\caption{SMC-based ($M=1000$) TS and Bayes-UCB.}
	\end{subfigure}
	
	\begin{subfigure}[b]{0.46\textwidth}
		\centering
		\includegraphics[width=\textwidth]{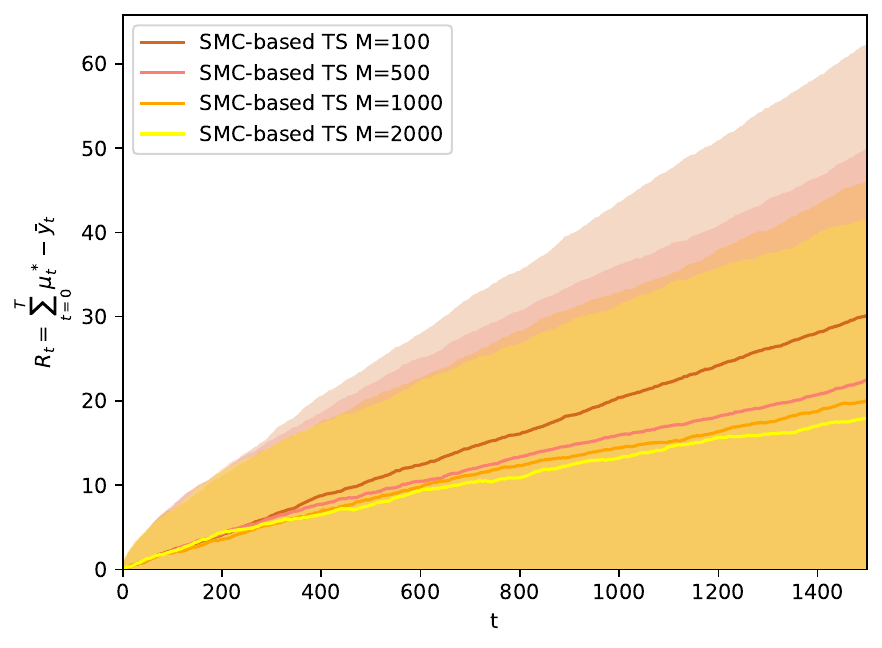}
		\caption{SMC-based TS: impact of $M$.}
	\end{subfigure}
	\begin{subfigure}[b]{0.46\textwidth}
		\centering
		\includegraphics[width=\textwidth]{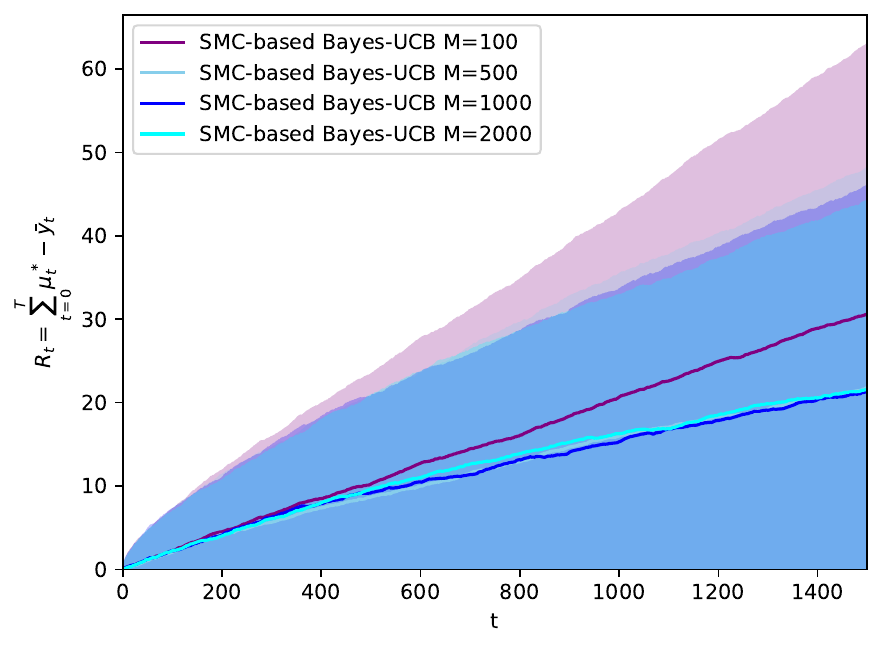}
		\caption{SMC-based Bayes-UCB: impact of $M$}
	\end{subfigure}
	
	\caption{Mean cumulative regret (standard deviation shown as the shaded region) of SMC-based Bayesian policies in
		stationary, two-armed contextual logistic bandits:
		$\theta_0=(-0.1,-0.1), \ \theta_1=(0.1,0.1)$.}
\end{figure}

\begin{figure}[!h]
	\centering
	\begin{subfigure}[b]{\textwidth}
		\centering
		\includegraphics[width=0.75\textwidth]{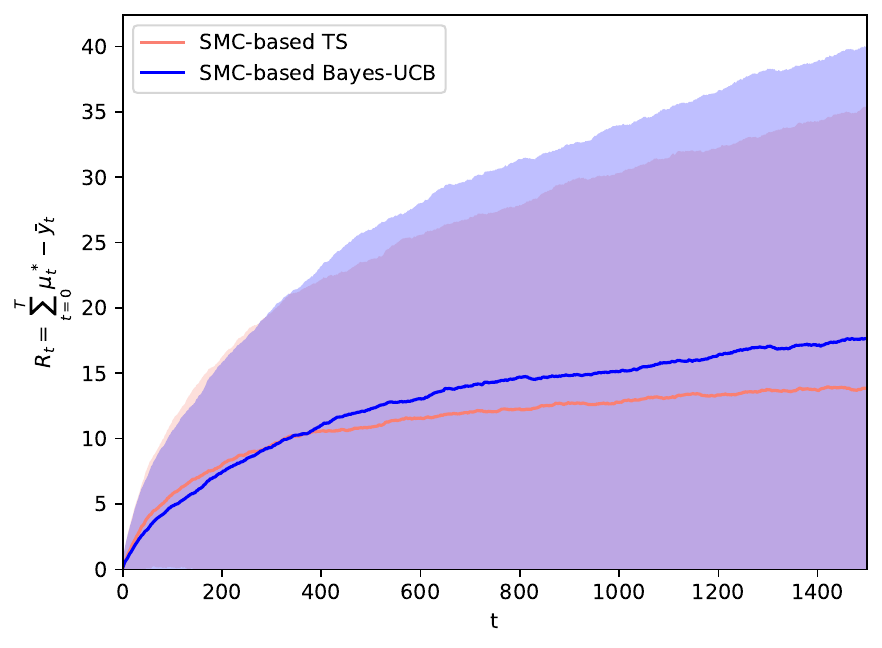}
		\caption{SMC-based ($M=1000$) TS and Bayes-UCB.}
	\end{subfigure}
	
	\begin{subfigure}[b]{0.46\textwidth}
		\centering
		\includegraphics[width=\textwidth]{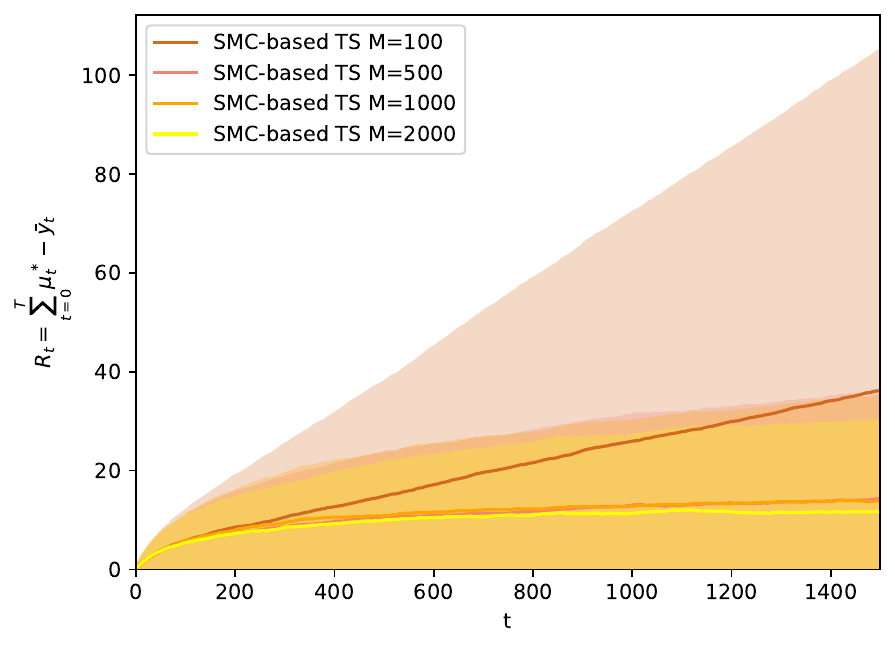}
		\caption{SMC-based TS: impact of $M$.}
	\end{subfigure}
	\begin{subfigure}[b]{0.46\textwidth}
		\centering
		\includegraphics[width=\textwidth]{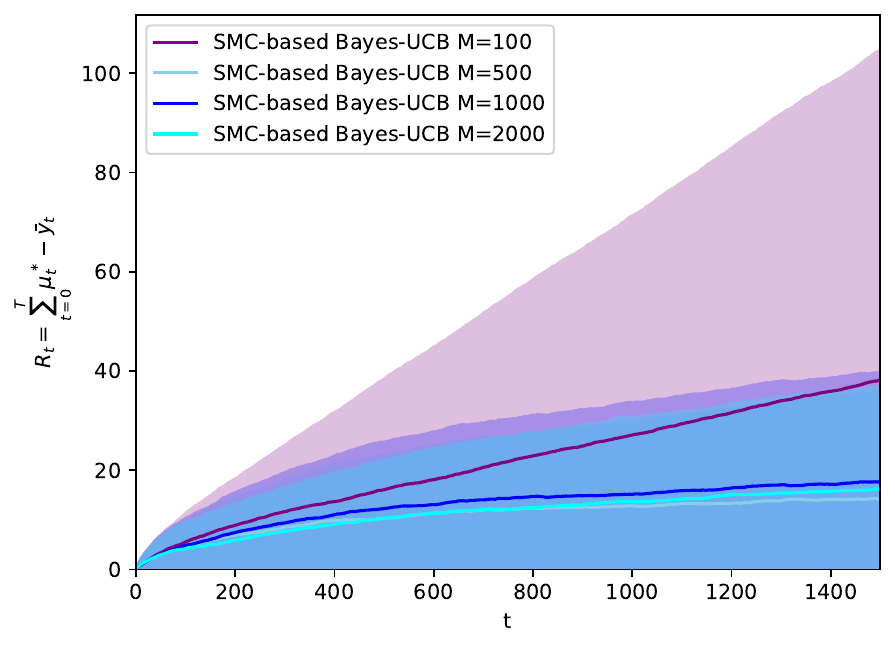}
		\caption{SMC-based Bayes-UCB: impact of $M$}
	\end{subfigure}
	
	\caption{Mean cumulative regret (standard deviation shown as the shaded region) of SMC-based Bayesian policies in
		stationary, two-armed contextual logistic bandits:
		$\theta_0=(-0.5,-0.5), \ \theta_1=(0.5,0.5)$.}
\end{figure}

\begin{figure}[!h]
	\centering
	\begin{subfigure}[b]{\textwidth}
		\centering
		\includegraphics[width=0.75\textwidth]{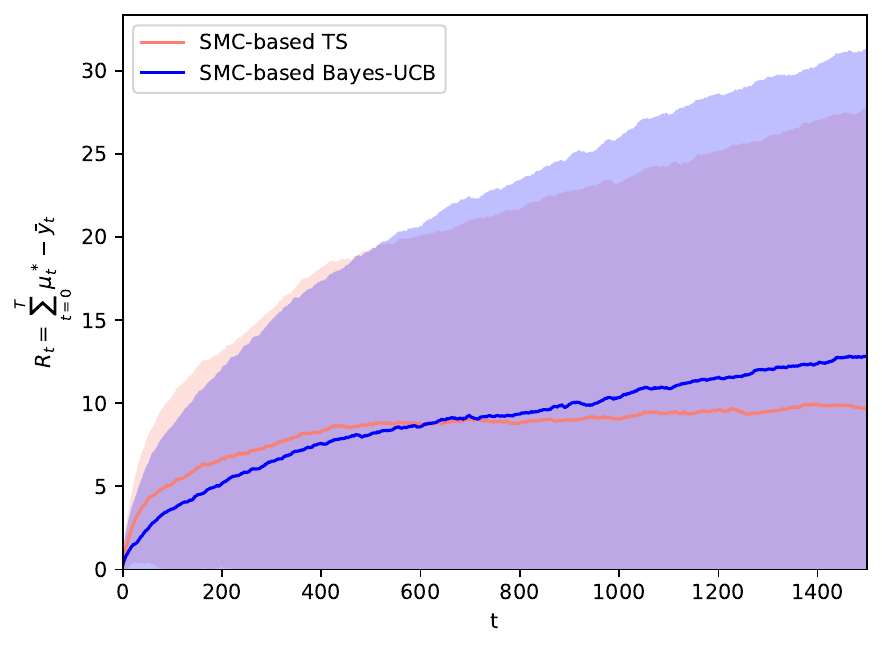}
		\caption{SMC-based ($M=1000$) TS and Bayes-UCB.}
	\end{subfigure}
	
	\begin{subfigure}[b]{0.46\textwidth}
		\centering
		\includegraphics[width=\textwidth]{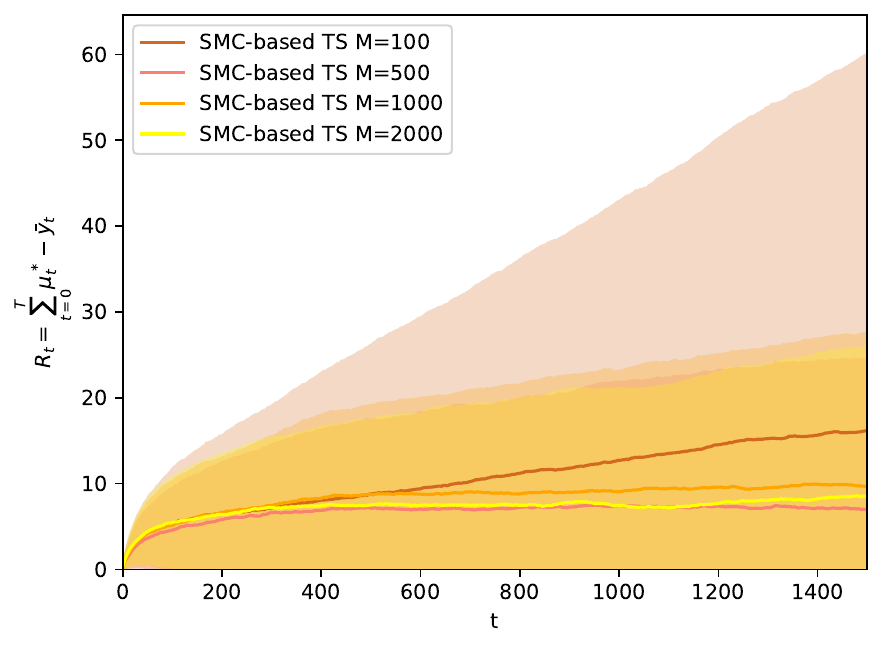}
		\caption{SMC-based TS: impact of $M$.}
	\end{subfigure}
	\begin{subfigure}[b]{0.46\textwidth}
		\centering
		\includegraphics[width=\textwidth]{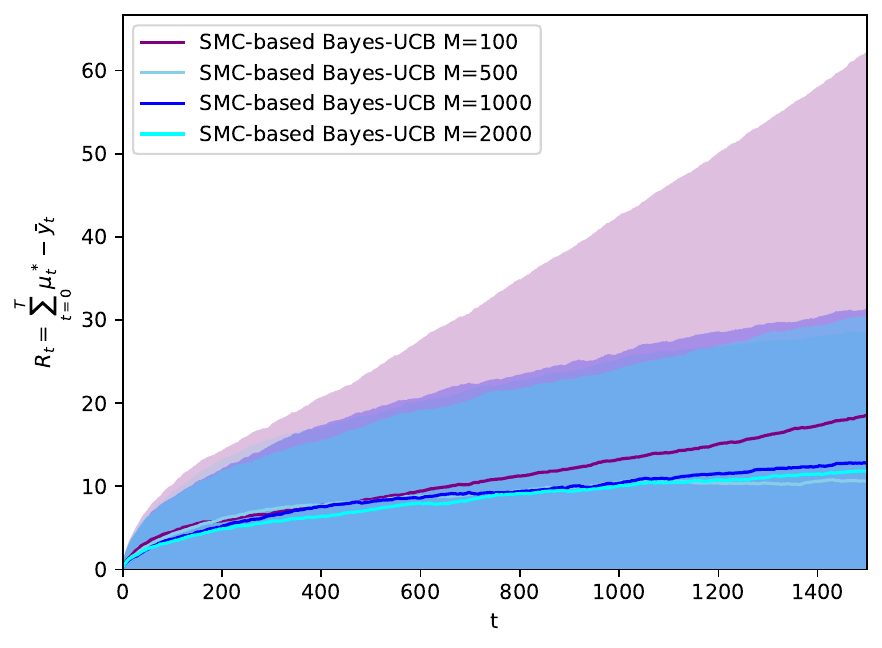}
		\caption{SMC-based Bayes-UCB: impact of $M$}
	\end{subfigure}
	
	\caption{Mean cumulative regret (standard deviation shown as the shaded region) of SMC-based Bayesian policies in
		stationary, two-armed contextual logistic bandits:
		$\theta_0=(-1.0,-1.0), \ \theta_1=(1.0,1.0)$.}
\end{figure}

\clearpage
\subsubsection{Contextual logistic bandits, A=5}
\label{asssec:static_bandits_logistic_5}

We present below cumulative regret results for different parameterizations of 5-armed, contextual logistic bandits.

\begin{figure}[!h]
	\centering
	\begin{subfigure}[b]{\textwidth}
		\centering
		\includegraphics[width=0.75\textwidth]{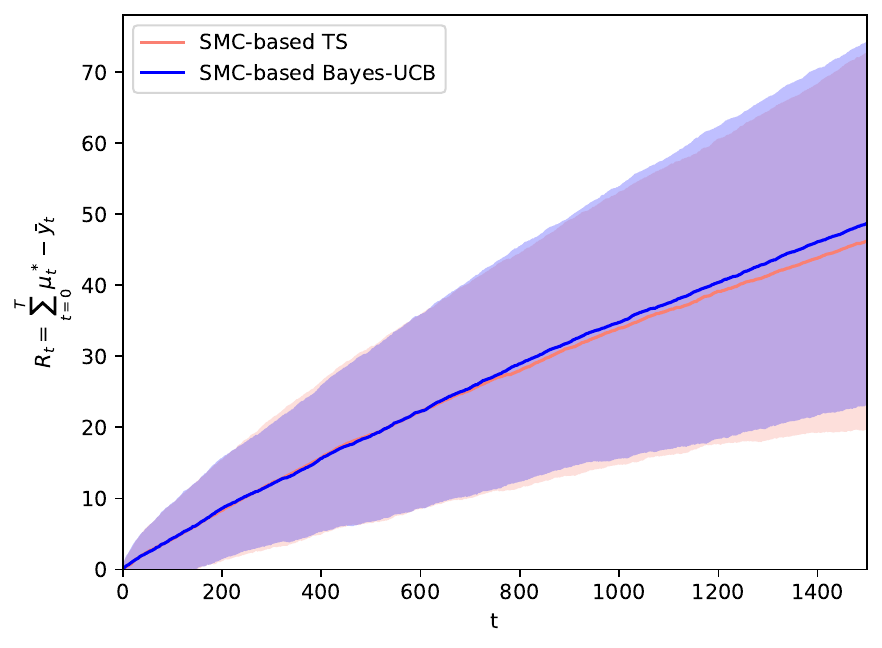}
		\caption{SMC-based ($M=1000$) TS and Bayes-UCB.}
	\end{subfigure}
	
	\begin{subfigure}[b]{0.46\textwidth}
		\centering
		\includegraphics[width=\textwidth]{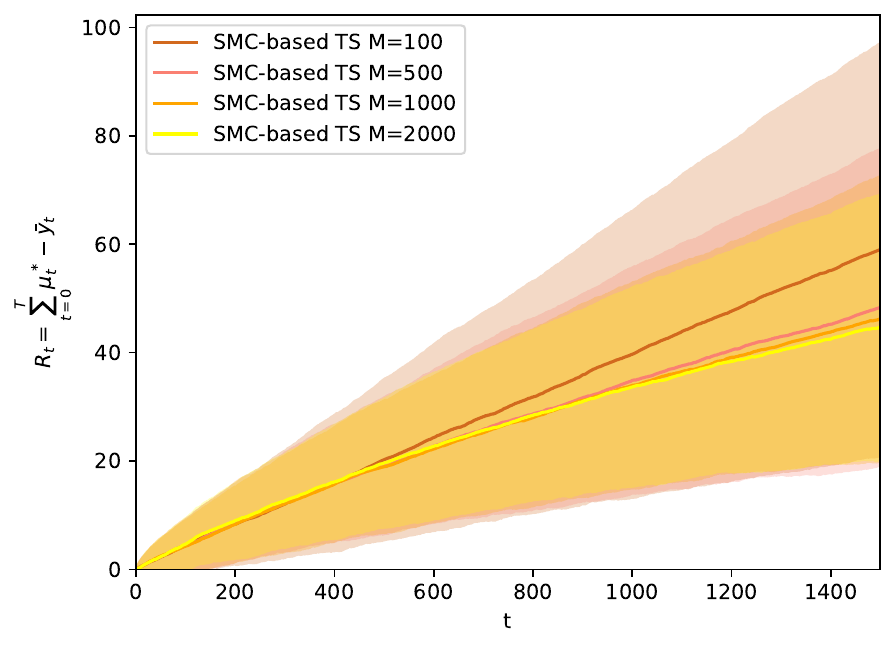}
		\caption{SMC-based TS: impact of $M$.}
	\end{subfigure}
	\begin{subfigure}[b]{0.46\textwidth}
		\centering
		\includegraphics[width=\textwidth]{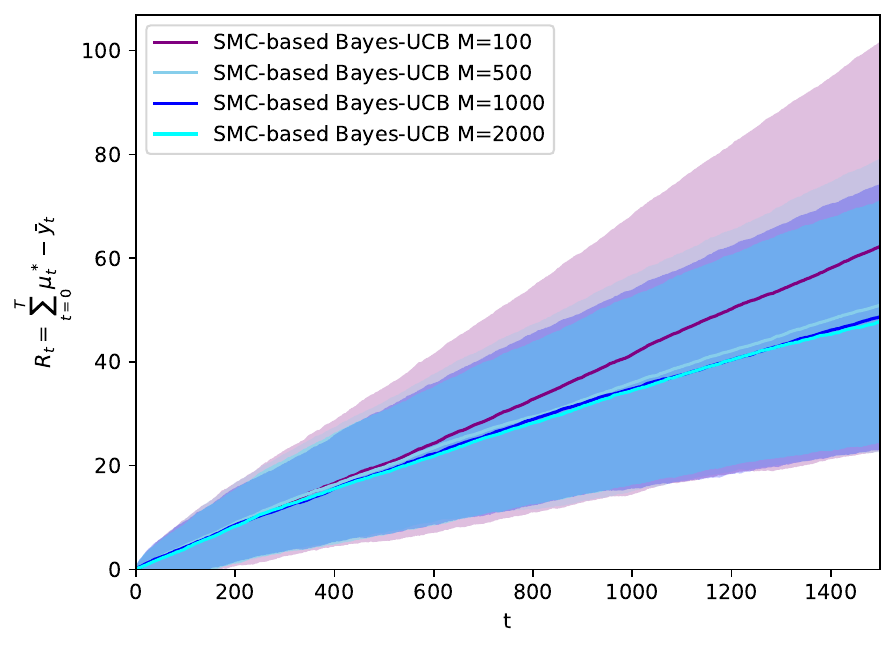}
		\caption{SMC-based Bayes-UCB: impact of $M$}
	\end{subfigure}
	
	\caption{Mean cumulative regret (standard deviation shown as the shaded region) of SMC-based Bayesian policies in
		stationary, five-armed contextual logistic bandits:
		$\theta_0=(-0.2,-0.2), \ \theta_1=(-0.1,-0.1), \ \theta_2=(0,0), \ \theta_3=(0.1,0.1), \ \theta_4=(0.2,0.2)$.
	}
\end{figure}

\begin{figure}[!h]
	\centering
	\begin{subfigure}[b]{\textwidth}
		\centering
		\includegraphics[width=0.75\textwidth]{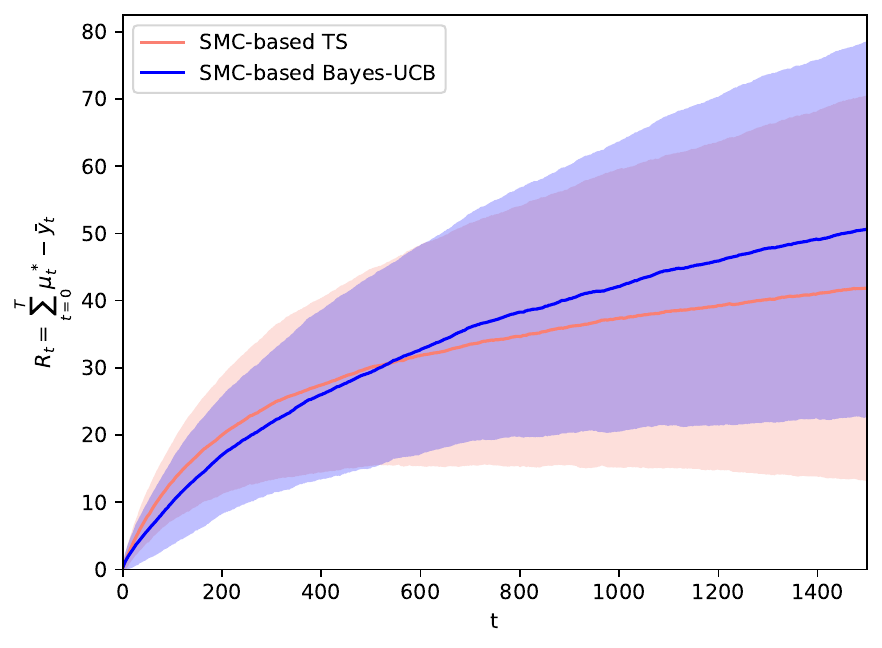}
		\caption{SMC-based ($M=1000$) TS and Bayes-UCB.}
	\end{subfigure}
	
	\begin{subfigure}[b]{0.46\textwidth}
		\centering
		\includegraphics[width=\textwidth]{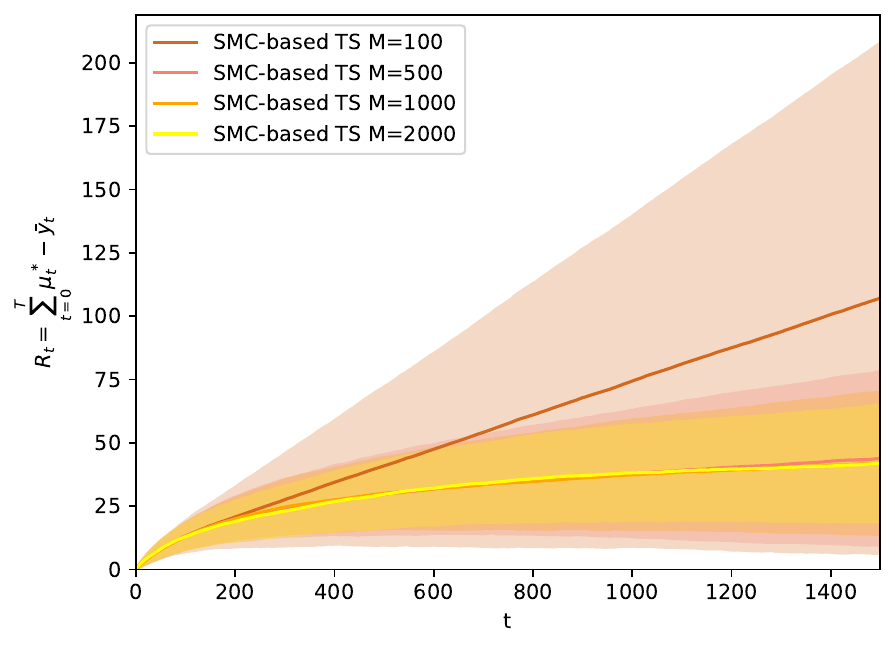}
		\caption{SMC-based TS: impact of $M$.}
	\end{subfigure}
	\begin{subfigure}[b]{0.46\textwidth}
		\centering
		\includegraphics[width=\textwidth]{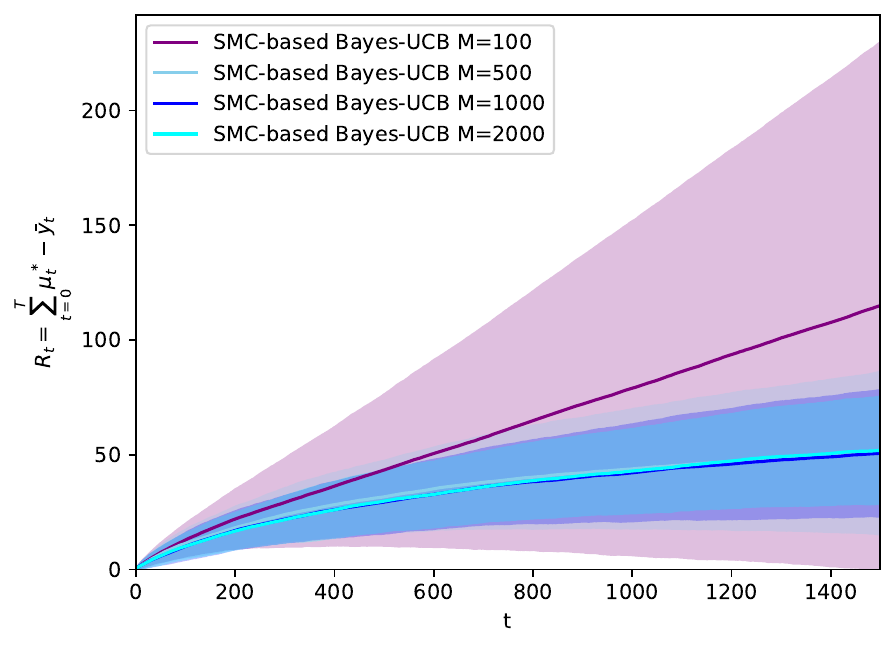}
		\caption{SMC-based Bayes-UCB: impact of $M$}
	\end{subfigure}
	
	\caption{Mean cumulative regret (standard deviation shown as the shaded region) of SMC-based Bayesian policies in
		stationary, five-armed contextual logistic bandits:
		$\theta_0=(-1.0,-1.0), \ \theta_1=(-0.5,-0.5), \ \theta_2=(0,0), \ \theta_3=(0.5,0.5), \ \theta_4=(1.0,1.0)$.
	}
\end{figure}

\begin{figure}[!h]
	\centering
	\begin{subfigure}[b]{\textwidth}
		\centering
		\includegraphics[width=0.75\textwidth]{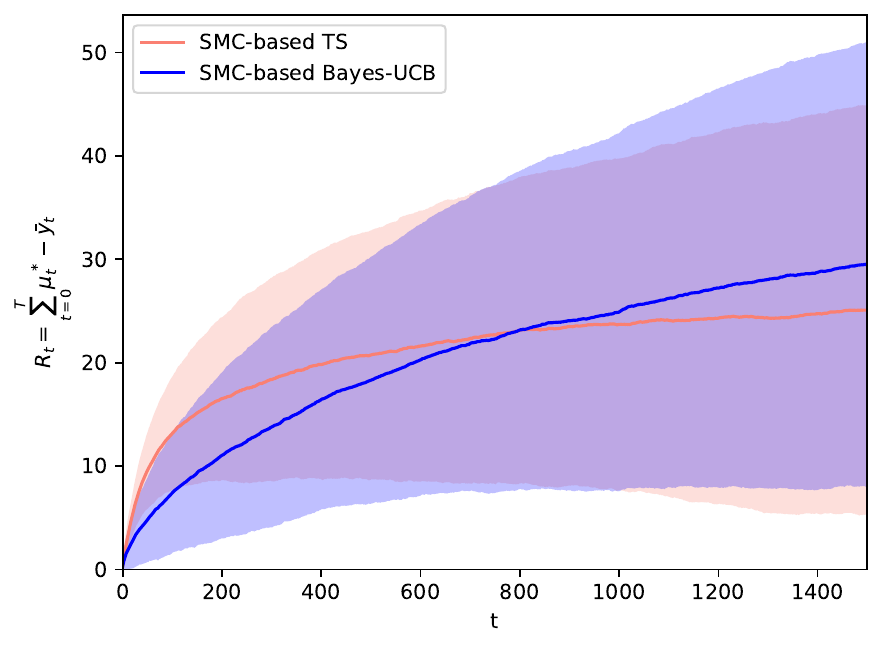}
		\caption{SMC-based ($M=1000$) TS and Bayes-UCB.}
	\end{subfigure}
	
	\begin{subfigure}[b]{0.46\textwidth}
		\centering
		\includegraphics[width=\textwidth]{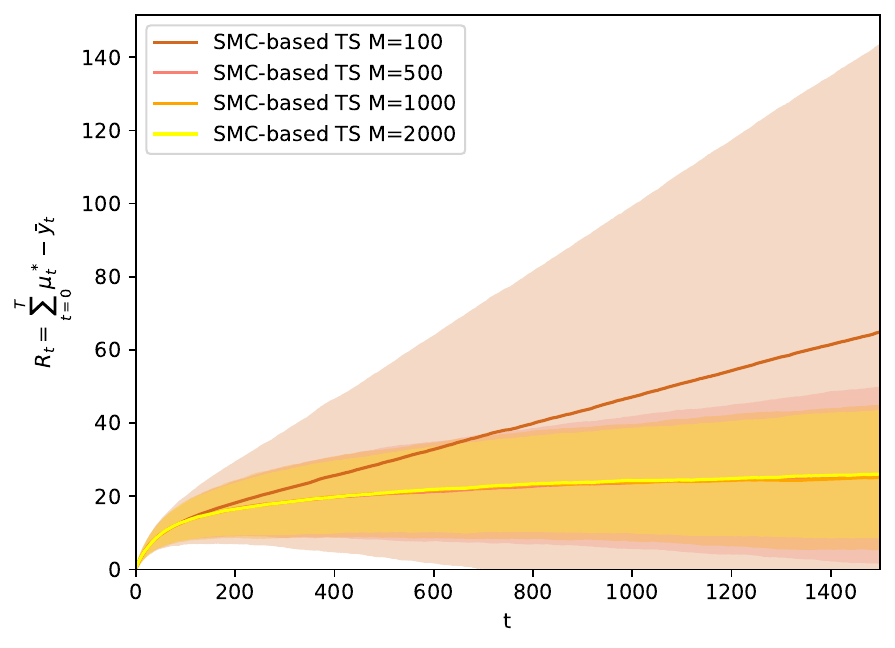}
		\caption{SMC-based TS: impact of $M$.}
	\end{subfigure}
	\begin{subfigure}[b]{0.46\textwidth}
		\centering
		\includegraphics[width=\textwidth]{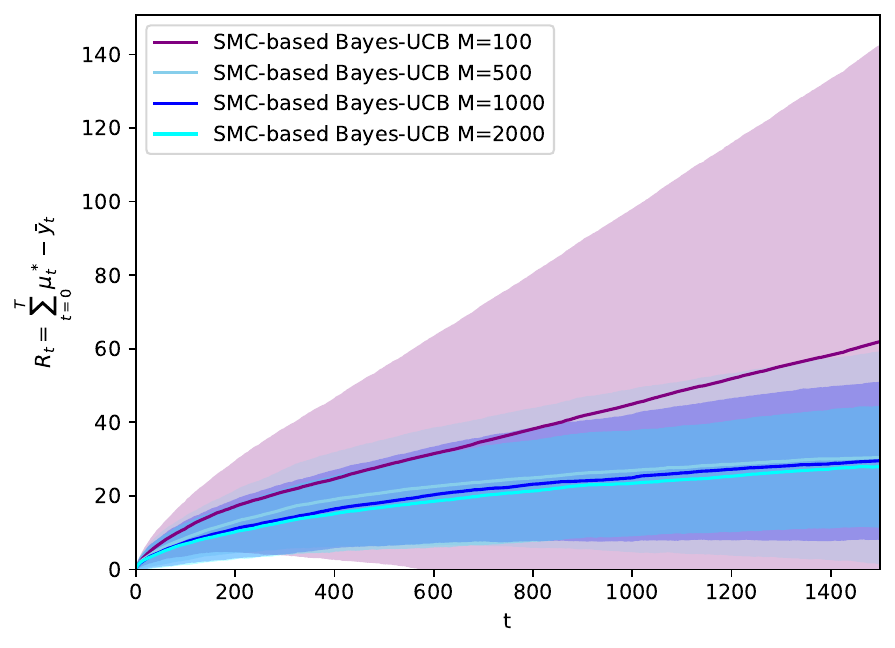}
		\caption{SMC-based Bayes-UCB: impact of $M$}
	\end{subfigure}
	
	\caption{Mean cumulative regret (standard deviation shown as the shaded region) of SMC-based Bayesian policies in
		stationary, five-armed contextual logistic bandits:
		$\theta_0=(-2.0,-2.0), \ \theta_1=(-1.0,-1.0), \ \theta_2=(0,0), \ \theta_3=(1.0,1.0), \ \theta_4=(2.0,2.0)$.
	}
\end{figure}

\clearpage
\section{SMC-based policies in non-stationary bandits}
\label{asec:dynamic_bandits}

We provide additional results below to assess 
the impact of different Monte Carlo sample size $M$ in SMC-based bandit policies, across the studied non-stationary environments.

\subsection{Non-stationary, linear Gaussian rewards}
\label{assec:dynamic_bandits_gaussian}

\begin{figure}[!h]
	\centering
	\begin{subfigure}[b]{0.45\textwidth}
		\includegraphics[width=\textwidth]{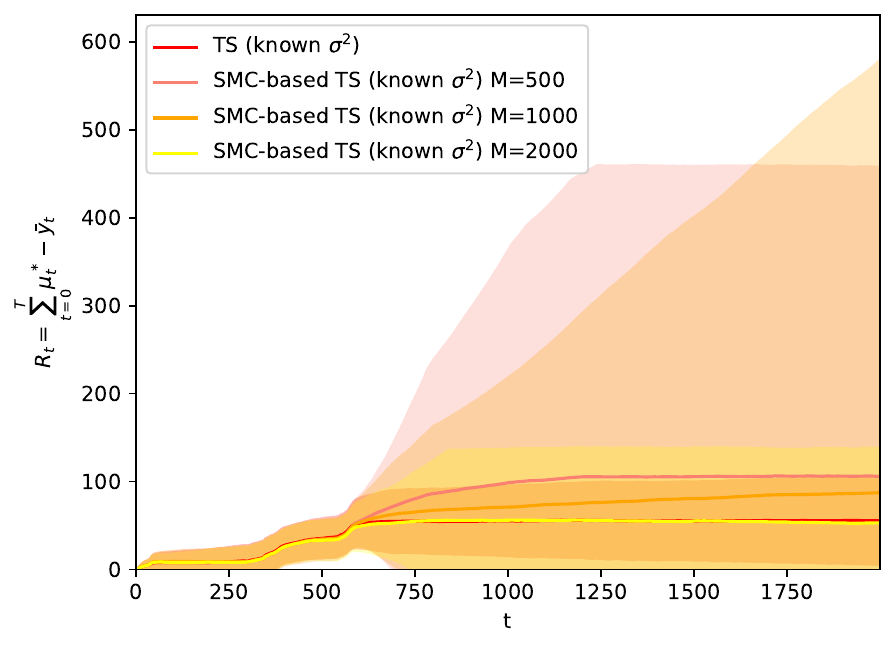}
		\caption{Cumulative regret for SMC-based TS in scenario A: known dynamic parameters.}
		\label{fig:dynamic_bandits_linearGaussian_a_ts_dknown_knownsigma_M}
	\end{subfigure}\qquad
	\begin{subfigure}[b]{0.45\textwidth}
		\includegraphics[width=\textwidth]{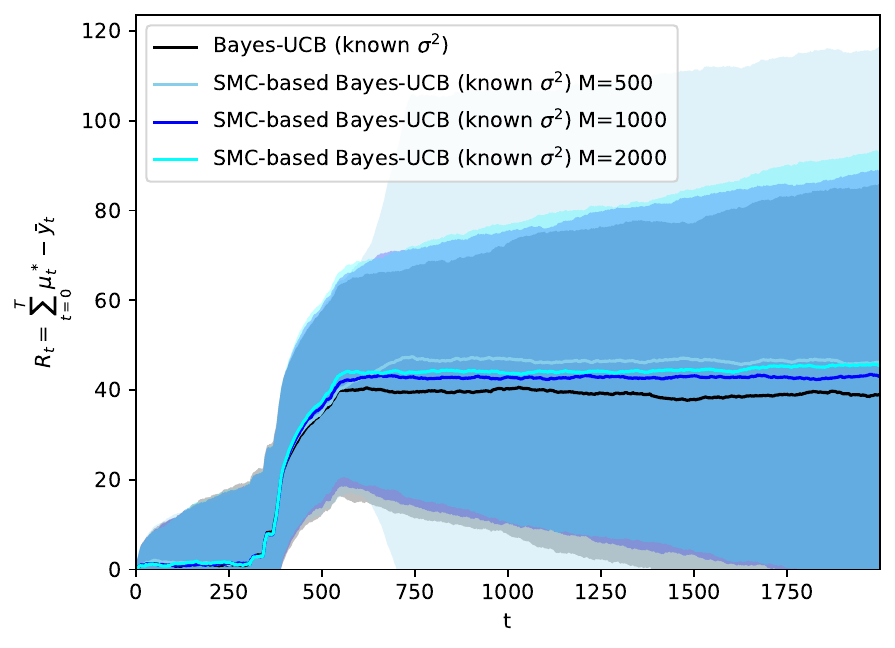}
		\caption{Cumulative regret for SMC-based Bayes-UCB in scenario A: known dynamic parameters.}
		\label{fig:dynamic_bandits_linearGaussian_a_bucb_dknown_knownsigma_M}
	\end{subfigure}
	
	\begin{subfigure}[b]{0.45\textwidth}
		\includegraphics[width=\textwidth]{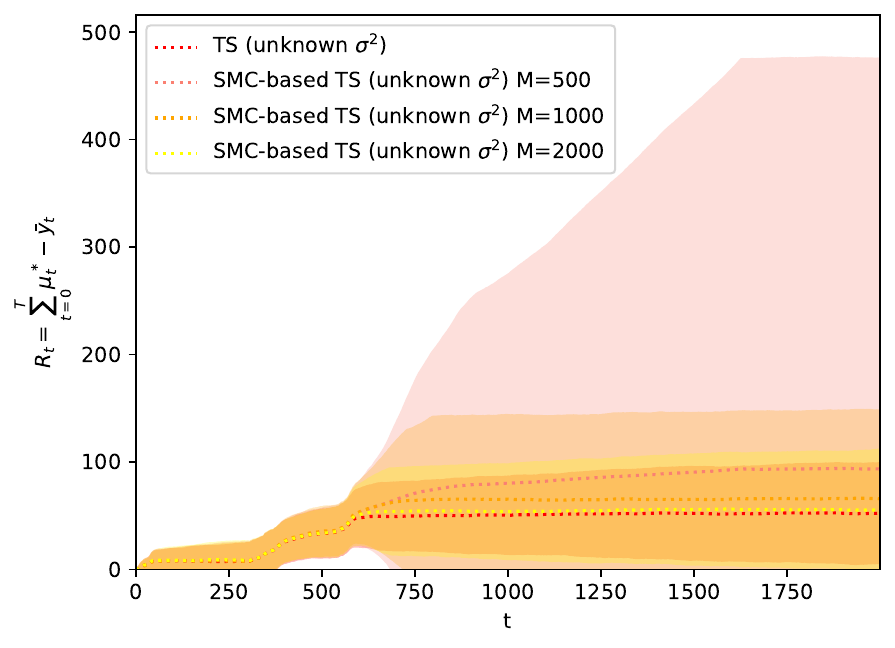}
		\caption{Cumulative regret for SMC-based TS in scenario A: known dynamic parameters, unknown $\sigma_a^2, \forall a$.}
		\label{fig:dynamic_bandits_linearGaussian_a_ts_dknown_unknownsigma_M}
	\end{subfigure}\qquad
	\begin{subfigure}[b]{0.45\textwidth}
		\includegraphics[width=\textwidth]{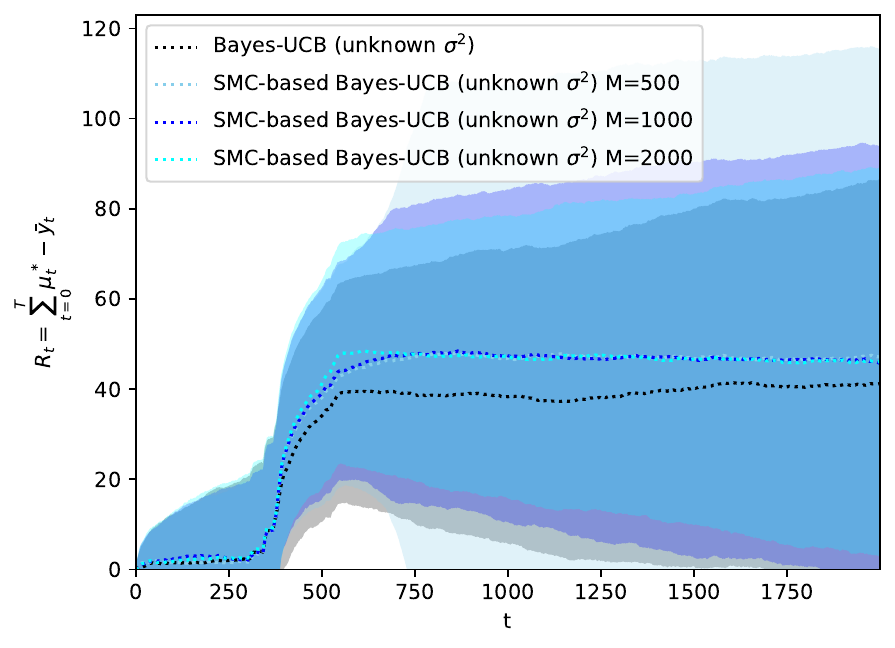}
		\caption{Cumulative regret for SMC-based Bayes-UCB in scenario A: known dynamic parameters, unknown $\sigma_a^2, \forall a$.}
		\label{fig:dynamic_bandits_linearGaussian_a_bucb_dknown_unknownsigma_M}
	\end{subfigure}
	
		\begin{subfigure}[b]{0.45\textwidth}
		\includegraphics[width=\textwidth]{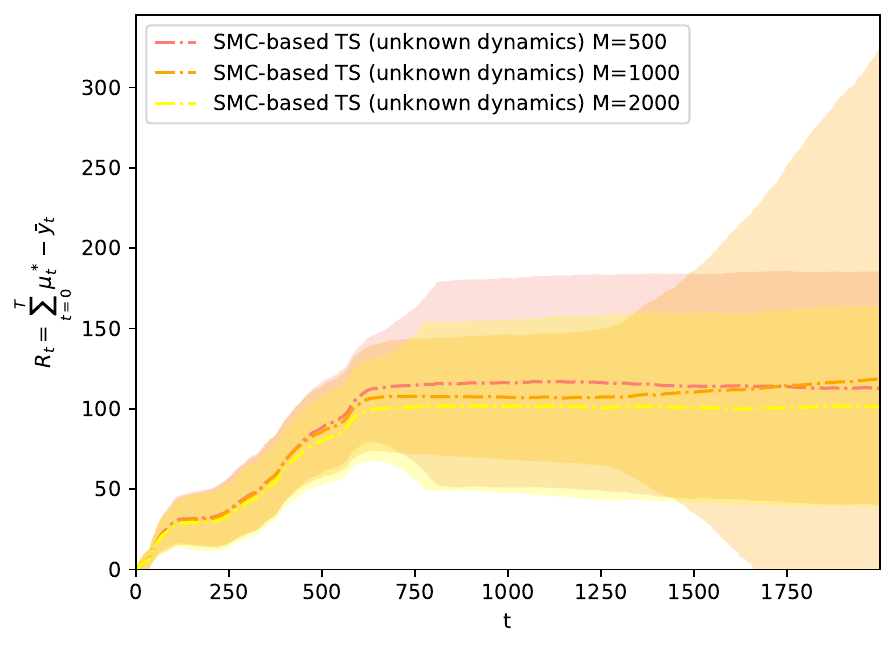}
		\caption{Cumulative regret for SMC-based TS in scenario A: unknown dynamic parameters $L_a,\Sigma_a,\sigma_a^2, \forall a$.}
		\label{fig:dynamic_bandits_linearGaussian_a_ts_dunknown_M}
	\end{subfigure}\qquad
	\begin{subfigure}[b]{0.45\textwidth}
		\includegraphics[width=\textwidth]{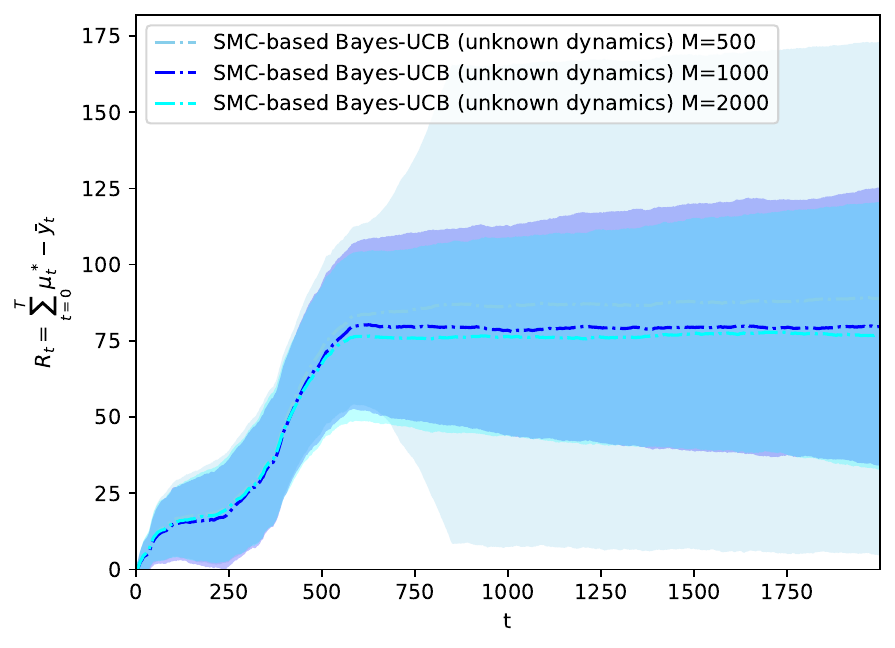}
		\caption{Cumulative regret for SMC-based Bayes-UCB in scenario A: unknown dynamic parameters $L_a,\Sigma_a,\sigma_a^2, \forall a$.}
		\label{fig:dynamic_bandits_linearGaussian_a_bucb_dunknown_M}
	\end{subfigure}

	\caption{
		Mean regret (standard deviation shown as the shaded region) in contextual, linear Gaussian bandit Scenario A
		described in Equation~\eqref{eq:linear_mixing_dynamics_a}.
		SMC-based policies' averaged cumulative regret is robust to different Monte Carlo sample sizes $M$,
		which impacts mostly the performance variability for $M=500$. 
	}
	\label{fig:dynamic_bandits_linearGaussian_a_M}
\end{figure}

\begin{figure}[!h]
	\centering
	\begin{subfigure}[b]{0.45\textwidth}
		\includegraphics[width=\textwidth]{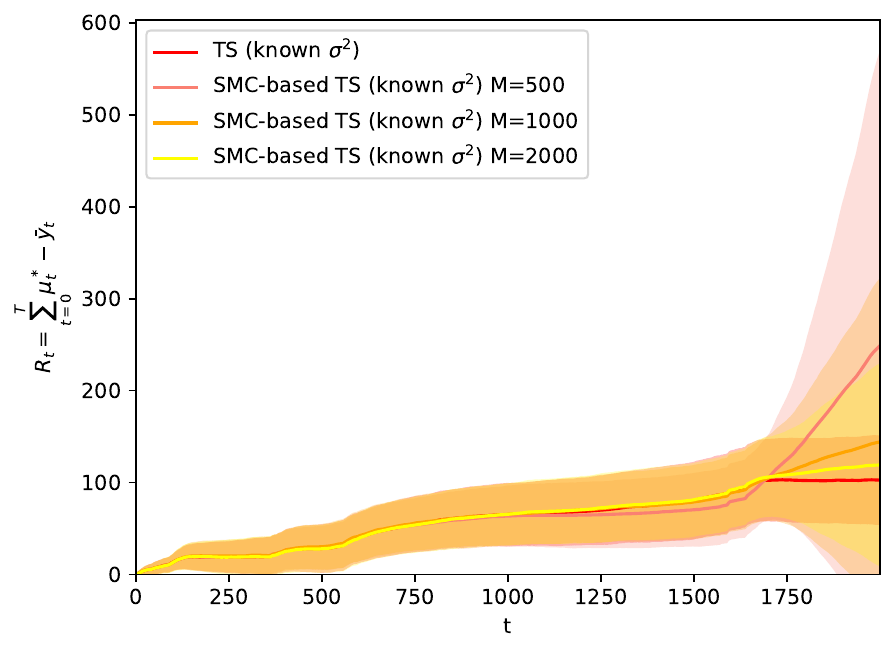}
		\caption{Cumulative regret for SMC-based TS in scenario B: known dynamic parameters.}
		\label{fig:dynamic_bandits_linearGaussian_b_ts_dknown_knownsigma_M}
	\end{subfigure}\qquad
	\begin{subfigure}[b]{0.45\textwidth}
		\includegraphics[width=\textwidth]{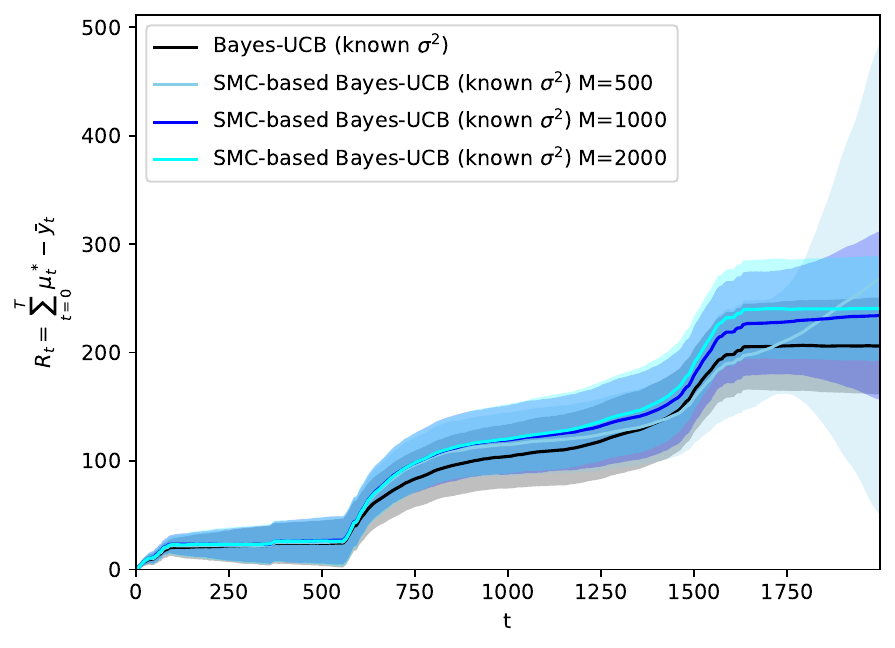}
		\caption{Cumulative regret for SMC-based Bayes-UCB in scenario B: known dynamic parameters.}
		\label{fig:dynamic_bandits_linearGaussian_b_bucb_dknown_knownsigma_M}
	\end{subfigure}
	
	\begin{subfigure}[b]{0.45\textwidth}
		\includegraphics[width=\textwidth]{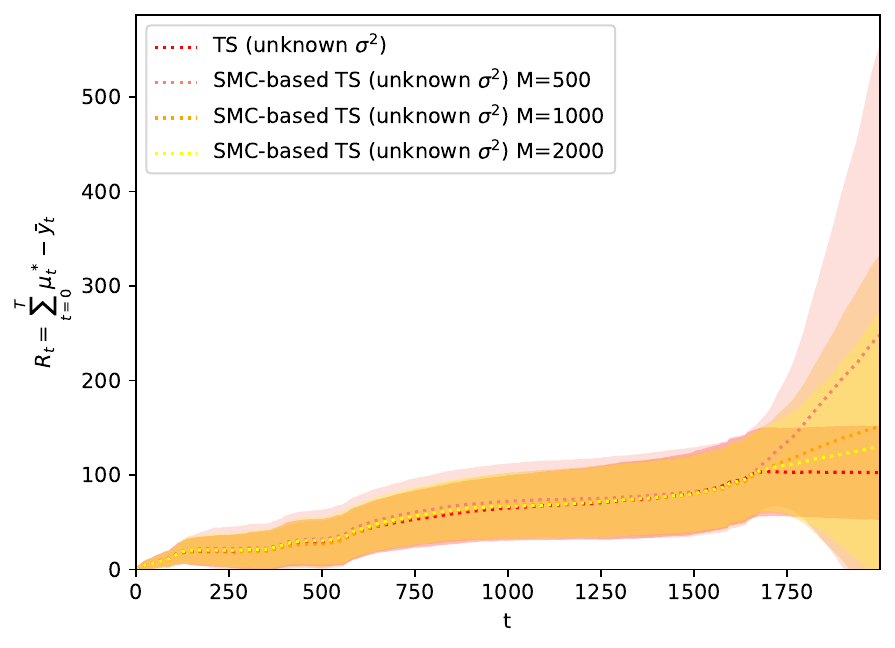}
		\caption{Cumulative regret for SMC-based TS in scenario B: known dynamic parameters, unknown $\sigma_a^2, \forall a$.}
		\label{fig:dynamic_bandits_linearGaussian_b_ts_dknown_unknownsigma_M}
	\end{subfigure}\qquad
	\begin{subfigure}[b]{0.45\textwidth}
		\includegraphics[width=\textwidth]{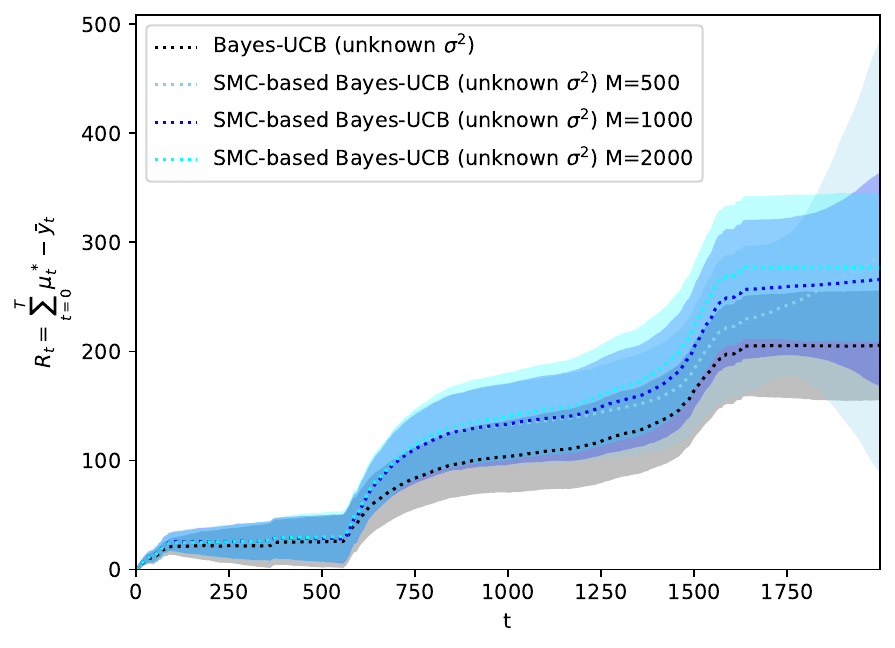}
		\caption{Cumulative regret for SMC-based Bayes-UCB in scenario B: known dynamic parameters, unknown $\sigma_a^2, \forall a$.}
		\label{fig:dynamic_bandits_linearGaussian_b_bucb_dknown_unknownsigma_M}
	\end{subfigure}
	
	\begin{subfigure}[b]{0.45\textwidth}
		\includegraphics[width=\textwidth]{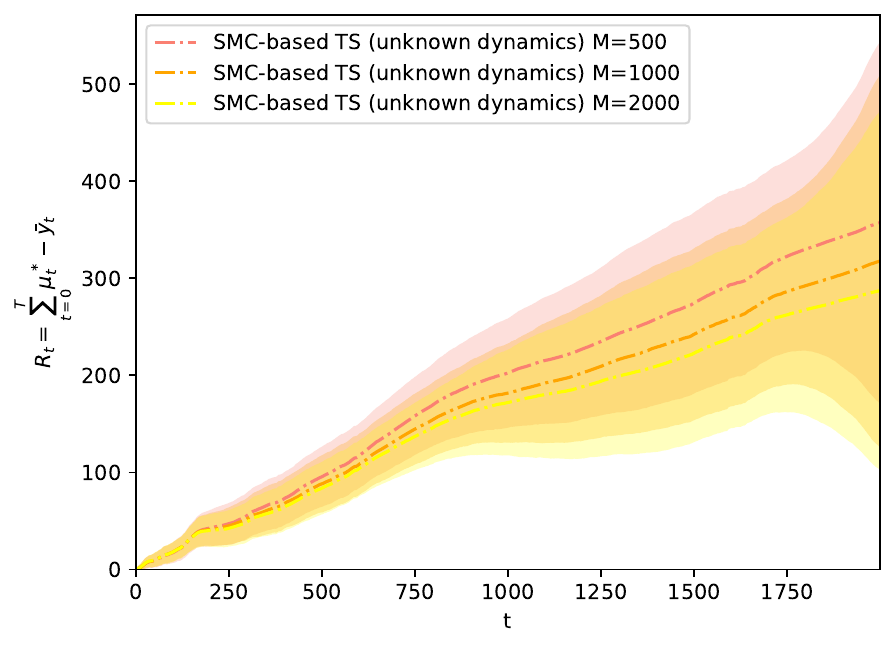}
		\caption{Cumulative regret for SMC-based TS in scenario B: unknown dynamic parameters $L_a,\Sigma_a,\sigma_a^2, \forall a$.}
		\label{fig:dynamic_bandits_linearGaussian_b_ts_dunknown_M}
	\end{subfigure}\qquad
	\begin{subfigure}[b]{0.45\textwidth}
		\includegraphics[width=\textwidth]{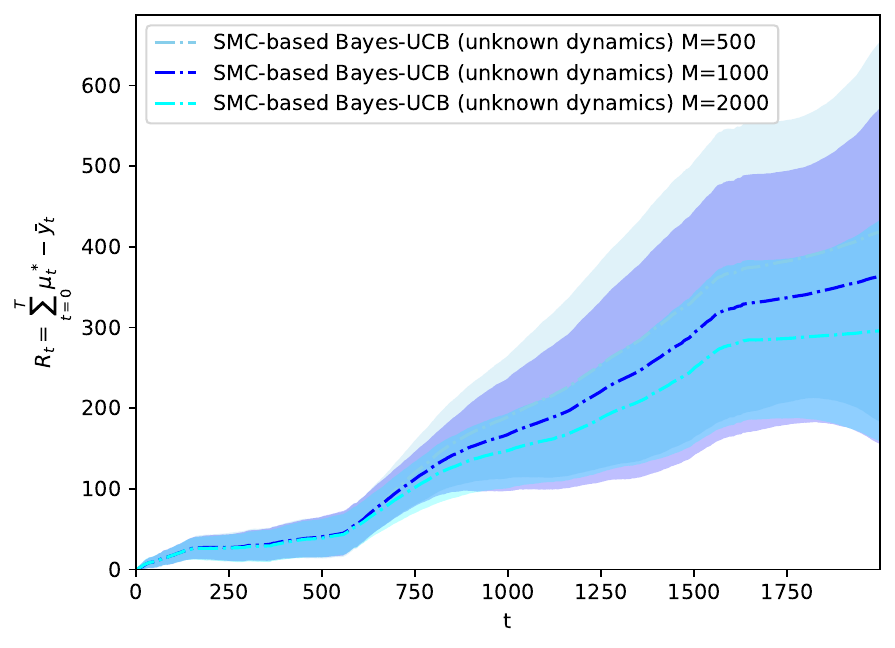}
		\caption{Cumulative regret for SMC-based Bayes-UCB in scenario B: unknown dynamic parameters $L_a,\Sigma_a,\sigma_a^2, \forall a$.}
		\label{fig:dynamic_bandits_linearGaussian_b_bucb_dunknown_M}
	\end{subfigure}
	
	\caption{
		Mean regret (standard deviation shown as the shaded region) in contextual, linear Gaussian bandit Scenario B
		described in Equation~\eqref{eq:linear_mixing_dynamics_b}.
		SMC-based policies' averaged cumulative regret is robust to different Monte Carlo sample sizes $M$,
		which impacts mostly the performance variability for $M=500$ ---specially so when optimal arm swaps occur later in time.
	}
	\label{fig:dynamic_bandits_linearGaussian_b_M}
\end{figure}

We assess the impact of Monte Carlo sample size $M$ in the performance of the proposed SMC-based Bayesian MAB policies:
In Figure~\ref{fig:dynamic_bandits_linearGaussian_a_M}, we present results for Scenario A defined by Equation~\eqref{eq:linear_mixing_dynamics_a},
for a realization of expected rewards as depicted in Figure~\ref{fig:linear_mixing_dynamics_a_gaussian};
In Figure~\ref{fig:dynamic_bandits_linearGaussian_b_M}, we  present results for Scenario B defined by Equation~\eqref{eq:linear_mixing_dynamics_b}, 
with a realization of expected rewards as depicted in Figure~\ref{fig:linear_mixing_dynamics_b_gaussian}.

\clearpage
\subsection{Non-stationary, logistic rewards}
\label{assec:dynamic_bandits_logistic}

\begin{figure}[!h]
	\centering
	\begin{subfigure}[b]{0.45\textwidth}
		\includegraphics[width=\textwidth]{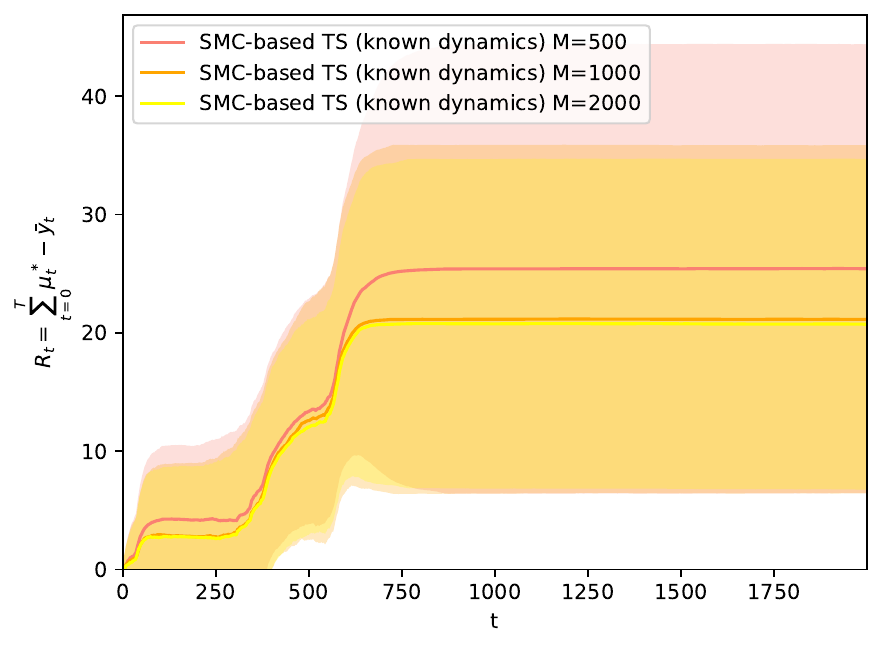}
		\caption{Cumulative regret for SMC-based TS in scenario C: known dynamic parameters.}
		\label{fig:dynamic_bandits_logistic_c_ts_dknown_M}
	\end{subfigure}\qquad
	\begin{subfigure}[b]{0.45\textwidth}
		\includegraphics[width=\textwidth]{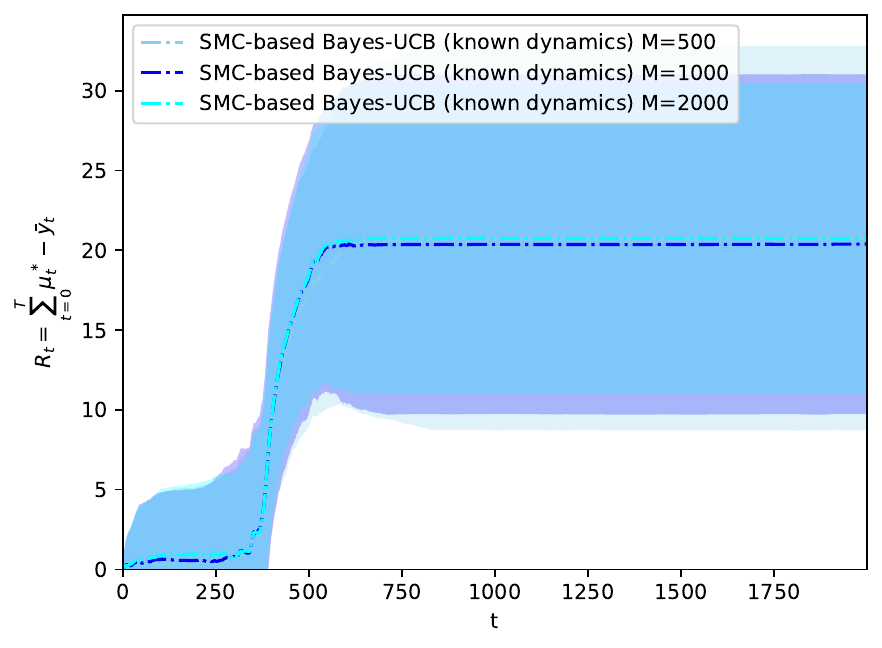}
		\caption{Cumulative regret for SMC-based Bayes-UCB in scenario C: known dynamic parameters.}
		\label{fig:dynamic_bandits_logistic_c_bucb_dknown_M}
	\end{subfigure}
	
	\begin{subfigure}[b]{0.45\textwidth}
		\includegraphics[width=\textwidth]{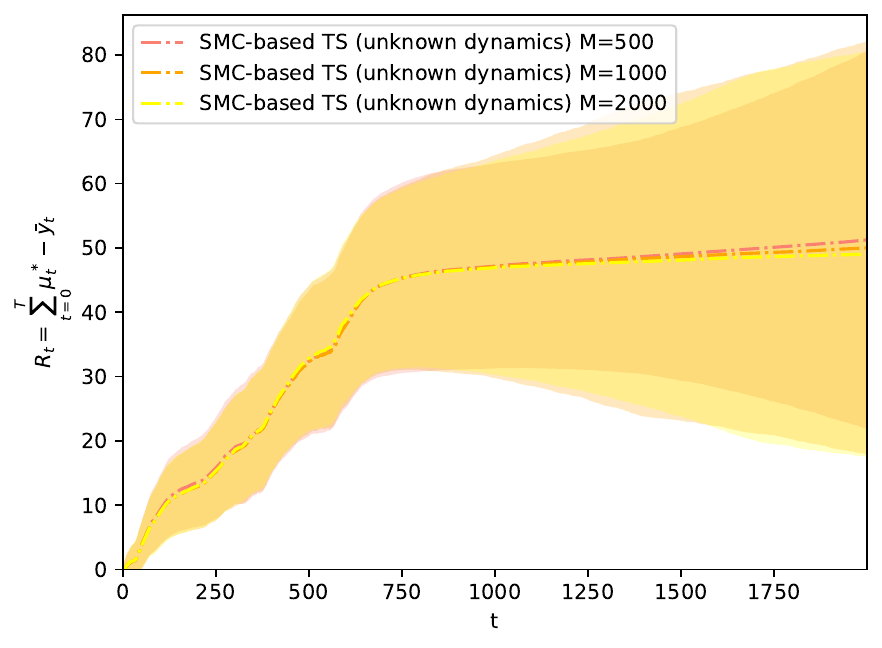}
		\caption{Cumulative regret for SMC-based TS in scenario C: unknown dynamic parameters $L_a,\Sigma_a,\sigma_a^2, \forall a$.}
		\label{fig:dynamic_bandits_logistic_c_ts_dunknown_M}
	\end{subfigure}\qquad
	\begin{subfigure}[b]{0.45\textwidth}
		\includegraphics[width=\textwidth]{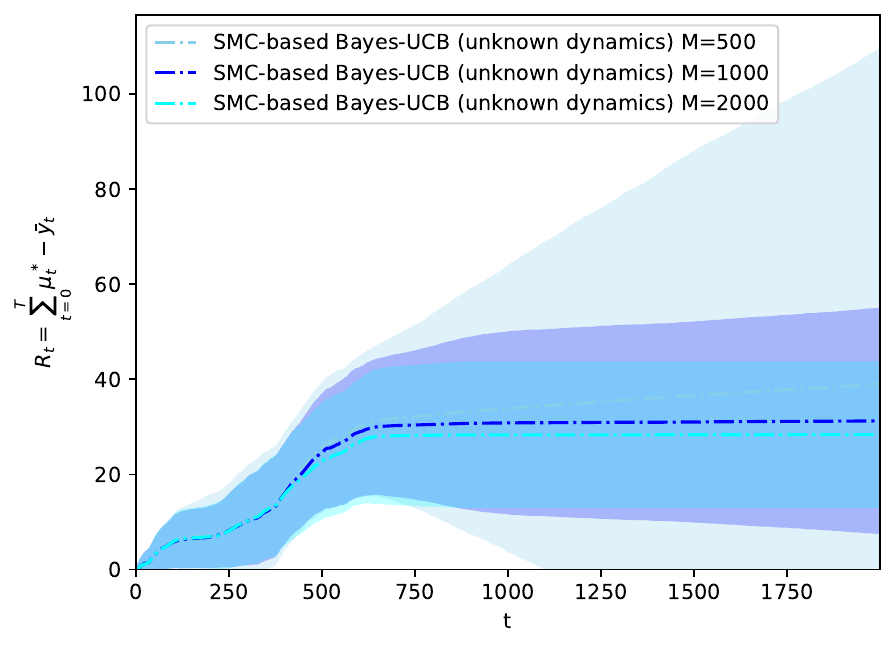}
		\caption{Cumulative regret for SMC-based Bayes-UCB in scenario C: unknown dynamic parameters $L_a,\Sigma_a,\sigma_a^2, \forall a$.}
		\label{fig:dynamic_bandits_logistic_c_bucb_dunknown_M}
	\end{subfigure}
	
	\caption{
		Mean regret (standard deviation shown as the shaded region) in contextual, logistic bandit Scenario C
		described in Equation~\eqref{eq:linear_mixing_dynamics_c}.
		SMC-based policies' averaged cumulative regret is robust to different Monte Carlo sample sizes $M$,
		which impacts mostly the performance variability for $M=500$. 
	}
	\label{fig:dynamic_bandits_logistic_c_M}
\end{figure}

We assess in Figure~\ref{fig:dynamic_bandits_logistic_c_M} the impact of Monte Carlo sample size $M$ in the performance of the proposed SMC-based Bayesian MAB policies
in Scenario C defined by Equation~\eqref{eq:linear_mixing_dynamics_c},
for a realization of expected rewards as depicted in Figure~\ref{fig:linear_mixing_dynamics_c_logistic}.

\begin{figure}[!h]
	\centering
	\begin{subfigure}[b]{0.45\textwidth}
		\includegraphics[width=\textwidth]{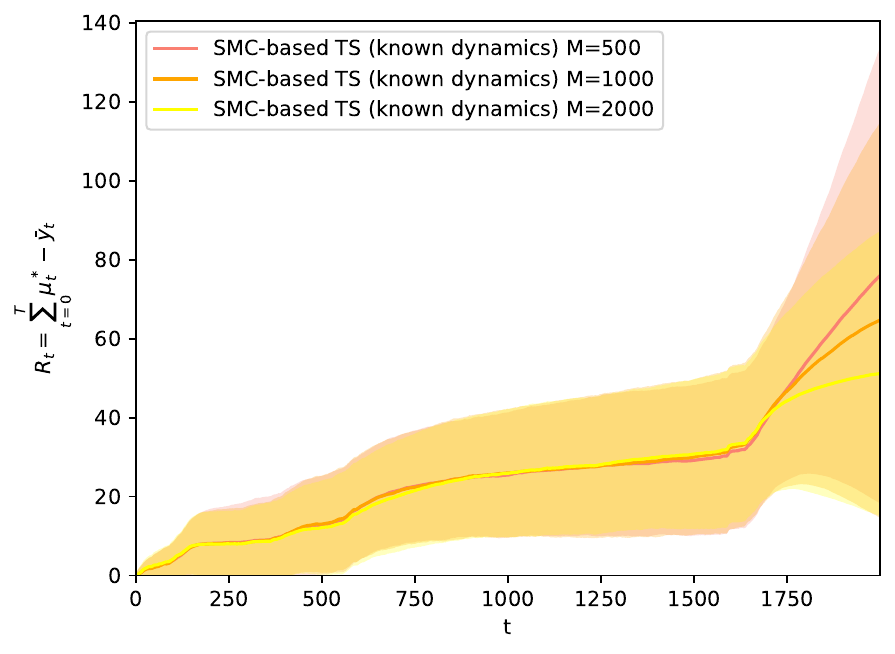}
		\caption{Cumulative regret for SMC-based TS in scenario D: known dynamic parameters.}
		\label{fig:dynamic_bandits_logistic_d_ts_dknown_M}
	\end{subfigure}\qquad
	\begin{subfigure}[b]{0.45\textwidth}
		\includegraphics[width=\textwidth]{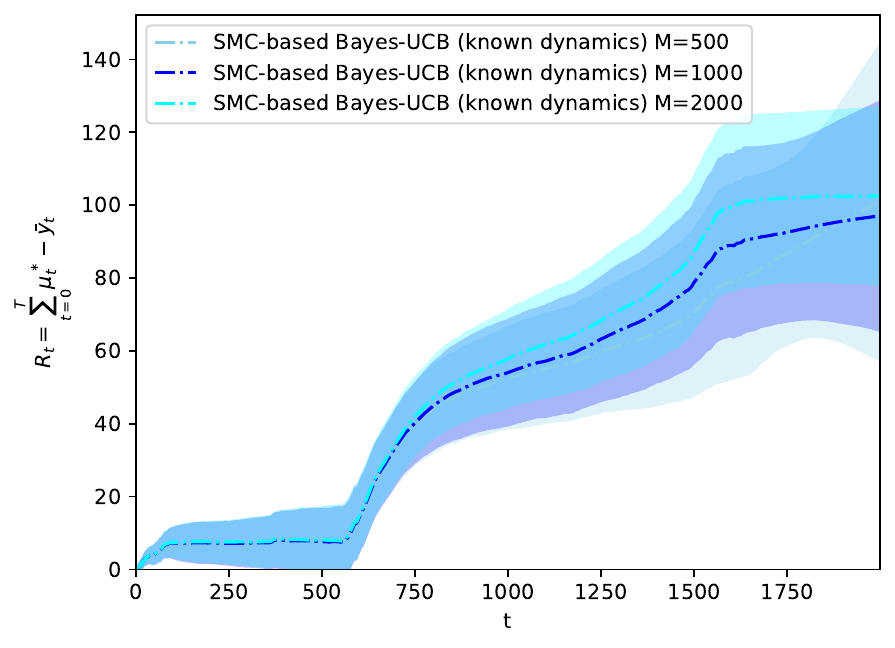}
		\caption{Dumulative regret for SMC-based Bayes-UCB in scenario D: known dynamic parameters.}
		\label{fig:dynamic_bandits_logistic_d_bucb_dknown_M}
	\end{subfigure}
	
	\begin{subfigure}[b]{0.45\textwidth}
		\includegraphics[width=\textwidth]{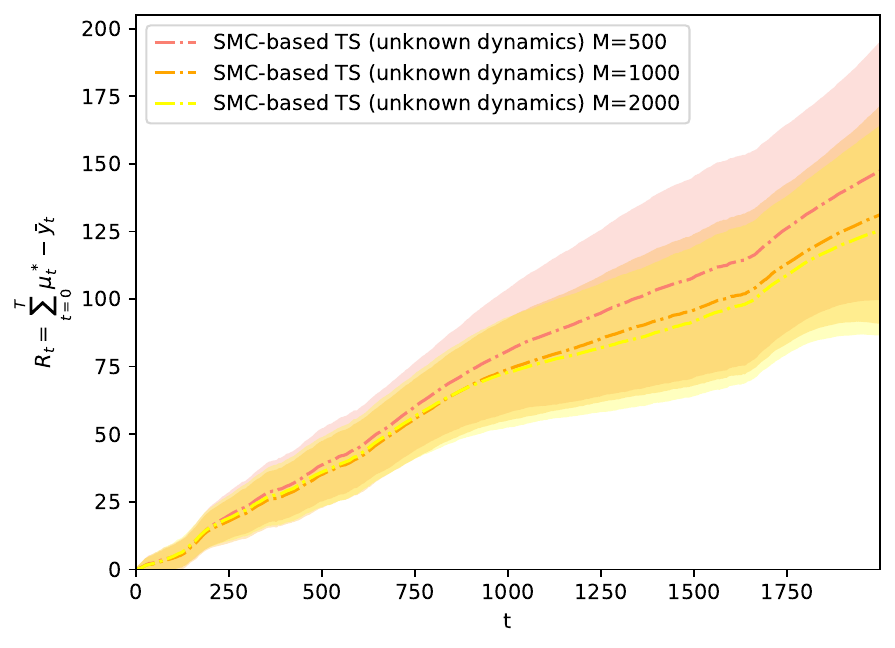}
		\caption{Cumulative regret for SMC-based TS in scenario D: unknown dynamic parameters $L_a,\Sigma_a,\sigma_a^2, \forall a$.}
		\label{fig:dynamic_bandits_logistic_d_ts_dunknown_M}
	\end{subfigure}\qquad
	\begin{subfigure}[b]{0.45\textwidth}
		\includegraphics[width=\textwidth]{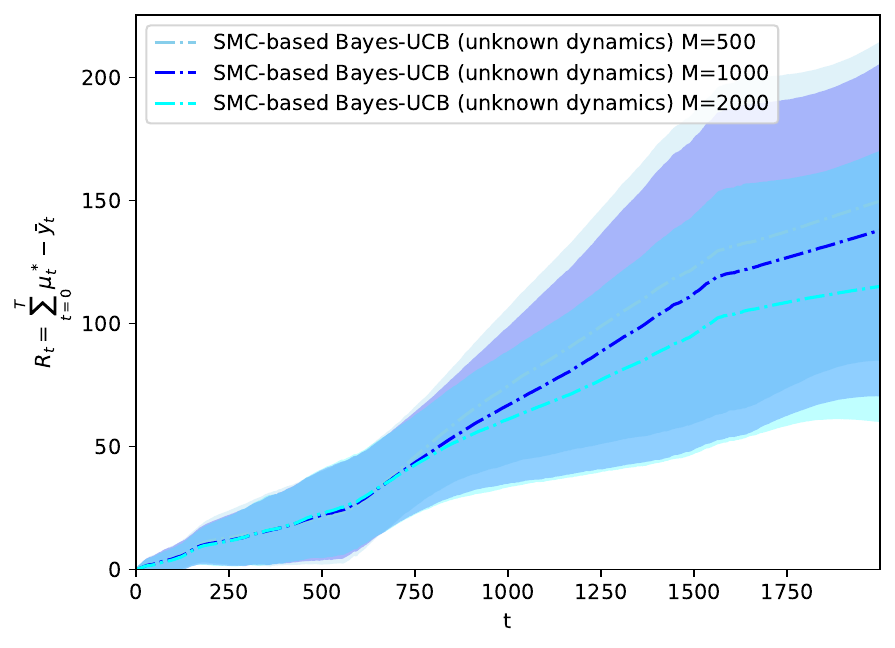}
		\caption{Cumulative regret for SMC-based Bayes-UCB in scenario D: unknown dynamic parameters $L_a,\Sigma_a,\sigma_a^2, \forall a$.}
		\label{fig:dynamic_bandits_logistic_d_bucb_dunknown_M}
	\end{subfigure}
	
	\caption{
		Mean regret (standard deviation shown as the shaded region) in contextual, logistic bandit Scenario D
		described in Equation~\eqref{eq:linear_mixing_dynamics_d}.
		SMC-based policies' averaged cumulative regret is robust to different Monte Carlo sample sizes $M$,
		which impacts mostly the performance variability for $M=500$ ---specially so when optimal arm swaps occur later in time.
	}
	\label{fig:dynamic_bandits_logistic_d_M}
\end{figure}

\clearpage
We assess in Figure~\ref{fig:dynamic_bandits_logistic_d_M} the impact of Monte Carlo sample size $M$ in the performance of the proposed SMC-based Bayesian MAB policies
in Scenario D defined by Equation~\eqref{eq:linear_mixing_dynamics_c},
for a realization of expected rewards as depicted in Figure~\ref{fig:linear_mixing_dynamics_d_logistic}.

\clearpage
\subsection{Non-stationary, categorical rewards}
\label{assec:dynamic_bandits_categorical}

We assess below the impact of Monte Carlo sample size $M$ in the performance of the proposed SMC-based Bayesian MAB policies
in Scenario E defined by Equation~\eqref{eq:linear_mixing_dynamics_e},
for a realization of expected rewards as depicted in Figure~\ref{fig:linear_mixing_dynamics_e_softmax}.

\begin{figure}[!h]
\centering
\begin{subfigure}[b]{0.45\textwidth}
\includegraphics[width=\textwidth]{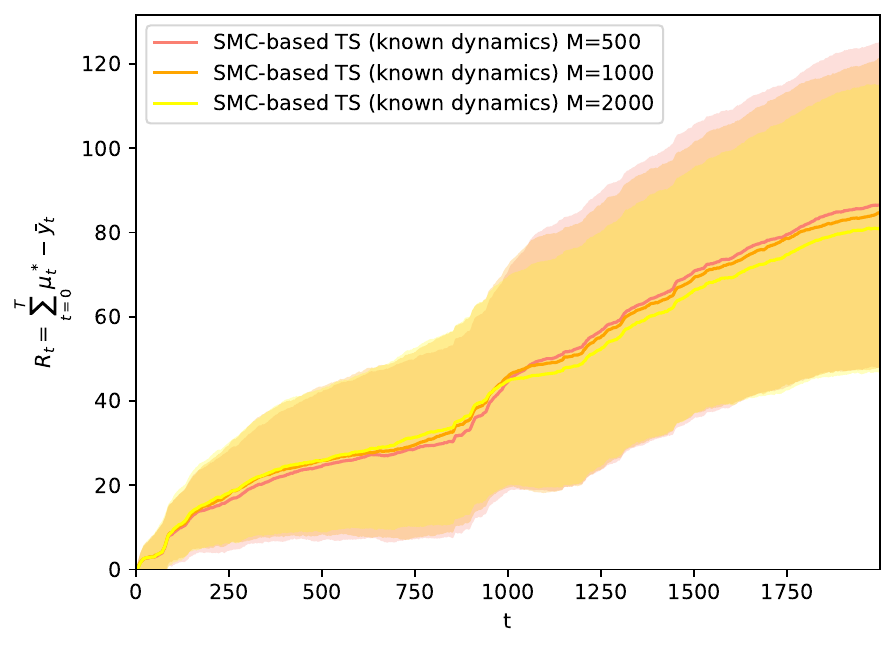}
\caption{Cumulative regret for SMC-based TS in scenario E: known dynamic parameters.}
\label{fig:dynamic_bandits_softmax_e_ts_dknown_M}
\end{subfigure}\qquad
\begin{subfigure}[b]{0.45\textwidth}
\includegraphics[width=\textwidth]{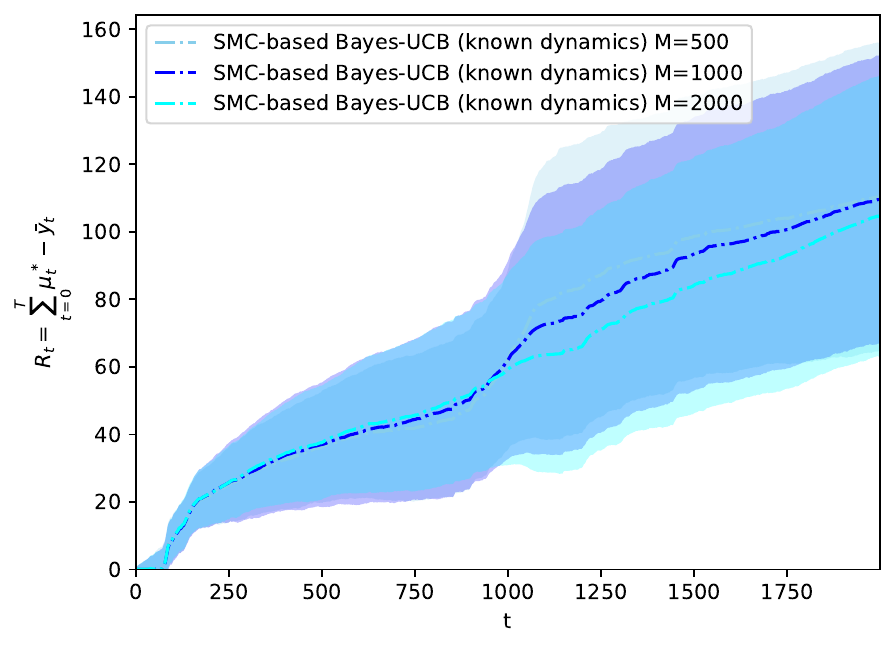}
\caption{Cumulative regret for SMC-based Bayes-UCB in scenario E: known dynamic parameters.}
\label{fig:dynamic_bandits_softmax_e_bucb_dknown_M}
\end{subfigure}

\begin{subfigure}[b]{0.45\textwidth}
\includegraphics[width=\textwidth]{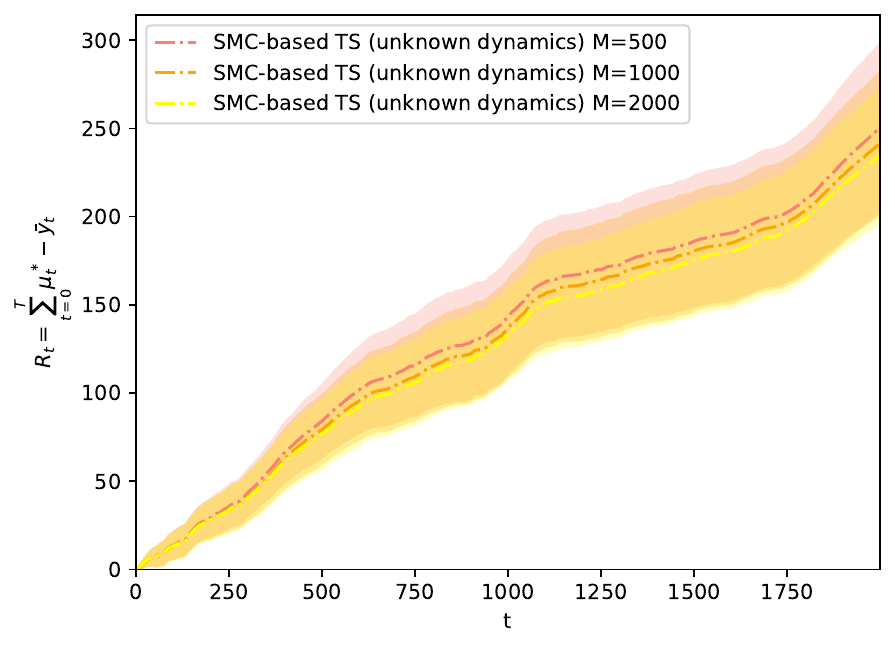}
\caption{Cumulative regret for SMC-based TS in scenario E: unknown dynamic parameters $L_a,\Sigma_a,\sigma_a^2, \forall a$.}
\label{fig:dynamic_bandits_softmax_e_ts_dunknown_M}
\end{subfigure}\qquad
\begin{subfigure}[b]{0.45\textwidth}
\includegraphics[width=\textwidth]{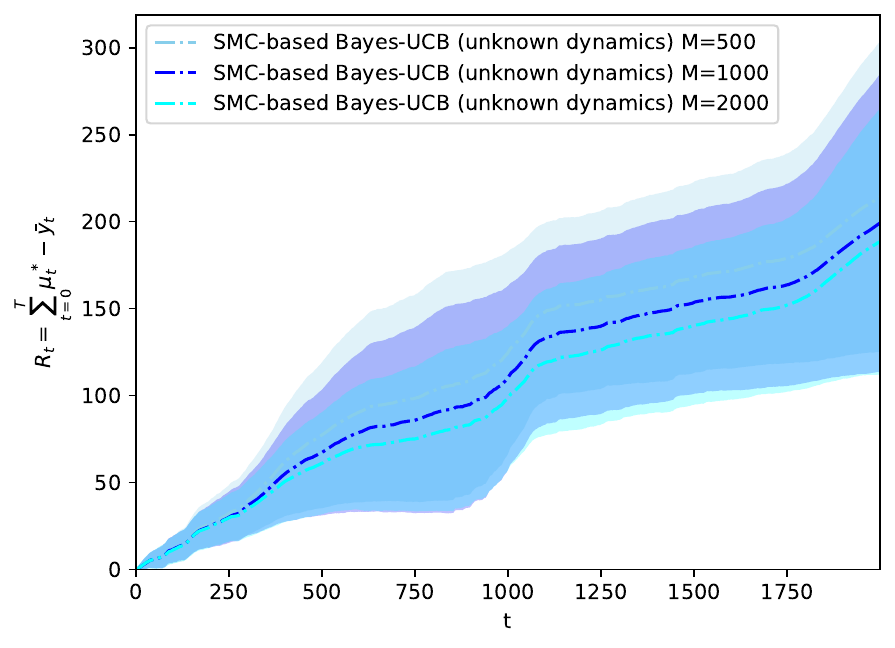}
\caption{Cumulative regret for SMC-based Bayes-UCB in scenario E: unknown dynamic parameters $L_a,\Sigma_a,\sigma_a^2, \forall a$.}
\label{fig:dynamic_bandits_softmax_e_bucb_dunknown_M}
\end{subfigure}

\caption{
Mean regret (standard deviation shown as the shaded region) in contextual, softmax bandit Scenario E
described in Equation~\eqref{eq:linear_mixing_dynamics_e}.
SMC-based policies' averaged cumulative regret is robust to different Monte Carlo sample sizes $M$,
which slightly impacts the performance variability for $M=500$.
}
\label{fig:dynamic_bandits_softmax_e_M}
\end{figure}

\clearpage
We assess below the impact of Monte Carlo sample size $M$ in the performance of the proposed SMC-based Bayesian MAB policies
in Scenario F defined by Equation~\eqref{eq:linear_mixing_dynamics_f},
for a realization of expected rewards as depicted in Figure~\ref{fig:linear_mixing_dynamics_f_softmax}.

\begin{figure}[!h]
\centering
\begin{subfigure}[b]{0.45\textwidth}
\includegraphics[width=\textwidth]{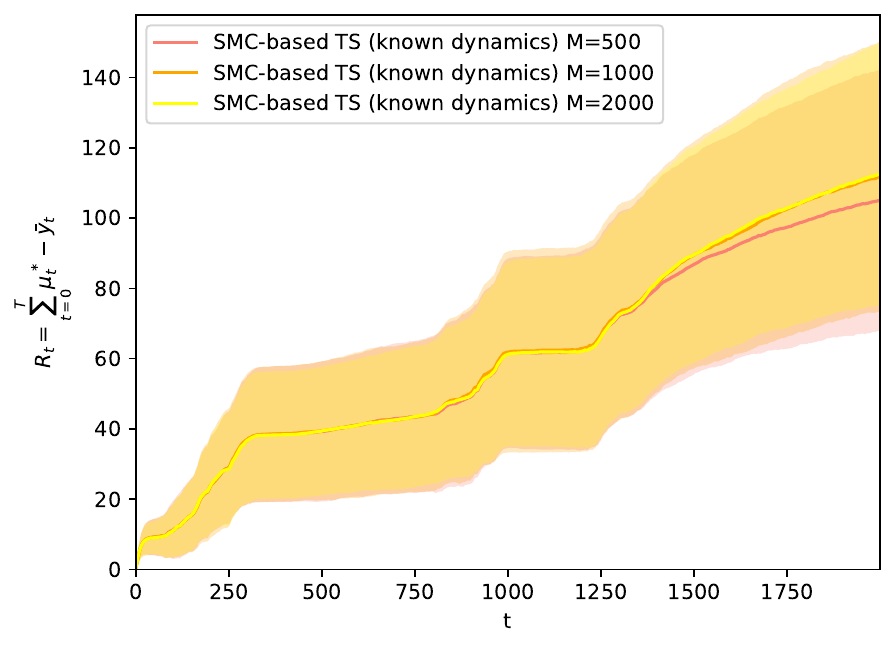}
\caption{Cumulative regret for SMC-based TS in scenario F: known dynamic parameters.}
\label{fig:dynamic_bandits_softmax_f_ts_dknown_M}
\end{subfigure}\qquad
\begin{subfigure}[b]{0.45\textwidth}
\includegraphics[width=\textwidth]{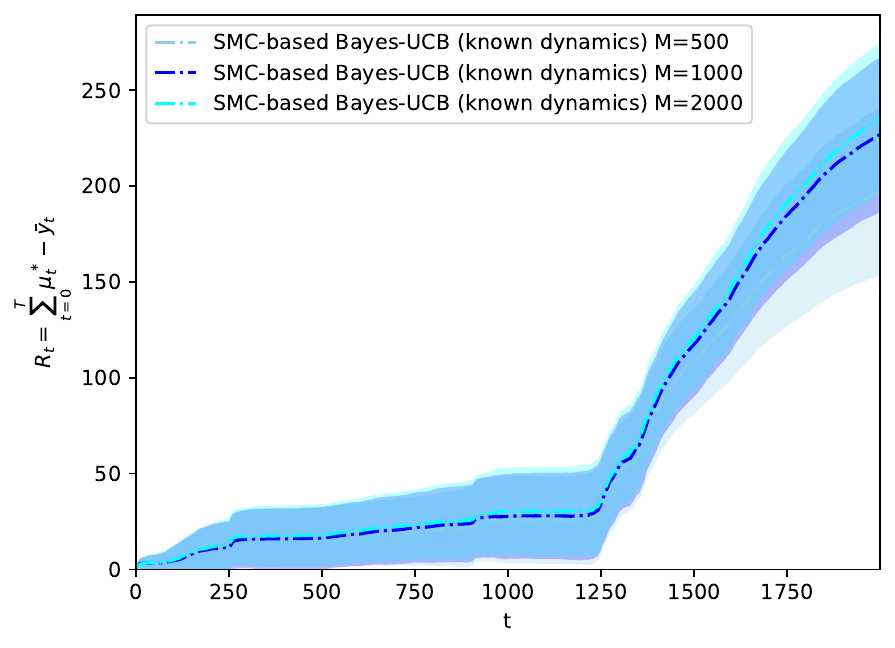}
\caption{Dumulative regret for SMC-based Bayes-UCB in scenario F: known dynamic parameters.}
\label{fig:dynamic_bandits_softmax_f_bucb_dknown_M}
\end{subfigure}

\begin{subfigure}[b]{0.45\textwidth}
\includegraphics[width=\textwidth]{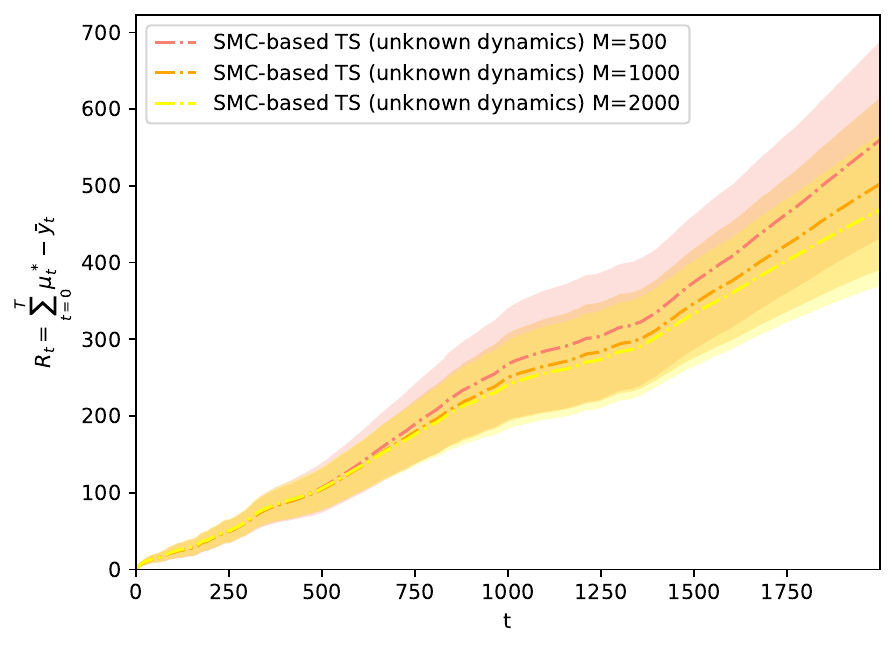}
\caption{Cumulative regret for SMC-based TS in scenario F: unknown dynamic parameters $L_a,\Sigma_a,\sigma_a^2, \forall a$.}
\label{fig:dynamic_bandits_softmax_f_ts_dunknown_M}
\end{subfigure}\qquad
\begin{subfigure}[b]{0.45\textwidth}
\includegraphics[width=\textwidth]{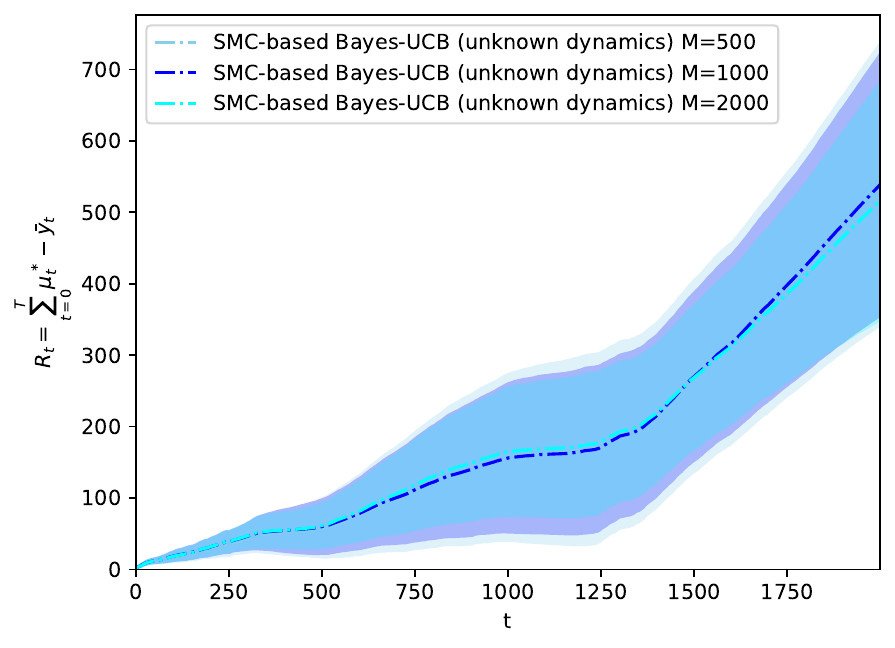}
\caption{Cumulative regret for SMC-based Bayes-UCB in scenario F: unknown dynamic parameters $L_a,\Sigma_a,\sigma_a^2, \forall a$.}
\label{fig:dynamic_bandits_softmax_f_bucb_dunknown_M}
\end{subfigure}

\caption{
Mean regret (standard deviation shown as the shaded region) in contextual, softmax bandit Scenario F
described in Equation~\eqref{eq:linear_mixing_dynamics_f}.
SMC-based policies' averaged cumulative regret is robust to different Monte Carlo sample sizes $M$,
which slightly impacts the performance variability for $M=500$.
}
\label{fig:dynamic_bandits_softmax_f_M}
\end{figure}

\end{document}